\documentclass{article}

\PassOptionsToPackage{numbers, compress}{natbib}

\usepackage[main, final]{neurips_2026}
\usepackage[utf8]{inputenc} 
\usepackage[T1]{fontenc}    
\usepackage{url}            
\usepackage{booktabs}       
\usepackage{graphicx}       
\usepackage{afterpage}      
\usepackage{amsmath}        
\usepackage{amsfonts}       
\usepackage{amssymb}        
\usepackage{nicefrac}       
\usepackage{microtype}      
\usepackage[table]{xcolor}         
\usepackage{subfiles}
\usepackage{titletoc}
\usepackage{enumitem}
\usepackage{wrapfig}
\usepackage{float}
\usepackage{flafter}
\usepackage{algorithm}
\usepackage{listings}
\usepackage{capt-of}        
\usepackage{array}
\usepackage{ragged2e}

\usepackage{soul}
\usepackage{hyperref}       

\lstdefinestyle{torchalgorithm}{
  language=Python,
  basicstyle=\ttfamily\footnotesize,
  commentstyle=\color{gray!70!black},
  keywordstyle=\color{blue!60!black},
  stringstyle=\color{green!40!black},
  showstringspaces=false,
  columns=fullflexible,
  keepspaces=true,
  breaklines=true,
  aboveskip=0.25em,
  belowskip=0pt
}

\newcommand\DoToC{%
  \startcontents
  \printcontents{}{1}{\vskip3pt\hrule\vskip5pt}
  \vskip15pt\hrule\vskip5pt
}

\providecommand{\ctxyes}{\textcolor{green!60!black}{\ensuremath{\checkmark}}}
\providecommand{\ctxno}{\textcolor{gray}{--}}
\providecommand{\ctxpartial}[1]{\textcolor{orange}{$\triangle$}~(#1)}

\newcommand{\proposed}{{\sffamily scTrilemma}}
\definecolor{methodhighlight}{HTML}{E2ECF7} 

\newcolumntype{L}[1]{>{\RaggedRight\arraybackslash}p{#1}}

\title{scTrilemma: Balancing Identity, Invariance, and Fidelity in Single-Cell Representation Learning}

\author{
    Yunhak Oh$^{1}$\thanks{Work done during an internship at HITS}, Yoonho Lee$^{1}$, Junseok Lee$^{1}$, Namkyeong Lee$^{1}$, 
    \\ \textbf{Sang-Yeon Hwang$^{2}$, Yinhua Piao$^{1}$, Hyomin Kim$^{1}$, Seonghwan Kim$^{1}$, Jaechang Lim$^{2}$, }
    \\ \textbf{Woo Youn Kim$^{1,2}$, Sungsoo Ahn$^{1}$, Chanyoung Park$^{1}$\thanks{Corresponding Author}} 
    \\ $^{1}$ KAIST ~~~ $^{2}$ HITS ~~~
    \\ \texttt{\{yunhak.oh, cy.park\}@kaist.ac.kr}
}

\begin{document}

\maketitle

\begin{abstract}
  Single-cell RNA-seq representation learning is fundamentally label-free: cell identities, states, and contexts are not fixed training targets, so what constitutes signal or nuisance is analysis-dependent.
  A single representation must therefore preserve biological identity and state, remain robust to nuisance context, and retain the gene-level variation needed for expression analysis, three demands we call the \textit{representation trilemma}.
  To tackle this problem, we introduce \proposed{}, a latent-bottleneck VAE that routes expression-derived variation to the embedding, the decoder, or the prior rather than forcing all of it through one embedding.
  It gates gene tokens by expression, routes the cell representation through the decoder, and conditions the prior on unlabeled pseudo-bulk context, under a single reconstruction objective and without target annotations or auxiliary representation losses.
  In release-based zero-shot evaluation on successive CZ CELLxGENE Census releases, \proposed{} leads all three demands at once and preserves biological-state, differential-expression, and pathway structure across multiple disease settings.
  Latent interventions further show that context can be removed at almost no cost to the other demands, leaving identity against fidelity as the remaining tension.
  Code is publicly available at \textcolor{magenta}{\url{https://github.com/yunhak0/scTrilemma}}.
\end{abstract}


\section{Introduction}
\label{sec:introduction}
\begin{wrapfigure}[15]{r}{0.45\textwidth}
    \centering
    \raisebox{0pt}[\dimexpr\height-1.0\baselineskip\relax]
    {\includegraphics[width=0.9\linewidth]{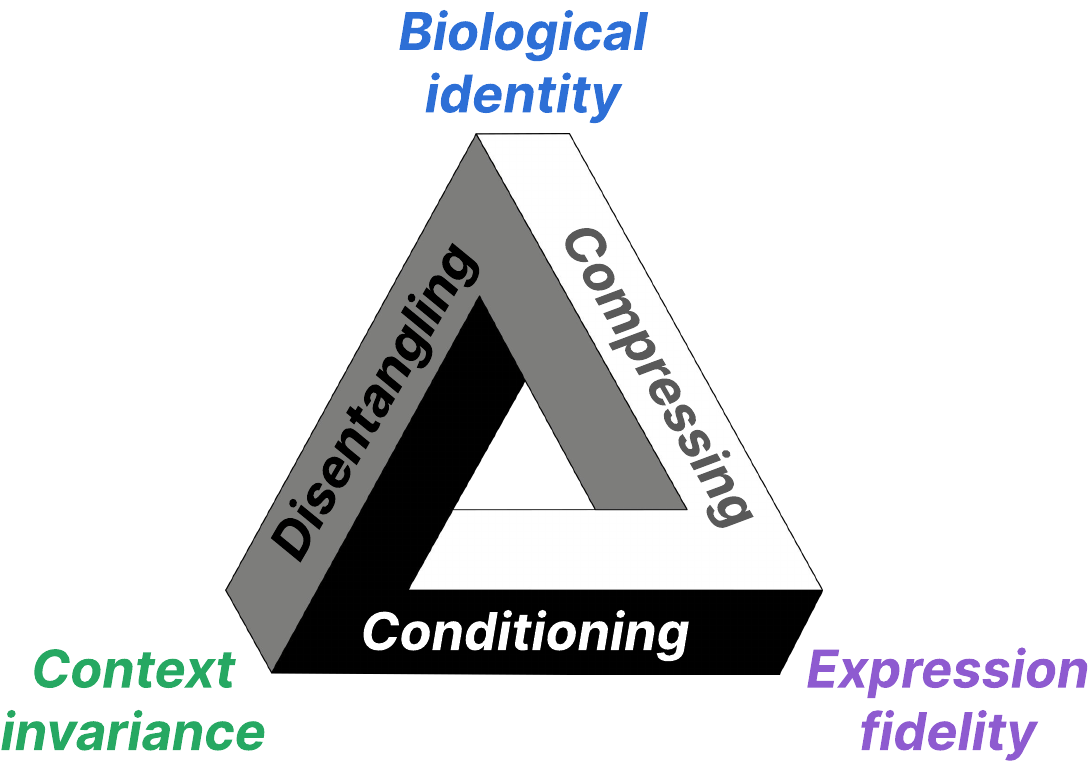}}
    \caption{The \textit{representation trilemma}; edge labels name the operation each pair contends over.
    \textbf{Biological Identity} preserves cell-type and state structure.
    \textbf{Context Invariance} prevents nuisance context from defining cellular similarity.
    \textbf{Expression Fidelity} retains gene-level variation needed for biological analysis.}
    \label{fig:representation-trilemma}
\end{wrapfigure}

In single-cell RNA-seq, expression profiles mix many sources of variation, and whether a given source is signal or nuisance depends on the analysis: donor differences are a confounder when matching cell types across studies~\citep{luecken2022benchmarking,the2022tabula} and the signal when comparing patients~\citep{mathys2019single,boyeau2025deep}.
Cell identity and state are themselves assigned only after clustering and annotation~\citep{luecken2019current}, so representation learning is label-free and no annotation fixes these roles in advance.
The only information available to the model is therefore the expression profile itself, in which all of these sources of variation are entangled.
Not all of this variation should define cellular similarity, and a label-free model must decide which does, which should be treated as context, and which should remain available for expression-level analysis.

We define the tension among these roles as the \textit{representation trilemma}, illustrated in Figure~\ref{fig:representation-trilemma}:

\begin{itemize}[leftmargin=.1in,itemsep=1pt,topsep=2pt]
    \vspace{-1ex}
    \item \textbf{Biological Identity}: preserve cell-type and biological-state structure for cellular comparison.
    \item \textbf{Context Invariance}: remain robust to nuisance context without removing tissue-, disease-, or donor-associated biology by default.
    \item \textbf{Expression Fidelity}: retain gene-level variation needed for expression prediction and downstream analysis, including variation that need not define cellular similarity.
    \vspace{-1ex}
\end{itemize}
A uniform preserve-or-remove rule therefore cannot serve all three, and existing models lead one demand at the cost of another.
We argue that part of this conflict is not inherent but arises because all variation is forced through one embedding.
We therefore frame an \textit{information-routing} problem: which expression-derived variation should shape the embedding, condition the model, or remain available for gene-level prediction?

To tackle this problem, we introduce \proposed{}, a VAE whose latent bottleneck forces selective compression and whose encoder, decoder, and prior provide distinct routes for cell comparison, expression prediction, and context conditioning.
Reconstruction supplies a label-free expression-level learning signal, while classical VAE backbones remain competitive with larger foundation models in zero-shot settings~\citep{kedzierska2025zero}.
Within this architecture, \textbf{E-Gate} controls what reaches the evaluated embedding by modulating gene-identity features with expression before compression.
\textbf{C-Route} makes the evaluated cell representation an active part of reconstruction by routing it through the decoder.
\textbf{PB-Cond} places unlabeled pseudo-bulk context in the VAE prior rather than in the embedding.
Together, they route expression-derived information without cell labels, metadata supervision, or auxiliary representation losses.
We evaluate \proposed{} under a release-based zero-shot protocol on 89 datasets first appearing in a later CZ CELLxGENE Census release~\citep{czi2025cz}.
The evaluation covers all three demands, from embedding-level identity and context to reconstruction-level differential-expression and pathway fidelity.
In summary, our contributions are:
\begin{itemize}[leftmargin=.1in,itemsep=1pt,topsep=2pt]
    \vspace{-1ex}
    \item To our knowledge, we are the first to formulate label-free single-cell representation learning as a \textit{representation trilemma} among biological identity, context invariance, and expression fidelity, and to recast this tension as an information-routing problem.

    \item We introduce \proposed{}, a reconstruction-trained latent-bottleneck VAE that implements complementary expression-derived routing across the encoder, decoder, and prior.

    \item Release-based zero-shot evaluation shows that \proposed{} leads all three demands at once, and demand-targeted interventions show that routing removes nearly all of the cost of context invariance and leaves identity against fidelity as the remaining tension.
\end{itemize}

\section{Related works}
\label{sec:related}
\textbf{Single-cell representation learning.}
Deep generative models such as scVI learn cell embeddings from count vectors and condition reconstruction on known batch covariates~\citep{lopez2018deep}.
Pretrained models commonly use gene-token objectives.
scBERT bins expression for masked-token prediction~\citep{yang2022scbert}, 
Geneformer represents cells as rank-ordered gene sequences~\citep{theodoris2023transfer}, 
and scGPT jointly learns gene-value prediction and cell-embedding consistency~\citep{cui2024scgpt}.
CellPLM aggregates expression-weighted gene embeddings into cell tokens before modeling cell--cell relations within tissue-level groups~\citep{wen2024cellplm}.
scPRINT combines the protein-informed gene embeddings of UCE~\citep{rosen2026universal} with denoising and ontology- or metadata-supervised objectives~\citep{kalfon2025scprint}.
Rather than a new gene-level objective, our focus is the label-free question of how to route expression-derived information 
among cellular comparison, context conditioning, and gene-level prediction without requiring the evaluated embedding to serve all three.

\textbf{Integration and batch-correction trade-offs.}
Integration methods control how collection-associated variation enters cell representations.
Latent-variable models condition reconstruction on batch or study covariates~\citep{lopez2018deep,xu2021probabilistic,lotfollahi2022mapping}, 
population-aware models such as scPoli and MrVI incorporate sample- or donor-level structure~\citep{de2023population,boyeau2025deep}, 
and post-hoc methods align embeddings or neighborhood graphs~\citep{korsunsky2019fast,polanski2020bbknn,hie2019efficient}.
Learned batch or sample identifiers are less direct in release-based zero-shot settings, where new datasets arrive without target-specific identifiers.
Tissue context may capture both collection-level expression structure and tissue-resident programs rather than act only as a nuisance.
Integration benchmarks evaluate biological conservation and batch correction along distinct axes~\citep{luecken2022benchmarking}, and model objectives and regularization shift the balance between them~\citep{yi2025benchmarking}.
This embedding-level view leaves open how variation that need not define cellular similarity can remain available for expression prediction and downstream analysis.

\section{Preliminaries}
\label{sec:preliminaries}
\subsection{Problem statement}

\textbf{Notation.}
Let $\mathcal{G}$ denote the global gene vocabulary, and let $\mathbf{x}_i$ be the observed gene-count vector for cell $i$, typically sparse in scRNA-seq.
During training, we represent each cell using a gene subset $S_i \subset \mathcal{G}$, with gene indices $g_{ij} \in S_i$ and scalar raw counts $x_{ij} \in \mathbb{R}_{\ge 0}$.
We 
use $\ell_i$ to denote the raw-count total over the modeled gene subset $S_i$, which serves as the library-size factor for the count likelihood.
We distinguish raw counts $x_{ij}$, used as reconstruction targets for the count likelihood, from library-normalized and log-transformed values $\tilde{x}_{ij}$, used to construct encoder gene-expression tokens.
The exact normalization is described in Appendix~\ref{app:preprocessing}.

\textbf{Group-derived pseudo-bulk summaries.}
Let $b_i$ denote the dataset--donor group containing cell $i$, and let $\mathcal{C}_b = \{c : b_c=b\}$ denote the cells in group $b$. We use $\mathbf{p}_b \in \mathbb{R}^{|\mathcal{G}|}$ to denote the group's pseudo-bulk expression signature, obtained by averaging per-cell library-normalized expression profiles over $\mathcal{C}_b$ and applying an elementwise log transformation. Dataset and donor identifiers determine only the aggregation set; cell-type, disease, and tissue annotations are not used to construct $\mathbf{p}_b$. Its exact construction is provided in Appendix~\ref{app:pseudobulk-details}.

\textbf{Representation-learning task.}
Given single-cell expression profiles $(\mathbf{x}_i,S_i)$ without cell-level labels, we learn a fixed-dimensional cell embedding $\mathbf{r}_i$ for downstream cellular comparison. At evaluation, the embedding is used without target-label supervision or parameter updates to assess biological identity, neighborhood structure, and comparability across collection contexts. Gene-level reconstruction provides the label-free learning signal and is evaluated separately for expression fidelity. Expression fidelity itself does not presuppose a reconstruction objective; any expression-level objective, including denoising or masked prediction, yields an expression output to which the same demand applies, and we instantiate it with reconstruction as the native signal of a VAE.
The central requirement is not to carry every expression-derived signal through $\mathbf{r}_i$, but to effectively allocate the signal across token construction, the evaluated embedding, and reconstruction or conditioning pathways.

\subsection{Latent-Bottleneck Variational Backbone}

\textbf{Gene-expression tokens.}
For each gene $g_{ij}\in S_i$, the backbone constructs a gene-expression token from its gene identity and normalized expression value $\tilde{x}_{ij}$ through a token embedding function $\psi$:
\begin{equation*}
    \mathbf{e}_{ij} = \psi(g_{ij}, \tilde{x}_{ij}) \in \mathbb{R}^d.
\end{equation*}
The exact form of $\psi$ is a token-construction choice studied in our method. The backbone only requires that it produce a token embedding sequence $\mathbf{E}_i = (\mathbf{e}_{ij})_{j=1}^{|S_i|}$.

\textbf{Perceiver-style latent bottleneck.}
Instead of processing all genes into a single pooled vector, the encoder compresses the gene-token set through $K$ learned latent queries $\mathbf{Q} \in \mathbb{R}^{K \times d}$~\citep{jaegle2021perceiver}. Each encoder block updates the latent queries using pre-normalized cross-attention followed by a residual SwiGLU feed-forward sublayer~\citep{shazeer2020glu}:
{\small
\begin{equation*}
\begin{aligned}
    \widetilde{\mathbf{H}}_i^{(l)}
    &=
    \mathbf{H}_i^{(l)}
    +
    \mathrm{CrossAttn}\!\left(
        \mathrm{Norm}(\mathbf{H}_i^{(l)}),
        \mathrm{Norm}(\mathbf{E}_i)
    \right), \\
    \mathbf{H}_i^{(l+1)}
    &=
    \widetilde{\mathbf{H}}_i^{(l)}
    +
    \mathrm{FFN}_{\mathrm{SwiGLU}}\!\left(
        \mathrm{Norm}(\widetilde{\mathbf{H}}_i^{(l)})
    \right),
    \qquad \mathbf{H}_i^{(0)}=\mathbf{Q}.
\end{aligned}
\end{equation*}
}
This gives a fixed-size latent bottleneck $\mathbf{H}_i \in \mathbb{R}^{K \times d}$ independent of the number of input gene positions, creating a fixed-capacity compression interface shared by the evaluated cell embedding and the reconstruction pathway.
The active configuration uses $K=64$ latent tokens and gene crops of size $4096$.
Detailed architecture settings are provided in Appendix~\ref{app:model-architecture-hparams}, and Appendix~\ref{app:latent-query-analysis} analyzes the gene programs captured by the latent queries.

\begin{figure*}[t]
    \centering
    \includegraphics[width=0.99\linewidth]{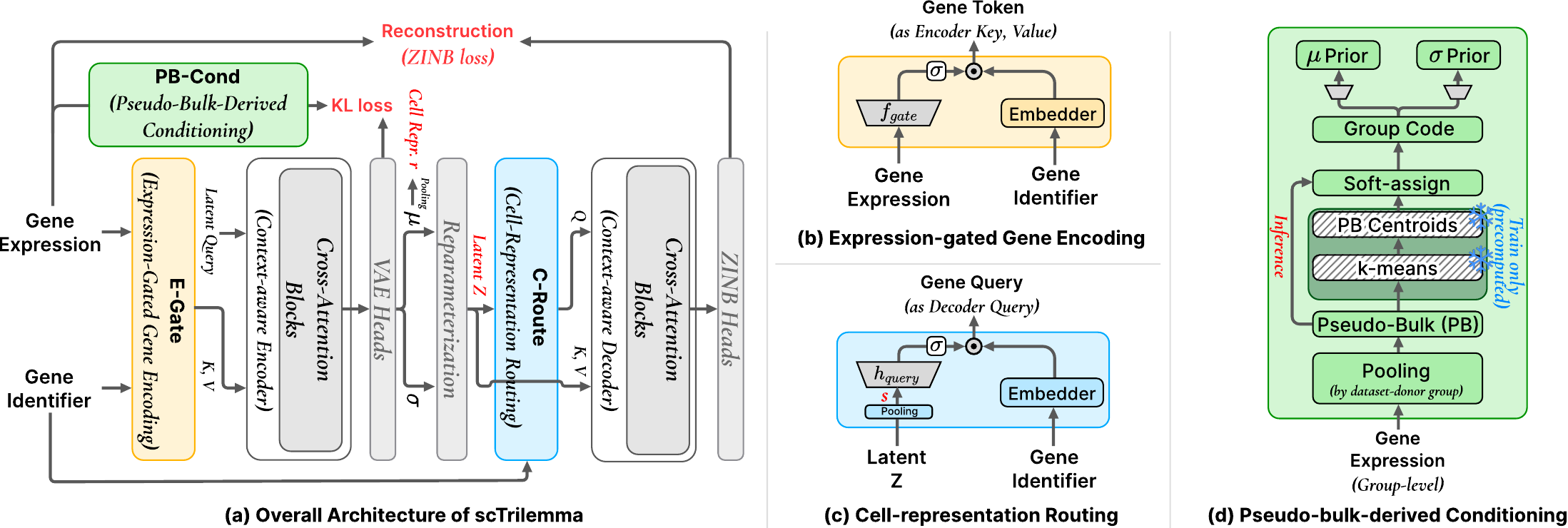}
    \caption{\proposed{} architecture.
    (a) Overall routing architecture trained with ZINB reconstruction and KL regularization.
    (b) E-Gate applies an expression-derived feature-wise gate before latent compression.
    (c) C-Route mean-pools posterior means for the evaluated cell representation and uses the pooled latent sample to modulate decoder queries, while full sampled latent tokens remain reconstruction memory.
    (d) PB-Cond softly assigns dataset--donor pseudo-bulks to fixed pretraining centroids to parameterize the conditional prior.}
    \label{fig:model-architecture}
    \vspace{-2ex}
\end{figure*}

\textbf{Variational posterior.}
The latent bottleneck parameterizes a diagonal Gaussian posterior over latent tokens:
{\small
\begin{equation*}
    q_\phi(\mathbf{Z}_i \mid \mathbf{x}_i, S_i) =
    \prod_{k=1}^{K}
    \mathcal{N}\left(\mathbf{z}_{ik}; \boldsymbol{\mu}_{ik}, \mathrm{diag}(\boldsymbol{\sigma}^2_{ik})\right),
\end{equation*}}
where $\boldsymbol{\mu}_i$ and $\log \boldsymbol{\sigma}_i^2$ are linear projections of the encoder output.
We sample $\mathbf{Z}_i$ using the standard reparameterization $\mathbf{Z}_i = \boldsymbol{\mu}_i + \boldsymbol{\sigma}_i \odot \boldsymbol{\epsilon}$, where $\boldsymbol{\epsilon} \sim \mathcal{N}(0,I)$.

\textbf{Cross-attention decoder and count likelihood.}
The decoder reconstructs the target genes using gene embeddings as base queries and the latent tokens $\mathbf{Z}_i$ as cross-attention memory~\citep{jaegle2022perceiver}.
After stacked decoder cross-attention blocks, the decoder parameterizes a ZINB likelihood over raw counts, following common scRNA-seq latent-variable and denoising models~\citep{lopez2018deep,eraslan2019single}. Likelihood and parameterization details are provided in Appendix~\ref{app:count-likelihood}. The pretraining objective is
\begin{equation}
    \mathcal{L}_{\mathrm{pretrain},i}
    =
    \mathcal{L}_{\mathrm{ZINB},i}
    +
    \lambda_{\mathrm{KL}}
    \mathcal{L}_{\mathrm{KL},i}.
    \label{eq:pretrain-objective}
\end{equation}
with a warm-up schedule for $\lambda_{\mathrm{KL}}$. 
Here $\mathcal{L}_{\mathrm{ZINB},i}$ is the reconstruction loss for $\mathbf{x}_i$, and $\mathcal{L}_{\mathrm{KL},i}$ is the KL divergence between its posterior and the active prior.
The default backbone uses a standard normal prior, while Section~\ref{sec:method-pbcond} introduces an expression-derived group-conditioned prior. 
This backbone serves as the shared foundation for the three routing components described in Section~\ref{sec:methodology}.

\section{Methodology}
\label{sec:methodology}

This section describes how \proposed{} routes expression-derived information within a latent-bottleneck VAE, summarized in Figure~\ref{fig:model-architecture}(a). 
The method comprises three routing components. 
Expression-gated gene encoding (E-Gate) modulates gene-identity features with expression before compression and controls what reaches the latent bottleneck and evaluated embedding, as described in Section~\ref{sec:method-egate}.
Cell-representation routing (C-Route) defines the evaluated cell embedding and couples it to gene-level reconstruction through decoder-query modulation while retaining token-level latent detail for prediction, as described in Section~\ref{sec:method-croute}.
Pseudo-bulk-derived conditioning (PB-Cond) derives group-level context from unlabeled pseudo-bulk expression and uses it to condition the VAE prior, as described in Section~\ref{sec:method-pbcond}.
Together, these components keep the reconstruction objective fixed while changing where and how expression-derived signals enter the encoder, decoder, and prior. 
PyTorch-style pseudocode for the full training step is provided in Appendix~\ref{app:pytorch-pseudocode}.

\subsection{Expression-Gated Gene Encoding (E-Gate)}
\label{sec:method-egate}

E-Gate controls how measured expression enters gene tokens before compression, as shown in Figure~\ref{fig:model-architecture}(b). A common alternative adds an expression embedding to the learned gene-identity embedding~\citep{cui2024scgpt}. This gives expression an independent additive contribution that can change both the direction and norm of the gene token before bottleneck compression.

E-Gate instead uses an expression-derived feature-wise gate:
\begin{equation}
    \mathbf{e}_{ij}^{\mathrm{gate}}
    =
    \mathbf{u}_{g_{ij}}
    \odot
    \sigma\!\left(f_{\mathrm{gate}}(\tilde{x}_{ij})\right),
    \qquad
    f_{\mathrm{gate}}:\mathbb{R}\rightarrow\mathbb{R}^{d},
    \label{eq:method-egate}
\end{equation}
Here, $\mathbf{u}_{g_{ij}}\in\mathbb{R}^{d}$ is a gene-identity embedding learned during pretraining, $\sigma$ is the element-wise sigmoid, and $f_{\mathrm{gate}}$ is an affine projection of $\tilde{x}_{ij}$ learned jointly during pretraining.
Because each gate coordinate lies between zero and one, $\lVert \mathbf{e}_{ij}^{\mathrm{gate}} \rVert \leq \lVert \mathbf{u}_{g_{ij}} \rVert$. Thus, expression reweights features of the gene-identity anchor without contributing an independent additive vector.

This constraint gives E-Gate an inductive bias. Gene identity provides the token anchor, while expression controls the feature-wise gain on that anchor.
Gate attenuation in the trained model is largest for genes that are both highly expressed and cell-type-specific, so E-Gate acts as a token-level filter that limits shared expression magnitude while retaining marker-aligned signal.
The contribution of E-Gate to the full model is evaluated by the ablation in Figure~\ref{fig:component-ablation}, and its gain-control mechanism is examined in Appendix~\ref{app:egate-analysis}.

\subsection{Cell-Representation Routing (C-Route)}
\label{sec:method-croute}

C-Route controls whether the pooled representation used for cellular comparison participates in reconstruction, as shown in Figure~\ref{fig:model-architecture}(c).
In a multi-token bottleneck, the decoder can reconstruct from the full latent-token state while the pooling operation that defines the evaluated representation remains post hoc.
C-Route therefore reuses the native ZINB decoder without introducing an auxiliary cell-to-gene prediction head.

The posterior mean tokens $\boldsymbol{\mu}_i\in\mathbb{R}^{K\times d}$ define the deterministic cell representation by mean pooling, $\mathbf{r}_i=K^{-1}\sum_{k=1}^{K}\boldsymbol{\mu}_{ik}$.
Mean pooling keeps the evaluated representation non-parametric, so no learned evaluation head can selectively emphasize tokens, and every difference between cells reflects the posterior itself.
For reconstruction, the sampled tokens $\mathbf{Z}_i\in\mathbb{R}^{K\times d}$ remain available as decoder memory, 
while their mean $\mathbf{s}_i=K^{-1}\sum_{k=1}^{K}\mathbf{z}_{ik}$ is the stochastic counterpart of $\mathbf{r}_i$,
satisfying $\mathbb{E}_{q_\phi(\mathbf{Z}_i\mid\mathbf{x}_i,S_i)}[\mathbf{s}_i]=\mathbf{r}_i$.
C-Route uses $\mathbf{s}_i$ to modulate each gene-identity query:
\begin{equation}
    \mathbf{q}_{ij}
    =
    \mathbf{u}_{g_{ij}}
    \odot
    \sigma\!\left(h_{\mathrm{query}}(\mathbf{s}_i)\right),
    \qquad
    h_{\mathrm{query}}:\mathbb{R}^{d}\rightarrow\mathbb{R}^{d}.
    \label{eq:method-croute-query}
\end{equation}
Here, $h_{\mathrm{query}}$ is a two-layer MLP with GELU between its linear layers, and $\sigma$ is the element-wise sigmoid.
This construction creates two complementary reconstruction routes.
The pooled route receives reconstruction gradients through the decoder queries under the same ZINB objective, while the full latent-token memory retains token-level expression detail.
C-Route thus couples the evaluated representation to reconstruction without making it the sole carrier of gene-level information or introducing an auxiliary loss.
Pooling sets cross-donor comparability while the decoder query sets identity and gene-level fidelity together in the analysis of Appendix~\ref{app:croute-analysis}, so the two routes trade against different demands rather than competing for the same one.
The contribution of decoder-query routing to the full model is evaluated by the ablation in Figure~\ref{fig:component-ablation}, 
and its routing mechanism is examined in Appendix~\ref{app:croute-analysis}.

\subsection{Pseudo-Bulk-Derived Conditioning (PB-Cond)}
\label{sec:method-pbcond}

PB-Cond controls where group-level expression context enters the VAE, as shown in Figure~\ref{fig:model-architecture}(d).
Direct context injection into the encoder or decoder can create a group-to-reconstruction shortcut, since group-level expression summaries carry study-specific effects alongside biological state~\citep{squair2021confronting,liu2026learning}.
PB-Cond instead uses expression-derived group context exclusively to condition the KL prior.

For each dataset--donor group $b$, PB-Cond maps its pseudo-bulk signature $\mathbf{p}_b$ to an expression-derived code.
We fit $M=32$ {k-means} centroids to pseudo-bulk signatures from the pretraining corpus and keep them fixed thereafter.
Soft assignment to these centroids produces $\mathbf{t}_b\in\Delta^{M-1}$.
Dataset and donor identifiers define only the aggregation groups, while cell-type, disease, and tissue labels are not used.
The fixed centroids allow the same construction to be applied to held-out groups without fitting group-specific parameters.
Further details on pseudo-bulk construction, soft assignment, and the corresponding settings are provided in Appendix~\ref{app:pseudobulk-details}, Algorithm~\ref{alg:overall-framework}, and Appendix~\ref{app:model-architecture-hparams}, respectively.

Two linear projections learned during pretraining map $\mathbf{t}_b$ to the mean and log-variance of a diagonal Gaussian prior shared across cells in group $b$ and their $K$ latent tokens:
\vspace{-1ex}
\begin{equation}
    \begin{aligned}
    p_{\eta}(\mathbf{Z}_i \mid \mathbf{t}_{b_i})
    &=
    \prod_{k=1}^{K}
    \mathcal{N}\left(
        \mathbf{z}_{ik};
        \boldsymbol{\mu}_{\eta}(\mathbf{t}_{b_i}),
        \mathrm{diag}\!\left(\boldsymbol{\sigma}^{2}_{\eta}(\mathbf{t}_{b_i})\right)
    \right), \\
    \mathcal{L}_{\mathrm{KL},i}^{\mathrm{PB}}
    &=
    D_{\mathrm{KL}}\!\left[
        q_{\phi}(\mathbf{Z}_i\mid\mathbf{x}_i,S_i)
        \,\Vert\,
        p_{\eta}(\mathbf{Z}_i\mid\mathbf{t}_{b_i})
    \right].
    \end{aligned}
    \label{eq:method-pbcond}
\end{equation}
The group code changes only the KL reference and is not provided to the encoder, evaluated readout, decoder queries, or decoder memory.
At zero-shot evaluation, embeddings and reconstructions follow the posterior-to-decoder path without the held-out group code.
The code is used only for conditional-prior diagnostics.
Giving the same pseudo-bulk code to a matched scVI as a conditional prior trades identity and cross-donor comparability for local donor mixing in Appendix~\ref{app:pbcond-controls}, so the balance comes from the routing design rather than from the code itself.
The contribution of PB-Cond to the full model is evaluated by the ablation in Figure~\ref{fig:component-ablation}, and its conditioning mechanism is examined in Appendix~\ref{app:pbcond-analysis}.

\section{Experiments}
\label{sec:experiments}
We organize our experiments around four research questions covering the overall zero-shot embedding benchmark and the three representation demands evaluated within matched biological populations:
\begin{itemize}[leftmargin=2em,itemsep=0pt,topsep=1pt]
    \item \textbf{RQ1. Overall zero-shot benchmark:}
    How does \proposed{} compare with existing methods on biological identity and context invariance across held-out datasets?

    \item \textbf{RQ2. Biological identity and state structure:}
    Does the evaluated cell embedding preserve cell-type identity and disease-associated state across donors?

    \item \textbf{RQ3. Context invariance:}
    Does the evaluated cell embedding limit donor- and modality-associated variation among biologically matched cells?

    \item \textbf{RQ4. Expression fidelity:}
    Does predicted expression preserve disease-associated gene-level changes and their pathway organization in the same held-out cohorts?

\end{itemize}
We then attribute the representation profile to the routing components and test which tensions among the demands remain.

\subsection{Experimental setup}

We evaluate \proposed{} under a release-based zero-shot protocol.
\proposed{} is trained on 62.6 million cells from 478 datasets in the January 2025 CZ CELLxGENE Census release~\citep{czi2025cz} and evaluated on 30.4 million cells from 89 datasets first appearing in the November 2025 release.
We compare \proposed{} with scVI, Geneformer, scGPT, CellPLM, and scPRINT.

RQ1 uses all 89 held-out datasets to assess broad zero-shot generalization, and the ablation and interventions of Section~\ref{sec:ablation} evaluate the reduced and intervened models on the same datasets with the same protocol.
RQ2--RQ4 use the same donor-balanced cells from five cohorts with normal and disease/state groups spanning kidney, liver, brain, tendon, and colon.
Each comparison fixes assay and tissue, includes multiple donors in both states, and contains at least three eligible cell types.
This common population allows the three representation demands to be evaluated jointly within one model.
Across the uncontrolled held-out datasets we score each demand with population-level embedding and reconstruction metrics, and within the controlled common cohorts with metrics defined by biological labels, so that each setting is scored at the level it supports.

No single common cohort can fully evaluate every demand.
Cell-type and disease-state labels do not capture all expression-derived biological variation, donor labels cannot distinguish nuisance effects from genuine individual biology, 
and the common cohorts do not directly test measurement modality or expression fidelity across broader diseases and datasets.
We therefore add RQ-specific held-out analyses.
Broader lineage and disease cohorts complement RQ2, biologically matched scRNA-seq and snRNA-seq populations complement RQ3, and additional disease cohorts and the full held-out benchmark complement RQ4.
All cohorts and contrasts are fixed using metadata, gene availability, and cell-count support before inspecting embeddings, reconstructions, or model performance.

\begin{table*}[t]
    \centering
    \caption{
        Release-based zero-shot embedding benchmark. 
        Values are 20-repeat means (SD) across datasets. 
        Bold and italics mark the best and second-best values; 
        stars mark gains over the strongest baseline by paired Wilcoxon test ($^{\ast}$:$p<$0.05, $^{\ast\ast\ast}$:$p<$0.001). 
        \textsuperscript{$\dagger$}Metadata-supervised pretraining.
    }
    \label{tab:main_zsb}
    \newcommand{\tabmethod}[1]{\begin{tabular}[c]{@{}l@{}}#1\end{tabular}}
    \newcommand{\tabmeanstd}[2]{\begin{tabular}[c]{@{}c@{}}$#1$\\[-2pt]{\scriptsize $(#2)$}\end{tabular}}
    \resizebox{0.99\linewidth}{!}{%
    \begin{tabular}{lcccccccc}
    \toprule
    & \multicolumn{5}{c}{\textbf{\textit{Biological identity}}} & \multicolumn{3}{c}{\textbf{\textit{Context invariance}}} \\
    \cmidrule(lr){2-6}\cmidrule(lr){7-9}
    \textbf{Method} & \textbf{NMI} & \textbf{ARI} & \textbf{ASW} & \textbf{cLISI} & \textbf{Iso. Label} & \textbf{BRAS} & \textbf{iLISI} & \textbf{PCR} \\
    \midrule
    \tabmethod{scVI} & \tabmeanstd{0.640}{0.138} & \tabmeanstd{\mathit{0.476}}{0.155} & \tabmeanstd{\mathit{0.570}}{0.049} & \tabmeanstd{0.973}{0.050} & \tabmeanstd{\mathit{0.598}}{0.072} & \tabmeanstd{0.730}{0.093} & \tabmeanstd{\mathit{0.218}}{0.174} & \tabmeanstd{0.195}{0.135} \\
    \tabmethod{Geneformer} & \tabmeanstd{0.558}{0.166} & \tabmeanstd{0.377}{0.156} & \tabmeanstd{0.545}{0.037} & \tabmeanstd{\mathbf{0.980}}{0.046} & \tabmeanstd{0.567}{0.062} & \tabmeanstd{0.732}{0.106} & \tabmeanstd{0.167}{0.169} & \tabmeanstd{0.112}{0.115} \\
    \tabmethod{scGPT} & \tabmeanstd{\mathit{0.645}}{0.141} & \tabmeanstd{0.470}{0.158} & \tabmeanstd{\mathbf{0.575}}{0.052} & \tabmeanstd{0.977}{0.047} & \tabmeanstd{\mathbf{0.600}}{0.081} & \tabmeanstd{0.728}{0.091} & \tabmeanstd{0.196}{0.173} & \tabmeanstd{0.215}{0.107} \\
    \tabmethod{CellPLM} & \tabmeanstd{0.595}{0.142} & \tabmeanstd{0.404}{0.150} & \tabmeanstd{0.559}{0.077} & \tabmeanstd{0.970}{0.049} & \tabmeanstd{0.598}{0.127} & \tabmeanstd{0.677}{0.087} & \tabmeanstd{\mathbf{0.236}}{0.178} & \tabmeanstd{\mathit{0.284}}{0.175} \\
    \tabmethod{scPRINT\textsuperscript{$\dagger$}} & \tabmeanstd{0.405}{0.154} & \tabmeanstd{0.240}{0.139} & \tabmeanstd{0.510}{0.017} & \tabmeanstd{0.914}{0.073} & \tabmeanstd{0.523}{0.035} & \tabmeanstd{\mathit{0.748}}{0.105} & \tabmeanstd{0.138}{0.135} & \tabmeanstd{0.025}{0.063} \\
    \midrule
    \rowcolor{methodhighlight}\tabmethod{\proposed{}} & \tabmeanstd{\mathbf{0.651}^{*}}{0.149} & \tabmeanstd{\mathbf{0.484}^{*}}{0.152} & \tabmeanstd{0.540}{0.032} & \tabmeanstd{\mathit{0.978}}{0.043} & \tabmeanstd{0.556}{0.049} & \tabmeanstd{\mathbf{0.840}^{***}}{0.073} & \tabmeanstd{0.181}{0.165} & \tabmeanstd{\mathbf{0.378}^{***}}{0.104} \\
    \bottomrule
    \end{tabular}
    }
\end{table*}

Dataset releases are described in Appendix~\ref{app:datasets}, preprocessing in Appendix~\ref{app:preprocessing}, baseline checkpoints and inference in Appendix~\ref{app:baseline-section}, and the common evaluation protocol in Appendix~\ref{app:evaluation-protocol}.
Metric definitions are provided in Appendix~\ref{app:evaluation-metrics}, common-cohort definitions and sampling procedures in Appendix~\ref{app:joint-cohort-selection}, and RQ-specific results in Appendix~\ref{app:supplementary-results}.

\subsection{Experimental results}

\textbf{RQ1: Overall zero-shot benchmark.}
RQ1 evaluates the frozen cell embedding with standard biological-conservation and batch-correction metrics, which Table~\ref{tab:main_zsb} groups by the demand they measure.
NMI and ARI measure global cell-type clustering.
ASW, cLISI, and isolated-label silhouette assess complementary aspects of cell-type separation.
BRAS and PCR comparison assess overall cross-donor comparability and donor-associated variance, while iLISI assesses local donor mixing.
In Table~\ref{tab:main_zsb}, \proposed{} leads NMI and ARI together with BRAS and PCR comparison.
scGPT is closest on NMI and leads ASW and isolated-label silhouette, scPRINT is closest on BRAS, and CellPLM leads iLISI, so no baseline is strong on both cell-type structure and cross-donor comparability at once.
A single frozen embedding thus combines strong global cell-type structure with strong overall cross-donor comparability.

To determine where this advantage arises, we relate per-dataset performance gaps to cell-type composition and donor-associated variation before integration in Figure~\ref{fig:rq1-dataset-characteristics}.
The NMI and ARI gaps increase with cell-type count and the fraction of cells assigned to rare cell types, whereas the isolated-label gap remains flat.
Thus, the global clustering advantage becomes stronger as cell-type composition grows more complex but does not extend uniformly to isolated cell types.
Because cLISI and iLISI evaluate fixed local neighborhoods and are sensitive to label cardinality and imbalance, we interpret cLISI with NMI and ARI, and iLISI with BRAS and PCR comparison.
BRAS decreases as donor-associated variation before integration increases, but less sharply for \proposed{}.
Its BRAS advantage therefore grows as cross-donor comparison becomes more difficult rather than being limited to datasets with little donor-associated variation.
The identity advantage and the comparability advantage thus grow with different dataset properties, so the two demands are not traded against each other within the same embedding.

RQ2--RQ4 subsequently evaluate the three representation demands in biologically controlled populations.

\begin{table*}[t]
    \centering
    \caption{
        Matched evaluation of the three representation demands on the common cohorts.
        Values are equal-cohort means; bold and italics mark the best and second-best values.
        Expression metrics are reported only for models with a full expression output ($^{\dagger}$: no decoder weights in the checkpoint; $^{\ddagger}$: metadata-supervised pretraining).
    }
    \label{tab:joint-trilemma}


    \resizebox{0.99\linewidth}{!}{%
    \begin{tabular}{@{}lccccc>{\columncolor{methodhighlight}}c@{}}
        \toprule
        \textbf{Metric} &
        \textbf{scVI} &
        \textbf{Geneformer}$^{\dagger}$ &
        \textbf{scGPT}$^{\dagger}$ &
        \textbf{CellPLM} &
        \textbf{scPRINT}$^{\ddagger}$ &
        \textbf{\proposed{}} \\
        \midrule

        \multicolumn{6}{l}{\textbf{\textit{Biological identity and state}}}
        & \cellcolor{methodhighlight} \\

        \hspace{1.5em} Cell-type BA $-$ chance
        & $0.6535$
        & $0.6112$
        & $\mathit{0.6667}$
        & $0.6041$
        & $0.4475$
        & $\mathbf{0.6697}$ \\

        \hspace{1.5em} Neighbor purity
        & $\mathit{0.6817}$
        & $0.6006$
        & $\mathbf{0.7041}$
        & $0.6459$
        & $0.4327$
        & $0.6795$ \\

        \hspace{1.5em} Disease-state BA
        & $0.5583$
        & $\mathit{0.5785}$
        & $0.5744$
        & $0.5507$
        & $0.5432$
        & $\mathbf{0.5856}$ \\ [2pt]

        \multicolumn{6}{l}{\textbf{\textit{Context invariance}}}
        & \cellcolor{methodhighlight} \\

        \hspace{1.5em} Residual donor invariance
        & $0.8889$
        & $0.8864$
        & $0.8778$
        & $\mathit{0.9161}$
        & $0.9016$
        & $\mathbf{0.9336}$ \\

        \hspace{1.5em} Conditional donor BRAS
        & $0.7113$
        & $0.7240$
        & $0.7078$
        & $0.6194$
        & $\mathit{0.7835}$
        & $\mathbf{0.8435}$ \\

        \hspace{1.5em} Local donor mixing
        & $\mathit{0.8216}$
        & $0.7879$
        & $0.7982$
        & $\mathbf{0.8653}$
        & $0.7582$
        & $0.7855$ \\ [2pt]

        \multicolumn{6}{l}{\textbf{\textit{Expression fidelity}}}
        & \cellcolor{methodhighlight} \\

        \hspace{1.5em} logFC Spearman
        & $0.2128$
        & --
        & --
        & $0.2201$
        & $\mathbf{0.4214}$
        & $\mathit{0.3392}$ \\

        \hspace{1.5em} DEG Jaccard
        & $0.1138$
        & --
        & --
        & $0.1670$
        & $\mathit{0.2202}$
        & $\mathbf{0.2688}$ \\

        \hspace{1.5em} DEG sign concordance
        & $0.7160$
        & --
        & --
        & $0.7051$
        & $\mathit{0.7423}$
        & $\mathbf{0.7861}$ \\

        \hspace{1.5em} Pathway Jaccard
        & $0.2525$
        & --
        & --
        & $0.2539$
        & $\mathit{0.2581}$
        & $\mathbf{0.3451}$ \\

        \hspace{1.5em} Pathway-score $\rho$
        & $-0.1108$
        & --
        & --
        & $\mathit{-0.0378}$
        & $-0.0824$
        & $\mathbf{0.2288}$ \\
        \bottomrule
    \end{tabular}
    }
\end{table*}

\textbf{RQ2: Biological identity and state structure.}
RQ2 tests whether cell type and disease state shape cellular similarity across donors, as required by the biological-identity demand.
Donor-associated variation may contain genuine biology, but same-donor neighbors can obscure whether this structure extends across individuals.
We therefore restrict candidate neighbors to other donors.
Chance-corrected cell-type balanced accuracy within disease state, disease-state balanced accuracy within cell type, and cell-type neighbor purity assess the corresponding biological structure across donors and in local
neighborhoods.
In Table~\ref{tab:joint-trilemma}, \proposed{} leads both balanced-accuracy metrics while remaining competitive in cell-type neighbor purity.
scGPT leads neighbor purity but trails on disease-state accuracy and on every context metric.
These results show that cell type and disease state shape cross-donor similarity while local cell-type structure is retained.

Beyond the common cohorts, in normal blood \proposed{} places in the top two for all nine marker programs and has the highest mean, showing that lineage-program similarity is consistently reflected in cross-donor neighborhoods.
Across four non-blood cohorts, it ranks first in most cohort--metric comparisons and never below third, with the highest equal-cohort averages for all three retrieval metrics, so disease-state structure extends across tissues and diseases rather than being driven by one cohort.

Together, the common-cohort and RQ2-specific analyses support the biological-identity demand across cell-type identity, lineage-associated expression programs, and disease-associated state.

\textbf{RQ3: Context invariance.}
RQ3 tests whether the cell embedding remains robust to nuisance context among biologically matched cells.
Because donor-associated variation can contain both individual biology and technical effects, context invariance does not require all donor information to be removed.
Using the same cells as RQ2, we compare cells sharing cell type and disease state and ask how much of their remaining embedding variation is associated with donor.
We measure the fraction of variation not explained by donor, cross-donor comparability using within-group BRAS, and donor composition among nearest neighbors.
In Table~\ref{tab:joint-trilemma}, \proposed{} leads the first two metrics.
This shows that donor accounts for less variation among biologically matched cells while cross-donor comparability remains high, complementing the biological structure established in RQ2.
Nearest-neighbor donor mixing provides a local view, but high mixing can reflect either reduced donor effects or mixing of unannotated biological states.
CellPLM leads local donor mixing yet has the lowest conditional donor BRAS, consistent with mixing that removes structure rather than aligning it.
We therefore interpret it with the two overall measures.

The RQ3-specific analysis turns to measurement modality, a more direct technical context, on matched scRNA-seq and snRNA-seq populations.
\proposed{} shows the least modality-associated variation under both a variance measure (modality $R^2$) and a distance measure (modality ASW), and this holds in nearly every matched donor--cell-type group rather than only on average.
Modality iLISI provides the corresponding nearest-neighbor view and yields a different method ordering.
Because no separate correction is applied, these results reflect the evaluated embedding itself.

Together, the donor and modality analyses support context invariance across distinct contexts without requiring all context information to be removed.

\begin{figure}[t]
    \centering
    \includegraphics[width=0.99\linewidth]{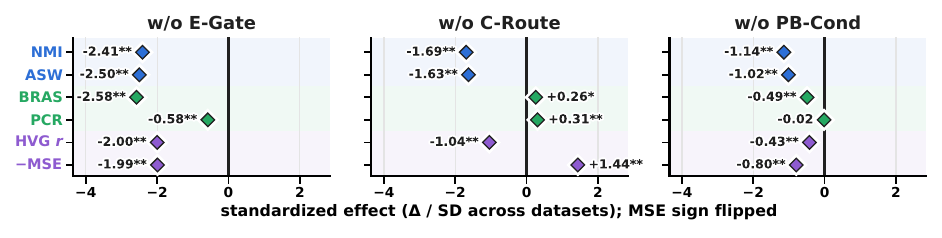}
    \vspace{-3ex}
    \caption{
    Component ablation of the full model.
    Diamonds show the mean score change without each component (w/o) relative to the full model, standardized by its across-dataset standard deviation.
    Negative values indicate degradation, MSE is sign-flipped, and $r$ denotes Pearson correlation.
    For C-Route, the mean-pooled readout is retained and only decoder-query routing is removed.
    Asterisks indicate significance after BH-FDR correction of paired two-sided Wilcoxon tests ($^{*}q<0.05$; $^{**}q<0.01$).
    }
    \label{fig:component-ablation}
\end{figure}

\textbf{RQ4: Expression fidelity.}
RQ4 evaluates whether disease-associated gene-level variation remains available in predicted expression, which for \proposed{} is the reconstruction.
Within each cell type, we compare disease-associated changes from observed and reconstructed expression at complementary levels.
LogFC Spearman assesses gene ranking, DEG overlap and sign concordance assess major gene-level changes, and pathway overlap and pathway-score Spearman assess their pathway organization.

All reconstructions are scored on one shared gene list per dataset, defined in Appendix~\ref{app:supp-expression-setup}, so that no model's own gene selection defines the comparison.
In Table~\ref{tab:joint-trilemma}, \proposed{} leads DEG overlap, sign concordance, and both pathway metrics, whereas logFC Spearman, which scores the ordering of effect sizes across all shared genes, yields a different ordering.
scPRINT leads logFC Spearman and is second on three of the four remaining fidelity metrics while trailing every identity metric, and Geneformer and scGPT provide no expression output for this comparison.
\proposed{} thus most consistently recovers which genes change, in which direction, and how these changes organize into pathways.
Across seven additional disease cohorts from the held-out datasets, \proposed{} leads DEG overlap in every cohort and the full pattern holds in most of them.
Because disease contrasts do not describe general reconstruction behavior, we add a secondary comparison of observed and reconstructed expression across all held-out datasets for the four models with reconstruction outputs.
\proposed{} has the highest Pearson and Spearman correlations, with the largest margin on genes detected in at least 20\% of cells, whereas MAE and MSE yield a different method ordering.
These contrasting orderings distinguish preservation of linear and rank relationships from exact pointwise agreement.
Because observed profiles contain sampling noise and technical variation, lower pointwise error need not indicate stronger preservation of biological variation~\citep{brennecke2013accounting}.
We therefore interpret this analysis as complementary to the disease-contrast and pathway results.

Across the common cohorts, additional disease cohorts, and full held-out benchmark, reconstructed expression retains disease-associated gene and pathway variation and the relative structure of expression, supporting the expression-fidelity demand.
The model that leads context invariance in Table~\ref{tab:joint-trilemma} also leads gene- and pathway-level fidelity, so reducing donor-associated variation in the embedding did not cost the reconstruction the structure that the fidelity demand measures.

\textbf{Representation trilemma.}
RQ2--RQ4 evaluate the three representation demands on the same sampled cells within one model.
Every baseline that leads a metric in Table~\ref{tab:joint-trilemma} does so within one demand at the cost of the others, whereas \proposed{} leads all three at once.
The RQ-specific analyses extend this joint evidence across lineage programs, diseases, measurement modalities, and the full held-out benchmark, supporting the information-routing balance targeted by the representation trilemma.

\begin{figure}[t]
    \centering
    \includegraphics[width=0.99\linewidth]{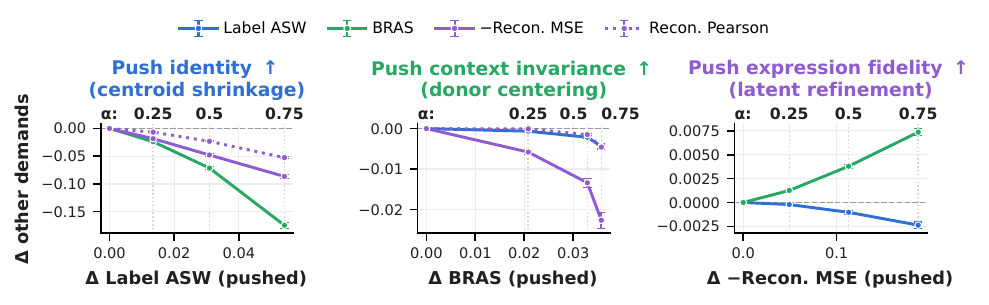}
    \vspace{-2ex}
    \caption{
    Demand-targeted interventions on the trained model.
    Each panel pushes one demand on the posterior-mean latent at inference (centroid shrinkage, donor centering, or latent refinement at strength $\alpha=0.25$, $0.5$, $0.75$) and plots the change of the other two demands against the change of the pushed metric; fidelity is shown as reconstruction Pearson and as MSE with its sign reversed.
    Points are dataset means with SEM.
    }
    \label{fig:demand-interventions}
\end{figure}

\subsection{Model analysis}
\label{sec:ablation}

\textbf{Component ablation.}
The ablation of Figure~\ref{fig:component-ablation} attributes the representation profile to the routing components by evaluating the full model without each one.
E-Gate is the precondition on which the other routes operate, since without it identity, context invariance, and expression fidelity fail together and nearly uniformly across datasets.
C-Route is what makes the evaluated embedding carry cell-level biology, since without it identity and correlation-based fidelity fall while cross-donor comparability improves slightly, the trade-off expected when a route strengthens biology in the same vector that must stay context-invariant.
The same removal lowers pointwise reconstruction error, so decoder-query routing spends pointwise agreement to preserve the gene-level structure that the fidelity demand measures.
PB-Cond refines how posteriors are organized rather than being the route that keeps donor-associated variation out of the embedding, since without it the donor-variance measure is unchanged and cross-donor comparability declines only moderately, whereas identity degrades most and both reconstruction measures degrade consistently across datasets.
Switching PB-Cond on within the three routing configurations of Appendix~\ref{app:pbcond-controls} changes gene-level reconstruction in every configuration but improves identity appreciably only once both routes are present, so its effect depends on the surrounding routes rather than being a fixed additive gain.
The PB-Cond effect is smaller than the other two because E-Gate and C-Route act on the reconstruction path itself and receive the full reconstruction gradient, whereas PB-Cond reaches the posterior only through the KL term, which is weighted by $\lambda_{\mathrm{KL}}$ during training and is not used at zero-shot evaluation.
Its contribution is therefore bounded by construction to how the prior organizes posteriors during training, and the shuffled-code control of Appendix~\ref{app:pbcond-stress} separates this contribution into a representation effect that is largely independent of the donor--code assignment and a reconstruction effect that requires matched codes.
PB-Cond controls and stress tests and hyperparameter sensitivity are reported in Appendix~\ref{app:model-analysis-section}, and the mechanism of each route is examined in Appendix~\ref{app:mechanistic-section}.

\textbf{Demand-targeted interventions.}
The interventions of Figure~\ref{fig:demand-interventions} ask which tensions remain once the routes are in place, by intervening on the posterior-mean latent of the trained model at inference and measuring all three demands on the same intervened latent, the pooled embedding for identity and invariance and the unchanged decoder for fidelity.
Each intervention is designed to raise one demand by construction and nothing else, so that any movement of the other two can only come from coupling in the representation, unlike a training hyperparameter or a removed component, which alter all three demands at once.
Identity is pushed by shrinking each cell toward its k-means centroid in the embedding, context invariance by subtracting a fraction of the donor mean offset, and expression fidelity by refining each latent toward the reconstruction of its own observed profile, with the strength $\alpha$ interpolating between the unmodified latent and the full intervention.
Pushing identity lowers both context invariance and expression fidelity in nearly every dataset, and pushing fidelity lowers identity only slightly in turn, so the identity--fidelity edge of the trilemma remains coupled, most strongly from identity to fidelity, and both demands depend on the same route, C-Route, whose removal lowers them together in the ablation.
Pushing context invariance leaves identity and correlation-based fidelity nearly unchanged and costs only pointwise reconstruction error, so the donor offset that remains in the embedding carries context-specific expression levels rather than the gene-level structure that the fidelity demand measures.
The tension is therefore a property of the shared latent rather than of any one design: the two demands of the remaining edge draw on the same posterior tokens, whereas context can be moved out of the evaluated vector at almost no cost.
The full metric set and per-dataset consistency for every intervention are reported in Appendix~\ref{app:demand-interventions}.

\section{Conclusion}
\label{sec:conclusion}

We studied label-free scRNA-seq representation learning as a representation trilemma among biological identity, context invariance, and expression fidelity, and recast this tension as an information-routing problem rather than as separate questions of clustering, batch correction, and reconstruction. Across the zero-shot benchmark, matched-cohort analyses, ablations, and demand-targeted interventions, scTrilemma improves this balance within a single model without target annotations or auxiliary representation losses: expression-derived routing through a latent bottleneck removes nearly all of the cost of context invariance and leaves identity against fidelity as the residual trade-off. The ablation and mechanistic analyses give each route a distinct role: E-Gate is the precondition on which the other routes operate, C-Route makes the evaluated embedding carry cell-level biology at a small cost in comparability, and PB-Cond raises identity and fidelity through a prior that is aligned with the posterior.
Limitations and stress-test directions are discussed in Appendix~\ref{app:limitations-section}.

\begin{ack}
This work was supported by the Institute of Information \& Communications Technology Planning \& Evaluation(IITP) grant funded by the Korea government(MSIT) (RS-2025-02304967, AI Star Fellowship(KAIST)), the National Research Foundation of Korea(NRF) grant funded by the Korea government(MSIT) (RS-2024-00335098), the National Research Foundation of Korea(NRF) funded by Ministry of Science and ICT (RS-2022-NR068758), and Institute for Information \& communications Technology Planning \& Evaluation(IITP) grant funded by the Korea government(MSIT) (RS-2019-II190075, Artificial Intelligence Graduate School Program(KAIST)).

\end{ack}

\bibliographystyle{abbrvnat}
\bibliography{neurips_2026}

\clearpage

\rule[0pt]{\linewidth}{3pt}
\begin{center}
    \Large 
    \textit{Supplementary Material for} \\ 
    {
        \bfseries 
        \textit{scTrilemma: Balancing Identity, Invariance, and Fidelity in Single-Cell Representation Learning} 
    } \par 
\end{center}
\vspace{2ex}
\DoToC

\clearpage

\appendix

\section{Experimental Setup}
\label{app:experimental-setup}
\subsection{Common Experimental Setup}
\label{app:common-experimental-setup}

\subsubsection{Datasets}
\label{app:datasets}

We use the CZ CELLxGENE Discover Census~\citep{czi2025cz}, a curated and standardized corpus of publicly contributed single-cell transcriptomic datasets.
Our pretraining corpus is built from the long-term-supported 2025-01-30 Census release, retaining \textit{Homo sapiens} observations marked as primary data and with non-missing suspension type.
The resulting pretraining corpus contains 62.6M cells from 478 datasets, spanning 29 assays, 794 annotated cell types, and 67 tissue-general categories.
For release-based zero-shot evaluation, we identify datasets first appearing in the subsequent 2025-11-08 Census release.

\subsubsection{Preprocessing}
\label{app:preprocessing}

We retain cells from both Census releases regardless of disease annotation.
Cell-type annotations are reserved for dataset characterization and evaluation.
We retain raw counts as reconstruction targets and construct encoder inputs by per-cell library-size normalization to $10^4$ counts followed by log transformation.
For memory-efficient training, each cell is represented by a stochastic crop of 4,096 gene positions, with gene identifier matching and crop construction details provided in Appendix~\ref{app:preprocessing-details}.
Dataset--donor pseudo-bulk profiles are computed from normalized expression without cell-type, disease, or tissue annotations.
k-means centroids are fitted on profiles from the 2025-01-30 pretraining corpus, and each group is softly assigned to them to obtain the expression-derived prior code used by PB-Cond.
Further details are provided for pseudo-bulk construction in Appendix~\ref{app:pseudobulk-details} and soft assignment in Algorithm~\ref{alg:overall-framework}.
Model architecture, hyperparameter, and computing-environment details are provided in Appendix~\ref{app:model-architecture-hparams} and Appendix~\ref{app:computing-environment}.

\subsubsection{Baselines}
\label{app:baseline-section}

We select baselines that expose frozen cell embeddings from count-vector, gene-token, or cell-token encoders for evaluation under the same zero-shot protocol.
We compare \proposed{} against scVI~\citep{lopez2018deep} and four pretrained single-cell representation models: Geneformer~\citep{theodoris2023transfer}, scGPT~\citep{cui2024scgpt}, CellPLM~\citep{wen2024cellplm}, and scPRINT~\citep{kalfon2025scprint}.
Together, these methods cover a count-vector VAE with batch-conditioned decoding, rank- and value-based gene-token encoders, tissue-level cell-token modeling, and metadata-supervised pretraining.
The sources, checkpoints, preprocessing procedures, and inference paths used for evaluation are specified below.

\providecommand{\ctxyes}{\textcolor{green!60!black}{\ensuremath{\checkmark}}}
\providecommand{\ctxno}{\textcolor{gray}{--}}
\providecommand{\ctxpartial}[1]{\textcolor{orange}{$\triangle$}~(#1)}


\begingroup
\raggedright
\begin{itemize}[leftmargin=*, itemsep=0.4em, topsep=0.2em]
    \item \textbf{scVI}~\citep{lopez2018deep}
    \begin{itemize}[leftmargin=1.5em, itemsep=0.1em, topsep=0.1em]
        \item Source: \url{https://github.com/scverse/scvi-tools}.
        \item Checkpoint: \url{s3://cellxgene-contrib-public/models/scvi/2025-01-30/homo_sapiens/model.pt}.
        \item Inference and preprocessing: The wrapper aligns target counts to the reference gene ordering and extracts posterior-mean latent embeddings through a frozen direct-module path verified against the released scVI query API. Unseen target batches follow the same mapping used by the query path.
    \end{itemize}


    \item \textbf{Geneformer}~\citep{theodoris2023transfer}
    \begin{itemize}[leftmargin=1.5em, itemsep=0.1em, topsep=0.1em]
        \item Source: \url{https://huggingface.co/ctheodoris/Geneformer}.
        \item Checkpoint: the root checkpoint of the HuggingFace \path{ctheodoris/Geneformer} release, selected with \path{--geneformer-variant main}.
        \item Inference and preprocessing: Inputs are converted to rank-based gene sequences with the variant-specific token dictionary and a maximum input length of 4,096, then mean-pooled over non-special-token encoder states from the frozen model.
    \end{itemize}

    \item \textbf{scGPT}~\citep{cui2024scgpt}
    \begin{itemize}[leftmargin=1.5em, itemsep=0.1em, topsep=0.1em]
        \item Source: \url{https://github.com/bowang-lab/scGPT}.
        \item Checkpoint: the official model-zoo \texttt{whole-human} checkpoint at \url{https://drive.google.com/drive/folders/1oWh_-ZRdhtoGQ2Fw24HP41FgLoomVo-y}.
        \item Inference and preprocessing: Embeddings are exported through the official scGPT inference path in a separate environment and then evaluated by the shared benchmark runner. The exporter uses gene-token inputs with the official vocabulary and a maximum sequence length of 1,200.
    \end{itemize}

    \item \textbf{CellPLM}~\citep{wen2024cellplm}
    \begin{itemize}[leftmargin=1.5em, itemsep=0.1em, topsep=0.1em]
        \item Source: \url{https://github.com/OmicsML/CellPLM}.
        \item Checkpoint: \path{20230926_85M.best.ckpt}, the release designated as the official pretrained model in the repository README.
        \item Inference and preprocessing: Input AnnData objects use the standardized Ensembl gene IDs supplied by CELLxGENE, with automatic Ensembl conversion disabled. The official pipeline generates per-dataset embeddings, which are cached and aligned to the shared benchmark by \path{soma_joinid}.
    \end{itemize}


    \item \textbf{scPRINT}~\citep{kalfon2025scprint}
    \begin{itemize}[leftmargin=1.5em, itemsep=0.1em, topsep=0.1em]
        \item Source: \url{https://github.com/cantinilab/scPRINT} and \url{https://huggingface.co/jkobject/scPRINT}.
        \item Checkpoint: the HuggingFace \path{large-v1.ckpt} model used to generate the transferred manuscript cache ($d_{\mathrm{model}}=512$, 16 transformer layers, 4 attention heads, 101.1M state-dict parameters, and a 1.24~GB checkpoint file).
        \item Inference and preprocessing: The generated embeddings are cached per dataset and aligned to the benchmark AnnData object by \path{soma_joinid}. The shared benchmark runner evaluates these fixed embeddings without target-label supervision or target-specific fine-tuning.
        \item Execution environment: scPRINT embedding extraction and reconstruction were executed on a separate Linux GPU node because the official scPRINT dependency pins, including \texttt{torch==2.2.0} and \texttt{lamindb==0.76.12}, are incompatible with the main training stack. The node had 8 NVIDIA L40S GPUs (sm\_89), each reporting 46,068 MiB of memory, with NVIDIA driver 550.107.02. The isolated \texttt{uv} environment used Python 3.11.13 on Linux 5.14.0 (RHEL 9.3); PyTorch reported CUDA 12.1 and cuDNN 8902. scPRINT was installed from the official v1.6.4 git tag with manual dependency resolution, and \texttt{flash-attn==2.5.6} was used for transformer attention.
    \end{itemize}
\end{itemize}
\endgroup


\subsubsection{Evaluation Protocol and Scope}
\label{app:evaluation-protocol}

We evaluate \proposed{} and all baselines under a release-based zero-shot protocol on the held-out benchmark.
The final \proposed{} checkpoint is determined by a fixed training-step budget, while baselines use publicly released checkpoints or fixed precomputed embeddings.
All \proposed{} parameters and PB-Cond centroids remain fixed during evaluation, and zero-shot baselines receive no target-specific parameter updates.
Target annotations are reserved for metric computation, and checkpoint selection uses neither these annotations nor benchmark performance.

For \proposed{}, reported embeddings and reconstructions are computed through the posterior and decoder pathways without using held-out group codes.
Held-out pseudo-bulk profiles are mapped to the fixed PB-Cond centroids only when conditional-prior quantities are evaluated.
Representation and downstream metrics are reported for methods that expose frozen cell embeddings, and reconstruction metrics only for methods with a comparable decoder output.
We distinguish off-the-shelf zero-shot models from supervised-pretrained zero-shot models, whose checkpoints use label or ontology supervision before target-dataset evaluation.

The frozen cell embedding is used to assess biological identity and state as well as context invariance, whereas reconstructed expression is used to assess expression fidelity.
Cohort construction and sampling are described in Appendix~\ref{app:evaluation-cohorts}, and metric definitions and normalization conventions are provided in Appendix~\ref{app:evaluation-metrics}.

\subsection{Evaluation Cohorts and Sampling}
\label{app:evaluation-cohorts}

\subsubsection{Release-Based Zero-Shot Benchmark (RQ1)}
\label{app:rq1-experimental-setup}

The candidate pool comprises all 92 datasets first appearing in the 2025-11-08 Census release.
Two datasets are excluded because they contain only one annotated cell type and therefore cannot support multi-class biological-conservation metrics.
We additionally exclude one 4.1M-cell dataset because complete embeddings could not be obtained for all evaluated methods under the common inference protocol.
The corresponding Census dataset-ID prefixes are \path{d973ea15} and \path{ed880090} for the single-cell-type datasets, and \path{37a17b78} for the 4.1M-cell dataset.
These exclusions were determined by metric eligibility and output availability under the common protocol, without inspecting model performance.
All remaining 89 datasets are included, forming a held-out zero-shot benchmark of 30.4M cells across 18 assays, 453 annotated cell types, and 33 tissue-general categories.

The benchmark datasets contain 2 to 81 annotated cell types and span broad variation in cell-type composition and donor-associated variation before integration.
Figure~\ref{fig:rq1-dataset-characteristics} examines how these characteristics relate to per-dataset performance differences.

A deterministic uniform sample caps each dataset at 100,000 cells before metric-specific sampling.
Each metric sample contains at most 10,000 cells, with up to 500 cells per annotated cell type while retaining cells from rare types.
Repeat and aggregation rules for each metric follow Appendix~\ref{app:rq1-evaluation-metrics}.
Donor labels define the RQ1 context variable.
Batch-correction metrics requiring multiple donors are evaluated on the 75 eligible datasets, while all 89 datasets remain included in biological-conservation evaluation.

\subsubsection{Common Cohorts for the Three Representation Demands (RQ2--4)}
\label{app:joint-cohort-selection}
Every cohort below was selected from Census metadata and cell support before inspecting model outputs or performance, and all methods are evaluated on the same sampled cells.

To compare the three representation demands within the same biological populations, we screened all 89 RQ1 datasets for normal-versus-disease/state contrasts within fixed assay and tissue combinations.
Fixing assay and tissue limits differences in measurement protocol and anatomical source.
Cell support was evaluated in the fixed cross-method pool.

For datasets with multiple contrasts, one was selected before support filtering by total cells, shared cell types, and lexicographic state order, in that order.
If this contrast failed the filter, no alternative was considered.
A cell type required at least 40 cells in each state and at least two donors per state with ten or more cells per donor.
A dataset required at least three such cell types, supporting repeated sampling, cross-donor comparison, and biological-identity evaluation across multiple cell types.

The metadata screen identified 16 candidates, of which five passed the support filter; all five form the common panel summarized in Table~\ref{tab:joint-cohort-support} and evaluated in Table~\ref{tab:joint-trilemma}.
The common pool contains cells aligned across the compared embedding methods, and expression-fidelity metrics use the shared gene list defined in Appendix~\ref{app:supp-expression-setup}.
BD Rhapsody WTA denotes BD Rhapsody Whole Transcriptome Analysis.

\begin{table}[htbp]
    \centering
    \caption{Metadata-defined cohorts and evaluation support for Table~\ref{tab:joint-trilemma} (RQ2--4). $^\dagger$ Evaluation-support entries report eligible cell types, common-pool cells, cells per repeat, and common genes, in that order.}
    \label{tab:joint-cohort-support}
    \small
    \begin{tabular}{@{}llL{0.20\linewidth}L{0.23\linewidth}l@{}}
        \toprule
        \textbf{Cohort}
        & \shortstack{\textbf{Dataset ID}\\\textbf{prefix}}
        & \textbf{Fixed context}
        & \shortstack{\textbf{Disease/state}\\\textbf{(vs. normal)}}
        & \shortstack{\textbf{Evaluation}\\\textbf{support}$^\dagger$} \\
        \midrule
        Kidney
        & \texttt{867757c1}
        & 10x multiome, kidney
        & obstructive nephropathy
        & 13 / 2,243 / 520 / 1,524 \\

        Liver
        & \texttt{e3ed2ba4}
        & BD Rhapsody WTA, liver
        & metastatic colorectal carcinoma
        & 11 / 2,073 / 440 / 2,135 \\

        Brain
        & \texttt{203025fe}
        & 10x multiome, dorsolateral prefrontal cortex
        & cognitive disorder
        & 7 / 830 / 280 / 1,821 \\

        Tendon
        & \texttt{acd544d0}
        & 10x 3' v3, quadriceps femoris tendon
        & injury
        & 4 / 854 / 160 / 1,555 \\

        Colon
        & \texttt{829a3cd1}
        & 10x 3' v3, sigmoid colon
        & colorectal cancer
        & 6 / 959 / 240 / 1,639 \\
        \bottomrule
    \end{tabular}
\end{table}

We perform 20 sampling repeats using seeds 42 through 61.
For each eligible cell type, a repeat selects two donors per state and ten cells per donor, yielding 40 cells.
The same selected cells are used to evaluate biological identity and state, context invariance, and expression fidelity.
Results are aggregated within cohorts and then averaged equally across the five cohorts.

The common panel thus provides directly comparable evidence for all three demands.
RQ-specific panels extend biological identity with broader marker-program and disease-state variation, context invariance with matched measurement modality, and expression fidelity with broader disease contrasts and dataset coverage.
Each panel is screened independently before inspecting model outputs or performance, so a dataset shared with the common panel may contribute a different contrast under its analysis-specific criteria.

\subsubsection{Supplementary Biological-Identity and State Analyses (RQ2)}
\label{app:supp-biological-setup}

To extend evaluation of the biological-identity demand beyond the common-cohort labels, we define two RQ2-specific held-out panels.
The normal-blood analysis examines expression-derived lineage-program consistency across donors, whereas the non-blood analysis examines disease-state retrieval across donors and diseases.

\paragraph{Normal-blood lineage-program consistency.}
We screened all 89 held-out datasets using only Census metadata and marker-gene availability, enumerating normal-blood cohorts within each assay.
Eligibility required at least 8,000 cells, 20 donors, seven annotated cell types with at least 20 cells each, and all nine prespecified marker programs with at least two measured genes per program.
Applying these criteria to 20 assay-level candidates identified the 10x 5$'$ v2 cohort with dataset-ID prefix \path{c838aec3} as the sole eligible cohort.

The full cohort contains 1,265,624 cells from 625 donors, with 32 annotated cell types and nine measurable marker programs.
Using seed 42, we draw an annotated-cell-type-stratified evaluation subset of 8,000 cells from 623 donors and 32 annotated cell types.
All nine programs are reported separately, and their unweighted mean provides the macro summary.
Marker genes, CP10K normalization, and the cross-donor neighborhood metric are defined in Appendix~\ref{app:evaluation-metrics}.

\paragraph{Non-blood disease-state retrieval.}
Across all 89 held-out datasets, we enumerated normal-versus-disease contrasts within each fixed assay and anatomical tissue combination.
A cell type was eligible when each state contained at least 20 cells from at least two donors.
Within each dataset, one contrast was retained by ordering candidates by eligible cell-type support, followed by donor support and total cell count.

The metadata audit identified 16 dataset-level normal-contrast candidates.
Excluding the two blood candidates left 14 non-blood candidates.
The four candidates with the largest eligible cell-type support were retained, using donor support and total cell count as tie-breakers.
The four retained cohorts and their evaluation support are summarized in Table~\ref{tab:rq2-cohort-selection}.

\begin{table}[htbp]
    \centering
    \caption{Metadata-defined non-blood disease-state cohorts used in RQ2. Cohorts were fixed from metadata and cell-count support before model evaluation. Eligible cell types contain at least 20 cells from at least two donors in each state. $^\dagger$ Evaluation-support entries report donors, eligible cell types, and sampled cells, in that order.}
    \label{tab:rq2-cohort-selection}
    \small
    \begin{tabular}{@{}lp{0.27\linewidth}p{0.24\linewidth}l@{}}
        \toprule
        \textbf{Dataset ID prefix} &
        \textbf{Disease (vs. normal)} &
        \textbf{Fixed context} &
        \textbf{Evaluation support}$^\dagger$ \\
        \midrule
        \texttt{16023185}
        & Colon adenocarcinoma
        & 10x 3$'$ v2, colorectum
        & 43 / 38 / 12,000 \\
        \texttt{f14bc322}
        & Interstitial lung disease
        & 10x 5$'$ v2, lung
        & 93 / 33 / 12,000 \\
        \texttt{d68a8b48}
        & Bronchopulmonary dysplasia
        & 10x 3$'$ v3, middle lobe of right lung
        & 21 / 35 / 12,000 \\
        \texttt{19053a82}
        & Ulcerative colitis
        & 10x 3$'$ v2, colon
        & 11 / 40 / 12,000 \\
        \bottomrule
    \end{tabular}
\end{table}

Within each cohort, we use seed 42 to draw a 12,000-cell sample stratified by annotated cell type, disease state, and donor.
For each query cell, candidate neighbors are restricted to cells with the same annotated cell type from other donors.
Neighborhood metrics and equal-cell-type and equal-cohort aggregation are defined in Appendix~\ref{app:evaluation-metrics}.

\subsubsection{Supplementary Context-Invariance Analysis (RQ3)}
\label{app:supp-context-setup}

To extend evaluation of the context-invariance demand beyond donor-associated variation in the common cohorts, we examine measurement modality as a more direct technical context.
scRNA-seq and snRNA-seq measure whole-cell and nucleus-derived RNA, respectively, and can capture systematically different RNA compartments and transcript compositions~\citep{denisenko2020systematic}.
Matching study collection, anatomical source, donor, and annotated cell type allows us to evaluate robustness to measurement modality while controlling major biological variables.

\paragraph{Cohort selection.}
Using Census metadata, we screened all 89 held-out datasets for collections that separately profiled the same anatomical source using conventional scRNA-seq and snRNA-seq.
Collections represented in the 2025-01-30 pretraining corpus were excluded.
The screen identified three within-collection modality pairs.
An eligible pair required at least one shared donor.
Matched-stratum support was then evaluated in the fixed cross-method evaluation pool, where a donor--cell-type--tissue stratum was eligible when it contained at least 50 observations from each modality.
The ocular-surface atlas did not meet the shared-donor criterion, leaving two eligible pairs.
The RPE/choroid atlas provided 24 eligible strata across four donors and nine cell types, compared with three strata across two donors and three cell types in the anterior-segment atlas.
We therefore retained the RPE/choroid atlas using eligible matched-stratum support as the fixed selection criterion.
The candidate audit and evaluation support are summarized in Table~\ref{tab:rq3-candidate-selection}.

\begin{table}[htbp]
    \centering
    \caption{Metadata-defined screening of matched scRNA-seq/snRNA-seq candidates for RQ3. Eligible-stratum counts are evaluated in the fixed cross-method evaluation pool after matching donor, annotated cell type, and tissue, with at least 50 observations from each modality. $^\dagger$ Evaluation-support entries report represented donors and cell types, in that order.}
    \label{tab:rq3-candidate-selection}
    \small
    \begin{tabular}{@{}lrrrl@{}}
        \toprule
        \textbf{Candidate collection} &
        \shortstack{\textbf{Shared}\\\textbf{donors}} &
        \shortstack{\textbf{Eligible}\\\textbf{strata}} &
        \shortstack{\textbf{Evaluation}\\\textbf{support}$^\dagger$} &
        \textbf{Outcome} \\
        \midrule
        RPE/choroid atlas & 4 & 24 & 4 / 9 & Selected \\
        Anterior-segment atlas & 8 & 3 & 2 / 3 & Lower matched support \\
        Ocular-surface atlas & 0 & 0 & 0 / 0 & No shared donor \\
        \bottomrule
    \end{tabular}
    \vspace{-1ex}
\end{table}

\paragraph{Matched-population construction.}
The selected pair comprises the snRNA-seq dataset with dataset-ID prefix \path{1a9f3b38} and the scRNA-seq dataset with prefix \path{5ed90bc0}.
Both belong to the Single Cell Atlas of the Human Retinal Pigment Epithelium and Choroid collection with collection-ID prefix \path{48c15b0c}.
Both datasets first appear in the 2025-11-08 evaluation release, and no dataset from the same collection is present in the pretraining corpus.

Both datasets use the chorioretinal region as the anatomical source, while their suspension types are nucleus and cell, respectively.
Because anatomical source is fixed, a donor--cell-type stratum is eligible when both modalities contain at least 50 observations.
Within each eligible stratum, we sample without replacement and balance the two modalities to the smaller available population, with at most 100 observations per modality.
This produces 24 strata across four donors and nine cell types.
Their composition and per-repeat sample sizes are summarized in Table~\ref{tab:rq3-strata-composition}.

\begin{table}[htbp]
    \centering
    \caption{Composition of the 24 eligible donor--cell-type strata used in RQ3.}
    \label{tab:rq3-strata-composition}
    \small
    \begin{tabular}{@{}lp{0.61\linewidth}rr@{}}
        \toprule
        \textbf{Donor} & \textbf{Eligible cell types} & \textbf{Strata} & \shortstack{\textbf{Cells per}\\\textbf{repeat}} \\
        \midrule
        \texttt{BCM\_22\_0500} & endothelial cell of venule, fenestrated endothelial cell, fibroblast, macrophage, melanocyte & 5 & 878 \\
        \texttt{BCM\_22\_0698} & endothelial cell of venule, fenestrated endothelial cell, fibroblast, macrophage, melanocyte, pericyte & 6 & 1,068 \\
        \texttt{BCM\_22\_0769} & T cell, endothelial cell of venule, fibroblast, macrophage, melanocyte, retinal pigment epithelial cell & 6 & 1,120 \\
        \texttt{BCM\_22\_0784} & T cell, endothelial cell of venule, fibroblast, macrophage, melanocyte, monocyte, retinal pigment epithelial cell & 7 & 1,290 \\
        \midrule
        \textbf{Total} & 9 unique cell types & 24 & 4,356 \\
        \bottomrule
    \end{tabular}
\end{table}

\paragraph{Evaluation protocol.}
We repeat cell sampling 20 times using seeds 42 through 61, with all methods evaluated on the same sampled cells in each repeat.
Each method is evaluated using its frozen or fixed cell embedding, and every embedding vector is L2-normalized before metric calculation.
Metrics are computed within each donor--cell-type stratum and averaged with equal weight across the 24 strata.
This prevents abundant cell types or those represented across more donors from dominating the aggregate score.
Donor and cell-type annotations are used only for matching and evaluation, not for representation extraction.

\subsubsection{Supplementary Expression-Fidelity Analyses (RQ4)}
\label{app:supp-expression-setup}

To extend evaluation of the expression-fidelity demand beyond the common-cohort disease contrasts, we conduct differential-expression and pathway analysis in disease cohorts selected from the full 89-dataset benchmark.
We also use the full benchmark for a secondary diagnostic of agreement with observed expression.

\paragraph{Cohort selection.}
Using Census metadata, we screened all 89 held-out datasets for explicit normal-versus-disease contrasts.
Cell-count and donor support were then evaluated in the deterministic seed-42 evaluation samples.
A cell type was eligible when both states contained at least 40 cells and each state included at least two donors contributing at least 10 cells each.
A dataset was eligible when at least three cell types met these requirements.
These criteria require disease-associated comparisons to be supported across multiple donors and cell types.

When multiple disease labels qualified within a dataset, one was selected deterministically by the number of supported cell types, the sum of their minimum state counts, total cell support, donor count, total cell count, and lexical order, in that order.
The donor-supported screen identified eight datasets.
Because RQ4 focuses on disease-associated changes, the injury contrast in dataset prefix \path{acd544d0} was excluded, leaving seven disease cohorts.
All seven disease-labeled datasets that passed the screen were included.

\paragraph{Sampling and contrast construction.}
Using seed 42, we sample at most 2,500 cells from each dataset while stratifying by annotated cell type and disease state.
Supported cell types are ordered by minimum state count, total cell count, donor support, and lexical order.
The first six under this ordering are retained per cohort, producing 41 within-cell-type disease-versus-normal contrasts.
Type 2 diabetes contributes five contrasts, and each of the other six cohorts contributes six.
Metrics are averaged within each cohort before cohorts are assigned equal weight.
Table~\ref{tab:rq4-cohort-selection} summarizes the selected cohorts and their evaluation support.

\begin{table}[htbp]
    \centering
    \caption{Disease cohorts retained by the donor-supported screen of all 89 held-out datasets for the supplementary RQ4 expression-fidelity analysis. All seven eligible disease-labeled datasets are included and sampled to 2,500 cells each. $^\dagger$ Evaluation support reports the numbers of supported cell types, selected contrasts, and common genes, in that order. Common genes are the shared gene list of Appendix~\ref{app:supp-expression-setup} after the 2\% detection filter.}
    \label{tab:rq4-cohort-selection}
    \small
    \begin{tabular}{@{}lp{0.48\linewidth}r@{}}
        \toprule
        \textbf{Dataset ID prefix} & \textbf{Disease (vs. normal)} & \shortstack{\textbf{Evaluation}\\\textbf{support}$^\dagger$} \\
        \midrule
        \texttt{30c2a6fd} & Open-angle glaucoma & 15 / 6 / 1,230 \\
        \texttt{84cfa5aa} & Intestinal failure-associated liver disease & 14 / 6 / 2,045 \\
        \texttt{867757c1} & Obstructive nephropathy & 13 / 6 / 1,524 \\
        \texttt{e3ed2ba4} & Metastatic colorectal carcinoma & 11 / 6 / 2,327 \\
        \texttt{203025fe} & Alzheimer disease & 8 / 6 / 1,877 \\
        \texttt{829a3cd1} & Colorectal cancer & 8 / 6 / 1,811 \\
        \texttt{e51bae9a} & Type 2 diabetes mellitus & 5 / 5 / 2,211 \\
        \bottomrule
    \end{tabular}
\end{table}

\paragraph{Shared gene list.}
Each model reconstructs a different gene subset by default: scVI its 8,000 highly variable genes, CellPLM its full 19,374-gene decoder, scPRINT a highly variable subset of its vocabulary, and \proposed{} the 4,096 most expressed genes of the dataset.
Intersecting these subsets would let the most restrictive selection, and hence one model's own choice of readily reconstructed genes, define the evaluation universe.
We therefore decode all four models on one shared list per dataset, fixed before any reconstruction is inspected as the genes expressed in the dataset's 2,500-cell evaluation sample and present in the output gene set or vocabulary of every model, capped at 4,096 genes by variance rank, a cap no dataset reaches (1,875--3,572 genes).
scVI and CellPLM already output these genes, and scPRINT and \proposed{} are decoded on the list directly.
Within each comparison, genes detected in fewer than 2\% of the evaluated cells are excluded, since no model reconstructs them above noise, leaving 1,230--2,327 genes per cohort in the supplementary analysis and 1,524--2,135 in the common cohorts.

\paragraph{Reconstruction alignment and preprocessing.}
Cells are aligned across the observed matrix and the four reconstructions.
Observed expression and each reconstruction are independently normalized to $10^4$ counts per cell and log-transformed.
Applying the same normalization independently prevents model-specific output scales or library-size parameterizations from determining the concordance metrics.

CellPLM reconstruction uses its 19,374-gene negative-binomial decoder.
The held-out datasets do not have batch or dataset identifiers represented during CellPLM pretraining.
We therefore use the official decoder fallback, which replaces each unavailable covariate embedding with the mean of the corresponding learned embeddings.
CellPLM is otherwise evaluated without target-specific parameter updates.

\paragraph{Pathway analysis.}
For each contrast, genes are ranked separately by positive and negative log fold change, and the top 100 genes in each direction are submitted to Enrichr~\citep{kuleshov2016enrichr}.
Enrichment is evaluated using MSigDB Hallmark 2020, Reactome Pathways 2024, and GO Biological Process 2025~\citep{liberzon2015molecular,gillespie2022reactome,gene2015gene}.
The DEG and pathway analyses use the same sampled cells and selected contrasts.

\paragraph{Direct reconstruction agreement.}
As a secondary diagnostic, we compare reconstructed and observed expression across all 89 release-held-out datasets.
We use a deterministic seed-42 sample of at most 2,500 cells per dataset, stratified by annotated cell type and disease when available.
The evaluation contains 220,796 cells in total.

Within each dataset, the comparison uses the shared gene list after the 2\% detection filter, which leaves 891 to 2,842 genes per dataset with a median of 1,848.
Dataset-level scores are given equal weight in the final mean and sample standard deviation.

\section{Implementation Details}
\label{app:implementation-details}
\subsection{Count likelihood details}
\label{app:count-likelihood}

Following common scRNA-seq latent variable and denoising models~\citep{lopez2018deep, eraslan2019single}, we use a zero-inflated negative binomial (ZINB) likelihood over raw counts. For a target count $x_{ij}$, mean $\mu_{ij} > 0$, inverse-dispersion parameter $\theta_{g_{ij}} > 0$, and zero-inflation probability $\pi_{ij} \in (0,1)$, the likelihood is
\[
    p(x_{ij} \mid \mu_{ij}, \theta_{g_{ij}}, \pi_{ij})
    =
    \pi_{ij}\mathbf{1}\{x_{ij}=0\}
    +
    (1-\pi_{ij})
    \mathrm{NB}(x_{ij}; \mu_{ij}, \theta_{g_{ij}}),
    \quad x_{ij}\in\mathbb{Z}_{\ge 0}.
\]

Here $\mathrm{NB}(x; \mu, \theta)$ denotes the negative binomial distribution parameterized by mean $\mu$ and inverse dispersion $\theta$:
\[
    \mathrm{NB}(x; \mu, \theta)
    =
    \frac{\Gamma(x+\theta)}{\Gamma(\theta)\Gamma(x+1)}
    \left(\frac{\theta}{\theta+\mu}\right)^\theta
    \left(\frac{\mu}{\theta+\mu}\right)^x .
\]
In our implementation, decoder gene states parameterize the mean and zero-inflation terms, while $\theta_{g_{ij}}$ is represented by a learned gene-specific parameter. In the active softmax parameterization, decoder outputs define normalized gene proportions $\rho_{ij}$ over the modeled gene subset, and the ZINB mean is scaled as $\mu_{ij}=\ell_i\rho_{ij}$. Here, $\ell_i$ is the raw-count total of the sampled gene crop, rather than the fixed target of $10^4$ used to normalize encoder inputs. 
The resulting negative log-likelihood defines the reconstruction term $\mathcal{L}_{\mathrm{ZINB},i}$ in Equation~\ref{eq:pretrain-objective}.

\subsection{Gene identifier matching and crop construction}
\label{app:preprocessing-details}

Genes are indexed by Ensembl gene IDs rather than gene symbols to avoid ambiguity from non-unique symbols. When matching genes across downloaded shards and the model vocabulary, we strip trailing version-like suffixes such as ``.1'' or ``-1'' and map genes by their base Ensembl ID.

Each training crop contains 4,096 gene positions. If a cell has more than 4,096 expressed genes, we sample expressed genes without replacement. Otherwise, we retain all expressed genes and fill the remaining positions with sampled zero-count genes, which are treated as valid reconstruction targets.

\subsection{Group-derived pseudo-bulk construction}
\label{app:pseudobulk-details}

For a dataset--donor group $b$, $\mathcal{C}_b$ contains all cells assigned to that group across the dataset's shards. Let $x_{c,g}$ denote the raw count of gene $g$ in cell $c$ after mapping genes to the model vocabulary. The pseudo-bulk signature is computed as
\begin{equation}
    p_{b,g}
    =
    \log\left(
        1+
        \frac{1}{|\mathcal{C}_b|}
        \sum_{c\in\mathcal{C}_b}
        \frac{10^4 x_{c,g}}{\sum_{g'\in\mathcal{G}}x_{c,g'}}
    \right),
    \qquad g\in\mathcal{G}.
    \label{eq:pseudobulk-construction}
\end{equation}
Raw counts are therefore restricted to the full model vocabulary, normalized to $10^4$ separately for each cell, averaged across cells with equal cell weight, and transformed by elementwise $\log(1+\cdot)$. The signature is precomputed independently of the sampled 4,096-gene crop and the training mini-batch. The same deterministic procedure is applied to pretraining and held-out groups.

\subsection{Computing environment}
\label{app:computing-environment}

Unless otherwise stated, the main training runs were executed on a Linux GPU node equipped with 8 NVIDIA B200 GPUs. Full pretraining and ablation runs used 4 GPUs on the same node type. Each GPU reported 183,359 MiB of memory, and the NVIDIA driver version was 570.211.01. The software environment was managed with \texttt{uv} and used Python 3.11.14 on Linux 6.8.0. PyTorch reported CUDA 12.8 and cuDNN 91002.
The corresponding Python package versions are provided in Table~\ref{tab:package-versions}.

\begin{table}[htbp]
    \centering
    \small
    \caption{Main Python package versions in the training environment.}
    \label{tab:package-versions}
    \begin{tabular}{ll}
        \toprule
        \textbf{Package} & \textbf{Version} \\
        \midrule
        \texttt{torch} & 2.8.0+cu128 \\
        \texttt{lightning} / \texttt{pytorch-lightning} & 2.6.0 \\
        \texttt{hydra-core} / \texttt{omegaconf} & 1.3.2 / 2.3.0 \\
        \texttt{numpy} / \texttt{scipy} / \texttt{pandas} & 2.3.5 / 1.16.3 / 2.3.3 \\
        \texttt{anndata} / \texttt{scanpy} & 0.12.7 / 1.11.5 \\
        \texttt{scikit-learn} & 1.8.0 \\
        \texttt{torchmetrics} & 1.8.2 \\
        \texttt{cellxgene-census} & 1.17.0 \\
        \texttt{scib-metrics} & 0.5.7 \\
        \texttt{faiss} & 1.13.2 \\
        \bottomrule
    \end{tabular}
\end{table}

\subsection{Runtime and memory profiling}
\label{app:runtime-memory-profiling}

We profile direct embedding extraction on a fixed sample of 2,000 cells from the same held-out dataset for all profiled models.
Each model is run once for warm-up and three times for measurement, with median wall-clock time and throughput reported in Table~\ref{tab:runtime-memory-profile}.
Peak allocated GPU memory is measured with \texttt{torch.cuda.max\_memory\_allocated()} after resetting the allocator statistics.
This process-local value reflects tensor allocations during embedding extraction rather than total device memory reported by \texttt{nvidia-smi}.
For \proposed{}, this path includes the encoder and cell-representation readout but not the reconstruction decoder.

scVI, Geneformer, and \proposed{} were profiled in the shared main environment, and scGPT and CellPLM in their own inference environments on the same GPU with the same protocol.
scPRINT benchmark inference was conducted in a separate environment compatible with its official inference stack and is therefore not included in this local profile.

The peak memory of \proposed{} primarily reflects activation memory rather than parameter storage.
With crop length $N=4096$, latent-token count $K=64$, hidden size $d=512$, and batch size $B=256$, a single dense token tensor of shape $B \times N \times d$ already occupies about 2.15GB in fp32.
Several such tensors and attention workspaces coexist during token construction and cross-attention, so peak allocated memory is larger than the 58.3M trainable parameters alone would suggest.

Measured throughput and peak allocated memory serve as the primary efficiency measures because standard FLOP profilers differ in whether they report multiply-accumulate operations or doubled floating-point operations and inconsistently count sparse preprocessing, embedding lookup, fused attention kernels, and model-specific tokenization.
For a fixed input shape, the dominant encoder-side attention term of \proposed{} scales as $\mathcal{O}(L_{\mathrm{enc}} B K N d)$ for $L_{\mathrm{enc}}$ latent cross-attention blocks, while token construction and feed-forward projections add dense projection costs that scale with the number of cropped gene positions.
FLOP-style operation counts are therefore treated as implementation-dependent estimates.

\begin{table}[htbp]
    \centering
    \small
    \caption{Direct embedding-extraction runtime and memory profile. All rows use a fixed 2,000-cell sample from the same held-out dataset. Throughput is computed from the median runtime. $^{\dagger}$CellPLM processes the sample as a single batch, as in its benchmark pipeline.}
    \label{tab:runtime-memory-profile}
    \resizebox{0.90\linewidth}{!}{%
    \begin{tabular}{@{}lrrrrrr@{}}
        \toprule
        \textbf{Model} & \textbf{Params} & \textbf{Emb. dim} & \textbf{Batch size} & \textbf{Median time (s)} & \textbf{Cells/s} & \textbf{Peak GPU memory} \\
        \midrule
        scVI & 8.0M & 50 & 512 & 0.041 & 49,225.2 & 0.09GB \\
        scGPT & 51.9M & 512 & 64 & 2.141 & 934.0 & 1.52GB \\
        CellPLM & 65.8M & 512 & 2,000$^{\dagger}$ & 0.396 & 5,056.6 & 0.88GB \\
        Geneformer & 316.3M & 1152 & 16 & 32.349 & 61.8 & 3.15GB \\
        \rowcolor{methodhighlight}\proposed{} & 58.3M & 512 & 256 & 6.433 & 310.9 & 19.79GB \\
        \bottomrule
    \end{tabular}
    }
    \vspace{-3ex}
\end{table}

\subsection{Model architecture and hyperparameter settings}
\label{app:model-architecture-hparams}

Table~\ref{tab:data-settings} lists the data, batching, and preprocessing settings for the final \proposed{} model, Table~\ref{tab:optimization-settings} the optimization, loss, and trainer settings, Table~\ref{tab:architecture-settings} the architecture settings, and Table~\ref{tab:pbcond-settings} the pseudo-bulk-derived conditioning settings.

\begin{table}[htbp]
    \centering
    \small
    \caption{Data, batching, and preprocessing settings.}
    \label{tab:data-settings}
    \resizebox{0.99\linewidth}{!}{%
    \begin{tabular}{
            L{0.30\linewidth}
            L{0.26\linewidth}
            L{0.35\linewidth}
        }
        \toprule
        \textbf{Category} & \textbf{Setting} & \textbf{Value} \\
        \midrule
        Training corpus & CELLxGENE Census release & 20250130 \\
        Zero-shot evaluation release & CELLxGENE Census release & 20251108 \\
        Held-out zero-shot benchmark & Release-based benchmark & 89 newly appearing datasets \\
        Counterfactual subset & Held-out blood/protocol subset & 10 blood datasets from the 20251108 release \\
        Random seed & Seed & 42 \\
        Batch size & Cells per batch & 256 per GPU (1,024 total across 4 GPUs) \\
        Crop size & Gene positions per cell & 4,096 \\
        Per-cell preprocessing & Normalize, log1p, target sum & Normalization and log transformation enabled; target sum $10^4$ \\
        Reconstruction target & Raw-expression target & Raw counts for the count likelihood \\
        Minimum expressed genes & Filtering threshold & 5 \\
        \bottomrule
    \end{tabular}
    }
\end{table}

\begin{table}[htbp]
    \centering
    \caption{Optimization, loss, and trainer settings.}
    \label{tab:optimization-settings}
    \resizebox{0.99\linewidth}{!}{%
    \begin{tabular}{
            L{0.30\linewidth}
            L{0.26\linewidth}
            L{0.35\linewidth}
        }
        \toprule
        \textbf{Category} & \textbf{Setting} & \textbf{Value} \\
        \midrule
        Optimizer & AdamW & learning rate $2\times10^{-4}$, weight decay $10^{-4}$, betas $(0.9,0.999)$ \\
        Scheduler & Cosine with linear warmup & 1,250 warmup steps (5\% of 25,000), then cosine decay to step 25,000 \\
        KL loss & KL coefficient, warmup & $5\times10^{-4}$, 2,500 steps (10\% of the 25,000-step recipe) \\
        Trainer & Precision, devices, strategy & bf16-mixed; 4 GPUs; DDP \\
        Trainer schedule & Steps, validation interval & 25,000, 5,000 \\
        Gradient clipping & Clip value & 1.0 \\
        \bottomrule
    \end{tabular}
    }
\end{table}

\begin{table}[htbp]
    \centering
    \small
    \caption{Architecture settings for the final \proposed{} model.}
    \label{tab:architecture-settings}
    \resizebox{0.95\linewidth}{!}{%
    \begin{tabular}{
            L{0.27\linewidth}
            L{0.25\linewidth}
            L{0.38\linewidth}
        }
        \toprule
        \textbf{Component} & \textbf{Setting} & \textbf{Role} \\
        \midrule
        Model family & Latent-bottleneck VAE with cross-attention encoder and decoder & Compresses gene-expression tokens into latent cell tokens before reconstruction. \\
        Trainable parameters & 58,347,460 & Counted from the active final model configuration. \\
        Gene vocabulary & 61,890 Ensembl genes & Gene IDs are indexed by the 20250130 CELLxGENE vocabulary. \\
        Token width & $d_{\mathrm{model}}=512$ & Shared width for gene tokens, latent tokens, attention blocks, and decoder states. \\
        Gene embedding & Learnable, 512-dimensional & Provides the base gene identity token. \\
        Expression-gated gene encoding & Multiplicative expression gate & Gates gene embeddings by measured expression before compression. \\
        Latent bottleneck & 64 latent tokens, each 512-dimensional & Learned latent queries attend to gene tokens and define the bottleneck. \\
        Encoder & 3 cross-attention blocks, 8 heads, dropout 0.1 & Uses RMSNorm pre-normalization and SwiGLU feed-forward layers with hidden multiplier 4.0. \\
        Posterior heads & Linear $\mu$ and $\log\sigma^2$ heads after encoder BatchNorm & Log-variance is clipped to $[-4,2]$. \\
        Evaluated cell embedding & Mean pooling over posterior mean tokens & Produces one 512-dimensional cell embedding per cell. \\
        Cell-representation routing & Z-modulated decoder queries with the same pooling rule as the cell embedding & Uses a pooled latent summary to gate decoder gene queries, while the evaluated embedding is mean-pooled from posterior mean tokens. \\
        Decoder & 3 cross-attention blocks, 8 heads, dropout 0.1 & Decoder queries attend to latent bottleneck tokens for expression reconstruction. \\
        Count head & Linear readout to mean and zero-inflation terms & Uses a learned gene-specific inverse-dispersion parameter for ZINB reconstruction. \\
        Mean activation & Softmax & Converts decoder mean logits into per-cell expression proportions before count scaling. \\
        Zero-inflation output & Probability output & Decoder outputs zero-inflation probabilities rather than logits. \\
        \bottomrule
    \end{tabular}
    }
\end{table}

\begin{table}[htbp]
    \centering
    \small
    \caption{Pseudo-bulk-derived conditioning settings.}
    \label{tab:pbcond-settings}
    \resizebox{0.95\linewidth}{!}{%
    \begin{tabular}{
            L{0.29\linewidth}
            L{0.26\linewidth}
            L{0.36\linewidth}
        }
        \toprule
        \textbf{Component} & \textbf{Setting} & \textbf{Role} \\
        \midrule
        Grouping unit & Dataset-donor group & Defines the collection-level unit for pseudo-bulk profiles. \\
        Pseudo-bulk profile & Per-cell CP10K normalization, group mean, then elementwise $\log(1+\cdot)$ & Produces the group-level expression signature used for conditioning. \\
        Centroid construction & k-means with $M=32$, random seed 42, and five random initializations & Learns expression-derived collection archetypes from pseudo-bulk profiles. \\
        Soft assignment & Euclidean distance to centroids with distance standardization & Produces a 32-dimensional soft pseudo-bulk code. \\
        Zero-shot-release assignment & Fixed 20250130 centroids; no centroid refit on 20251108 & Assigns unseen release groups by soft assignment to the precomputed centroids. \\
        Prior conditioning & Diagonal Gaussian prior & Conditions the VAE prior using the pseudo-bulk code. \\
        Prior parameters & Learned mean and log-variance & Two linear projections map the pseudo-bulk code to the prior mean and log-variance. \\
        \bottomrule
    \end{tabular}
    }
    \vspace{-1ex}
\end{table}

\subsection{PyTorch-style model pseudocode}
\label{app:pytorch-pseudocode}

The following pseudocode summarizes the active \proposed{} training path.
It is written to show where each expression-derived signal enters the model, rather than to reproduce implementation-specific batching, masking, or optimizer code.

\begin{algorithm}[H]
\caption{PyTorch-style pseudocode for \proposed{}.}
\label{alg:overall-framework}
\begin{lstlisting}[style=torchalgorithm,moredelim={**[il][]{@}}]
# C_b: raw counts from one dataset-donor group
# V: model vocabulary, A: frozen pseudo-bulk centroids
# sigma_pb: reference std of pseudo-bulk-to-centroid distances
# T: soft-assignment temperature for centroid weights
def pseudo_bulk_code(C_b, V, A, sigma_pb, T):
    raw = C_b[:, V]
    lib_size = raw.sum(dim=-1, keepdim=True).clamp_min(1.0)
    cpm = 1.0e4 * raw / lib_size
    p_b = torch.log1p(cpm.mean(dim=0))

    dist = torch.cdist(p_b[None, :], A).squeeze(0)
    # normalize distances before temperature-scaled soft assignment
    t_b = torch.softmax(-dist / (sigma_pb * T), dim=0)
    return t_b

for batch in loader:
    x_expr, gene_id = batch["expr_values"], batch["gene_id"]
    y, target_gene_id = batch["target_x"], batch["target_gene_id"]
    valid, t_b = batch["target_mask"], batch["pb_code"]
    @lib = batch["library_size"]  # raw counts summed over the crop

    # expression-gated gene-token construction
    gene_tok = gene_embedding(gene_id)
    expr_feat = expr_proj(x_expr.unsqueeze(-1))
    enc_tok = gene_tok * torch.sigmoid(egate(expr_feat))

    # latent-token bottleneck
    mu_z, logvar_z = encoder(enc_tok, latent_queries)
    z = reparameterize(mu_z, logvar_z)
    cell_repr = mu_z.mean(dim=1)

    # pseudo-bulk-derived prior conditioning
    prior_mu, prior_logvar = pb_prior(t_b).chunk(2, dim=-1)
    prior_mu = prior_mu[:, None, :].expand_as(mu_z)
    prior_logvar = prior_logvar[:, None, :].expand_as(logvar_z)

    # cell-representation route for reconstruction
    cell_summary = z.mean(dim=1)
    query_gate = torch.sigmoid(decoder_query_gate(cell_summary))[:, None, :]
    dec_query = gene_embedding(target_gene_id) * query_gate
    dec_state = decoder(queries=dec_query, memory=z)

    @mean_frac, zi_prob = count_head(dec_state)   # softmax over target genes
    @mean = mean_frac * lib[:, None]
    theta = gene_dispersion(target_gene_id)
    @recon = zinb_nll(y, mean, theta, zi_prob, valid).sum(dim=1).mean()
    @kl = normal_kl(mu_z, logvar_z, prior_mu, prior_logvar).sum(dim=(1, 2)).mean()
    @loss = recon + lambda_kl(step) * kl          # lambda_kl warms up 0 -> 5e-4

    loss.backward()
    optimizer.step()
\end{lstlisting}
\end{algorithm}
\clearpage

\section{Evaluation Metrics}
\label{app:evaluation-metrics}
\subsection{Evaluation Overview}

The metrics fall into two families that score the same three demands at different levels: population-level embedding and reconstruction metrics for the held-out benchmark and the model analyses, and label-defined cohort metrics for the controlled common cohorts.
The metrics below correspond to the release-based RQ1 benchmark, the common-cohort RQ2--RQ4 evaluation, and the supplementary analyses.
Dataset selection and preprocessing are specified in Appendices~\ref{app:datasets} and~\ref{app:preprocessing}.
Cohort selection, cell sampling, and repeat aggregation are specified in Appendix~\ref{app:evaluation-cohorts}.
This section defines the quantities computed from those evaluation sets, and Table~\ref{tab:evaluation-metrics} summarizes them with their optimization direction.
Cell-type, donor, disease-state, and modality annotations are used only after representation extraction for evaluation.

\begin{table}[htbp]
    \centering
    
    \small
    \caption{
    Summary of the evaluation metrics defined in Appendix~\ref{app:evaluation-metrics}. 
    $^{\dagger}$Direction indicates whether higher ($\uparrow$) or lower ($\downarrow$) values are better.}
    \label{tab:evaluation-metrics}
    \resizebox{0.99\linewidth}{!}{%
    \begin{tabular}{
        L{0.30\textwidth}
        L{0.35\textwidth}
        L{0.25\textwidth}
        L{0.10\textwidth}
    }
    \toprule
    \textbf{Evaluation} & \textbf{Metrics} & \textbf{Reported on} & \textbf{Direction}$^{\dagger}$ \\
    \midrule
    RQ1 biological identity & NMI, ARI, ASW-label, cLISI, isolated labels & Cell embedding & $\uparrow$ \\
    RQ1 context invariance & BRAS, iLISI, PCR comparison & Cell embedding & $\uparrow$ \\
    RQ2 biological identity and state & Cell-type BA minus chance, neighbor purity, disease-state BA & Cell embedding & $\uparrow$ \\
    RQ3 context invariance & Residual donor invariance, conditional donor BRAS, local donor mixing & Cell embedding & $\uparrow$ \\
    RQ4 expression fidelity & logFC Spearman, DEG Jaccard and sign, pathway Jaccard and score Spearman & Reconstructed expression & $\uparrow$ \\
    Supplementary identity and state & Program-neighbor Spearman, purity, accuracy, balanced accuracy & Embedding neighborhoods and expression & $\uparrow$ \\
    Supplementary measurement modality & Modality $R^2$, ASW, conditional iLISI & Cell embedding & $\downarrow$, $\uparrow$, $\uparrow$ \\
    Supplementary direct agreement & Pearson, Spearman, MAE, MSE & Reconstructed expression & $\uparrow$, $\uparrow$, $\downarrow$, $\downarrow$ \\
    \bottomrule
    \end{tabular}
    }
\end{table}

\subsection{Release-Based Zero-Shot Benchmark Metrics (RQ1)}
\label{app:rq1-evaluation-metrics}

\paragraph{Biological identity.}
We evaluate whether frozen cell embeddings preserve annotated cell-type structure using scIB-style biological-conservation metrics~\citep{luecken2022benchmarking}. All RQ1 embedding metrics are computed with the \texttt{scib-metrics} package, except that the {k-means} clustering underlying NMI and ARI uses FAISS with 20 initialization seeds for speed, scored with the same scikit-learn functions.
For unsupervised clustering, we run {k-means} with the number of clusters set to the number of annotated cell types in each dataset.
Let $\mathcal{U}$ denote the {k-means} partition and $\mathcal{V}$ the held-out cell-type partition.
NMI is the mutual information between the two partitions normalized by the arithmetic mean of their entropies.
ARI is the chance-adjusted Rand index between the same two partitions.

For ASW-label, let $s_i^{\mathrm{label}}\in[-1,1]$ denote the silhouette coefficient of cell $i$ computed from its annotated cell type.
We report the scIB rescaling $\tfrac{1}{2}(1+\bar{s}^{\mathrm{label}})$, where $\bar{s}^{\mathrm{label}}$ averages $s_i^{\mathrm{label}}$ over cells.

Cell-type LISI (cLISI) measures cell-type purity within local neighborhoods~\citep{korsunsky2019fast,luecken2022benchmarking}.
For cell $i$, let $p_{ic}$ be the perplexity-weighted neighborhood proportion assigned to cell type $c$.
The local inverse Simpson index is
\begin{equation*}
    L_i^{\mathrm{celltype}}
    =
    \left(
        \sum_{c=1}^{C}p_{ic}^{2}
    \right)^{-1},
\end{equation*}
where $C$ is the number of annotated cell types.
We convert the median local index to a higher-is-better score:
\begin{equation*}
    \mathrm{cLISI}
    =
    \frac{
        C-\operatorname{median}_{i}L_i^{\mathrm{celltype}}
    }{
        C-1
    }.
\end{equation*}

For isolated-label silhouette, let $B_c$ denote the number of donors in which cell type $c$ occurs and define
\begin{equation*}
    \mathcal{C}_{\mathrm{iso}}
    =
    \left\{
        c:
        B_c=\min_{c'}B_{c'}
    \right\}.
\end{equation*}
The isolated-label score macro-averages the rescaled cell-type silhouette over the cell types in $\mathcal{C}_{\mathrm{iso}}$.
This metric therefore emphasizes cell types represented in the fewest donor groups.

\paragraph{Context comparability.}
Batch-removal-adapted silhouette (BRAS) evaluates donor comparability within each annotated cell type~\citep{rautenstrauch2026shortcomings}.
For cell $i$, the within-donor distance $a_i$ and cross-donor distance $b_i$ are computed using cosine distance among cells of the same annotated cell type.
Unlike the conventional silhouette, $b_i$ is the mean distance to all cells from other donors rather than the distance to the nearest donor cluster.
With
\begin{equation*}
    s_i^{\mathrm{donor}}
    =
    \frac{b_i-a_i}{\max(a_i,b_i)},
\end{equation*}
BRAS is
\begin{equation*}
    \mathrm{BRAS}
    =
    \frac{1}{|\mathcal{C}^{*}|}
    \sum_{c\in\mathcal{C}^{*}}
    \frac{1}{n_c}
    \sum_{i:y_i=c}
    \left(1-\left|s_i^{\mathrm{donor}}\right|\right),
\end{equation*}
where $\mathcal{C}^{*}$ contains cell types represented by at least two donors.

Integration LISI (iLISI) measures local donor mixing~\citep{korsunsky2019fast,luecken2022benchmarking}.
Let $p_{ib}$ denote the perplexity-weighted neighborhood proportion assigned to donor $b$ and let
\begin{equation*}
    L_i^{\mathrm{donor}}
    =
    \left(
        \sum_{b=1}^{B}p_{ib}^{2}
    \right)^{-1},
\end{equation*}
where $B$ is the number of donors.
We report the normalized score
\begin{equation*}
    \mathrm{iLISI}
    =
    \frac{
        \operatorname{median}_{i}L_i^{\mathrm{donor}}-1
    }{
        B-1
    }.
\end{equation*}
Both cLISI and iLISI use 90-nearest-neighbor graphs with effective perplexity
$\lfloor 90/3\rfloor=30$, following the \texttt{scib-metrics} defaults.

PCR comparison measures the relative reduction in variance explained by donor identity from the pre-integration expression representation to the evaluated embedding, $\mathrm{PCR}_{\mathrm{comp}}=\max\{0,(\mathrm{PCR}_{\mathrm{pre}}-\mathrm{PCR}_{\mathrm{post}})/\mathrm{PCR}_{\mathrm{pre}}\}$.
These metrics are evaluated only for the 75 datasets containing more than one donor.

NMI and ARI are averaged over 20 {k-means} initializations on a fixed stratified sample.
ASW-label, cLISI, isolated-label silhouette, BRAS, iLISI, and PCR comparison are averaged over 20 deterministic stratified samples of up to 10,000 cells per dataset.
Repeat scores are first averaged within each dataset, after which Table~\ref{tab:main_zsb} reports the mean and sample standard deviation across eligible datasets.

Unlike NMI and ARI, cLISI and iLISI summarize label composition at a fixed local-neighborhood scale.
They therefore remain sensitive to neighborhood size, label cardinality and imbalance, and the local feasibility of context mixing.
We interpret clustering, silhouette, local-neighborhood, and variance-removal metrics as complementary rather than interchangeable.

\subsection{Common-Cohort Metrics for the Three Representation Demands (RQ2--4)}
\label{app:common-demand-metrics}

All common-cohort metrics use the same sampled cells described in Appendix~\ref{app:joint-cohort-selection}, and all embedding-based metrics use L2-normalized representations.
Let $c_i$, $y_i$, and $b_i$ denote the annotated cell type, disease state, and donor of cell $i$, respectively.

\paragraph{Biological identity and state.}
We use label-balanced cross-donor retrieval to evaluate whether the embedding preserves cell identity and disease-associated state without relying on same-donor neighbors.
For cell-type retrieval, candidate references for query cell $i$ have the same disease state $y_i$ and a different donor $b_j\neq b_i$.
Before neighbor search, the candidate pool is subsampled to contain the same number of cells from each represented cell type.
The predicted cell type is the majority label among the $k=10$ nearest references by cosine similarity.
Balanced accuracy, $\mathrm{BA}_{\mathrm{celltype}}^{*}$, is computed separately within each disease state, averaged across states, and adjusted by the random baseline $1/C$, where $C$ is the number of eligible cell types in the cohort.
For the same neighborhoods $N_i$, cell-type neighbor purity is
\begin{equation*}
P_i^{\mathrm{celltype}}
=
\frac{1}{|N_i|}
\sum_{j\in N_i}
\mathbb{1}[c_j=c_i],
\end{equation*}
and is macro-averaged across cell types within each disease state and then across states.

Disease-state retrieval reverses the conditioning variables.
Candidate references have the same cell type $c_i$ and a different donor, and the reference pool is balanced across disease states before the $k=10$ neighbor search.
The majority-vote disease-state prediction is evaluated by balanced accuracy within each eligible cell type and then macro-averaged across cell types.

\paragraph{Context invariance.}
To measure donor variation only after controlling the biological variables evaluated above, all context metrics are computed within strata defined jointly by cell type and disease state.
For a stratum $s$, let $\hat{\mathbf{r}}_i$ be the L2-normalized embedding, $\bar{\mathbf{r}}_s$ its overall centroid, and $\bar{\mathbf{r}}_{sb}$ the centroid for donor $b$.
The between-donor and within-donor sums of squares are
\begin{equation*}
SS_{\mathrm{between},s}
=
\sum_b n_{sb}
\left\|\bar{\mathbf{r}}_{sb}-\bar{\mathbf{r}}_s\right\|_2^2,
\quad
SS_{\mathrm{within},s}
=
\sum_b\sum_{i\in(s,b)}
\left\|\hat{\mathbf{r}}_i-\bar{\mathbf{r}}_{sb}\right\|_2^2.
\end{equation*}
With $B_s$ donors, $n_s$ cells, $df_{\mathrm{between}}=B_s-1$, $df_{\mathrm{within}}=n_s-B_s$, and
$MS_{\mathrm{within},s}=SS_{\mathrm{within},s}/df_{\mathrm{within}}$, adjusted donor-associated variance is
\begin{equation*}
\omega_s^2
=
\operatorname{clip}_{[0,1]}
\left(
\frac{
SS_{\mathrm{between},s}
-df_{\mathrm{between}}MS_{\mathrm{within},s}
}{
SS_{\mathrm{between},s}
+SS_{\mathrm{within},s}
+MS_{\mathrm{within},s}
}
\right).
\end{equation*}
Residual donor invariance is $1-\operatorname{mean}_s\omega_s^2$, so higher values indicate that less residual embedding variance is attributable to donor after conditioning on cell type and disease state.

Conditional donor BRAS applies the BRAS definition in Appendix~\ref{app:rq1-evaluation-metrics} with each cell-type--state combination as the biological stratum and donor as the context label.
Local donor mixing measures chance-corrected same-donor enrichment among the $k=10$ nearest neighbors within each stratum.
If $o_s$ is the observed same-donor neighbor fraction and
\begin{equation*}
q_s
=
\frac{\sum_b n_{sb}(n_{sb}-1)}{n_s(n_s-1)}
\end{equation*}
is its random-mixing expectation, the stratum score is
\begin{equation*}
M_s
=
\operatorname{clip}_{[0,1]}
\left(
1-
\frac{o_s-q_s}{1-q_s}
\right).
\end{equation*}
The reported score is the equal-weight mean of $M_s$ across eligible strata.
A value of one denotes random or greater cross-donor mixing, whereas zero denotes complete same-donor segregation.

\paragraph{Expression fidelity.}
Let $\tilde{x}_{ig}$ denote the normalized observed expression of gene $g$ in cell $i$, consistent with the notation in Section~3.1, and let $\widehat{\tilde{x}}_{ig}^{(m)}$ denote the corresponding normalized reconstruction from model $m$.
Both quantities are independently normalized by log1p(CP10K).

For each eligible cell type, let $\mathcal{I}_{+}$ and $\mathcal{I}_{-}$ denote cells from the disease and normal groups, respectively.
The observed expression change is
\begin{equation*}
\Delta_g
=
\frac{1}{|\mathcal{I}_{+}|}
\sum_{i\in\mathcal{I}_{+}}\tilde{x}_{ig}
-
\frac{1}{|\mathcal{I}_{-}|}
\sum_{i\in\mathcal{I}_{-}}\tilde{x}_{ig},
\end{equation*}
and $\widehat{\Delta}_g^{(m)}$ is the same quantity computed from $\widehat{\tilde{x}}_{ig}^{(m)}$.

Let
\begin{equation*}
d_g
=
\frac{1}{|\mathcal{I}_{+}|}
\sum_{i\in\mathcal{I}_{+}}\mathbb{1}[x_{ig}>0]
+
\frac{1}{|\mathcal{I}_{-}|}
\sum_{i\in\mathcal{I}_{-}}\mathbb{1}[x_{ig}>0]
\end{equation*}
denote the sum of observed detection rates in the two groups.
Genes with $d_g>0.02$ are retained for evaluation.
LogFC Spearman, $\rho_{\mathrm{logFC}}^{(m)}$, is the Spearman correlation between $\{\Delta_g\}$ and $\{\widehat{\Delta}_g^{(m)}\}$ over the retained genes.

Let $T$ and $\widehat{T}^{(m)}$ denote the 100 genes with the largest absolute observed and reconstructed changes, respectively.
Top-100 DEG overlap, $J_{\mathrm{DEG}}^{(m)}$, is the Jaccard index of $T$ and $\widehat{T}^{(m)}$, and sign concordance, $S_{\mathrm{DEG}}^{(m)}$, is the fraction of genes in $T$ whose reconstructed change has the same sign as the observed one.

For pathway concordance, the 100 genes with the largest positive and negative changes are analyzed separately against MSigDB Hallmark 2020~\citep{liberzon2015molecular}, Reactome Pathways 2024~\citep{gillespie2022reactome}, and GO Biological Process 2025~\citep{gene2015gene}.
Within each regulation direction and pathway library, let $P$ and $\widehat{P}^{(m)}$ denote the ten highest-ranked terms from observed and reconstructed expression.
Pathway overlap, $J_{\mathrm{path}}^{(m)}$, is the Jaccard index of $P$ and $\widehat{P}^{(m)}$.
Pathway-score Spearman, $\rho_{\mathrm{path}}^{(m)}$, is computed over $P\cup\widehat{P}^{(m)}$ using Enrichr combined scores, with a score of zero assigned when a term is absent from one result.

LogFC Spearman evaluates the ranking of disease-associated changes, DEG Jaccard and sign concordance evaluate recovery of the strongest gene-level changes, and the two pathway metrics evaluate preservation of their pathway-level organization.
Scores are macro-averaged across eligible cell-type contrasts within each cohort and then aggregated with equal cohort weight as specified in Appendix~\ref{app:joint-cohort-selection}.
Pathway metrics are computed from expression changes averaged over the balanced repeats before ranking, rather than scored per repeat.

\subsection{Supplementary Biological-Identity and State Metrics (RQ2)}
\label{app:supp-biological-metrics}

The corresponding cohorts and sampling procedures are described in Appendix~\ref{app:supp-biological-setup}.

\paragraph{Cross-donor lineage-program consistency.}
For cell $i$ and marker program $p$ with gene set $G_p$, the expression-derived program score is
\begin{equation*}
    S_{i,p}
    =
    \frac{1}{|G_p|}
    \sum_{g\in G_p}
    \log\left(1+\frac{10^4 x_{ig}}{\ell_i}\right).
\end{equation*}
For each cell, we identify the $k=30$ cells from other donors with the highest cosine similarity in the evaluated embedding and average their program scores.
For each program, we report the Spearman correlation across evaluation cells between $S_{i,p}$ and the corresponding cross-donor neighbor mean.
A high correlation indicates that cells with similar lineage-associated program activity remain local neighbors after same-donor cells are excluded.
The macro score assigns equal weight to the nine reported marker programs.

\paragraph{Cross-donor disease-state retrieval.}
For query cell $i$, let $N_i$ be its $k=30$ nearest embedding neighbors that have the same annotated cell type and originate from other donors.
Neighborhood purity is defined as in Appendix~\ref{app:common-demand-metrics}, with the disease-state label $y_i$ in place of the cell type.
The neighborhood prediction is $\hat{y}_i=\operatorname{mode}\{y_j:j\in N_i\}$.
Accuracy is the fraction of queries for which $\hat{y}_i=y_i$, whereas balanced accuracy assigns equal weight to recall for the normal and disease states.
Purity, accuracy, and balanced accuracy are first averaged with equal weight across eligible cell types within each cohort and then with equal weight across cohorts.

\subsection{Supplementary Context-Invariance Metrics (RQ3)}
\label{app:supp-context-metrics}

The matched scRNA-seq and snRNA-seq cohort and its sampling procedure are described in Appendix~\ref{app:supp-context-setup}.
For each donor--cell-type stratum $s$, let $\hat{\mathbf{r}}_i$ denote the L2-normalized cell embedding and $m_i\in\{\mathrm{scRNA},\mathrm{snRNA}\}$ the measurement-modality label.

\paragraph{Modality $R^2$.}
Modality $R^2$ is the proportion of total within-stratum embedding variance explained by separation between the two modality centroids:
\begin{equation*}
    R_s^2
    =
    \frac{
    \sum_m n_{sm}
    \left\|\bar{\mathbf{r}}_{sm}-\bar{\mathbf{r}}_s\right\|_2^2
    }{
    \sum_{i\in s}
    \left\|\hat{\mathbf{r}}_i-\bar{\mathbf{r}}_s\right\|_2^2
    }.
\end{equation*}
Here, $\bar{\mathbf{r}}_{sm}$ is the centroid of modality $m$, $\bar{\mathbf{r}}_s$ is the overall stratum centroid, and $n_{sm}$ is the number of observations from modality $m$.

\paragraph{Modality ASW.}
Let $q_i$ be the silhouette of the binary modality label computed with cosine distance within stratum $s$.
We define
\begin{equation*}
    \mathrm{ASW}_s
    =
    1-\frac{1}{|s|}\sum_{i\in s}|q_i|.
\end{equation*}

\paragraph{Conditional iLISI.}
We construct a 30-nearest-neighbor graph separately within each donor--cell-type stratum.
If $p_{i,m}$ is the proportion of neighbors of cell $i$ measured with modality $m$,
\begin{equation*}
    \mathrm{iLISI}_s
    =
    \frac{1}{|s|}
    \sum_{i\in s}
    \left[
    \frac{1}{p_{i,\mathrm{scRNA}}^2+p_{i,\mathrm{snRNA}}^2}-1
    \right].
\end{equation*}
The score approaches zero when a neighborhood contains only one modality and one when both modalities are equally represented.
Metrics are computed within each stratum and then averaged with equal weight across strata.
Conditional iLISI measures local modality-label diversity and cannot by itself distinguish appropriate attenuation of technical variation from excessive mixing of unannotated biological states.
We therefore interpret modality $R^2$, ASW, and conditional iLISI as complementary summaries of overall variance, cell-level distance geometry, and local mixing.

\subsection{Supplementary Expression-Fidelity Metrics (RQ4)}
\label{app:supp-expression-metrics}

\paragraph{Supplementary cohort analysis.}
The supplementary seven-cohort analysis applies the differential-expression and pathway-concordance definitions in Appendix~\ref{app:common-demand-metrics} to the cohorts and contrasts specified in Appendix~\ref{app:supp-expression-setup}.
Scores are averaged over the selected cell-type contrasts, with up to six per cohort, while pathway metrics are additionally averaged over two regulation directions and three pathway libraries.

\paragraph{Direct reconstruction agreement.}
For the full 89-dataset diagnostic, let $\mathcal{G}$ denote the shared gene list of the dataset defined in Appendix~\ref{app:supp-expression-setup}.
$\mathrm{Pearson}^{(m)}$ and $\mathrm{Spearman}^{(m)}$ are computed across genes for each cell and then averaged across the $N$ cells.
$\mathrm{MAE}^{(m)}$ and $\mathrm{MSE}^{(m)}$ average $|\widehat{\tilde{x}}_{ig}^{(m)}-\tilde{x}_{ig}|$ and its square over all cells and genes in $\mathcal{G}$.

Pearson and Spearman evaluate preservation of the relative within-cell expression pattern and gene ranking.
MAE and MSE instead evaluate pointwise calibration to the observed profile.
HVG Pearson restricts the comparison to the 2,000 genes with the highest raw-count variance in each dataset and reports the cell-wise Pearson correlation between observed counts and the ZINB mean, averaged over cells and then over datasets.
Dataset-level scores are aggregated with equal weight across the 89 held-out datasets, and variability is reported as the sample standard deviation across datasets.

\section{Supplementary Experimental Results}
\label{app:supplementary-results}
\subsection{RQ1: Dataset-Characteristic Analysis}
\label{app:rq1-dataset-characteristics}

We examine whether the per-dataset RQ1 gaps vary systematically with held-out dataset characteristics.
Cell-type composition and context structure are computed from the deterministic uniform input cache of up to 100,000 cells per dataset, before metric-specific stratified sampling.
We measure annotated cell-type complexity, abundance-defined long-tail mass as the fraction of cells assigned to types representing less than 1\% of the cache, and raw context burden using the variance explained by the evaluation context before integration, $\mathrm{PCR}_{\mathrm{pre}}$.
To characterize where the aggregate advantages arise, we analyze NMI and ARI for global clustering and BRAS for cross-donor comparability, using isolated-label silhouette as a local control.
Per-dataset gaps are computed against the strongest external baseline for each metric in Table~\ref{tab:main_zsb}.
Associations use Spearman correlation with 2,000 bootstrap resamples and BH-FDR correction over the primary feature--outcome family.
All significant associations reported below remain significant after this correction.
We additionally control for total cell count and repeat the analysis against the strongest external model within each dataset.

\begin{figure*}[htbp]
    \centering
    \includegraphics[width=0.99\textwidth]{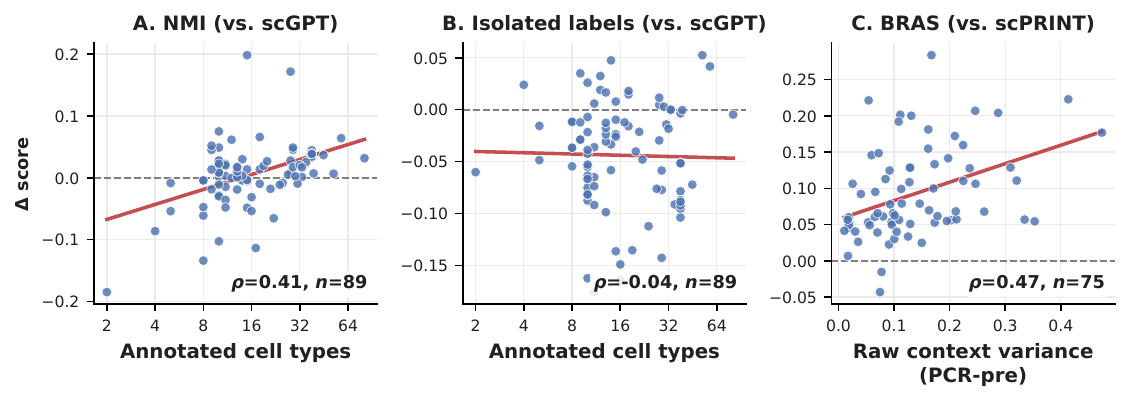}
    \caption{Dataset characteristics associated with RQ1 score differences. Each point is one held-out dataset, and $\Delta$ is \proposed{} minus the fixed external comparator. Panels A and B relate annotated cell-type count to NMI and isolated-label silhouette; Panel C relates raw context burden to BRAS. Red lines show least-squares trends on the displayed scale, dashed lines mark equal performance, and $\rho$ denotes Spearman correlation.}
    \label{fig:rq1-dataset-characteristics}
\end{figure*}

Figure~\ref{fig:rq1-dataset-characteristics} relates the per-dataset gaps to annotated cell-type count in Panels A and B and to raw context burden in Panel C.
The associations summarized in Section~\ref{sec:experiments} hold in these panels.
The identity and BRAS associations retain their direction after controlling for dataset size, replacing the fixed comparator with the per-dataset best external baseline, and leaving out individual datasets.
Tissue- and assay-stratified differences are treated as exploratory because these categories are uneven and correlated with context burden.

\subsection{Supplementary Biological-Identity and State Results (RQ2)}
\label{app:supp-biological-results}

These analyses complement the common-cohort biological-identity and state results in Table~\ref{tab:joint-trilemma} by examining biologically interpretable local structure in cohorts with analysis-specific support.
The cohort-selection and sampling procedures are described in Appendix~\ref{app:supp-biological-setup}, and the metric definitions are provided in Appendix~\ref{app:supp-biological-metrics}.

\paragraph{Cross-donor lineage-program consistency.}
We first test whether cell-to-cell differences in lineage-associated program activity are reflected in local embedding geometry, using a held-out normal-blood cohort and the nine marker programs.

\proposed{} ranks first for the CD4 T-cell, B-cell, and dendritic-cell programs and second for the other six, giving the highest macro score in Table~\ref{tab:rq2-lineage-programs}.
Its top-two placement across all nine programs shows that the result is not driven by a single program.

\begin{table*}[htbp]
    \centering
    \caption{Cross-donor marker-program consistency in held-out normal blood. Share is the annotated population percentage; ambiguous T-cell labels are excluded from subtype shares. Scores are Spearman correlations between cell program scores and their 30-neighbor cross-donor means. Macro mean weights the nine programs equally. Bold and italics mark the highest and second-highest values.}
    \label{tab:rq2-lineage-programs}
    \small
    \resizebox{0.99\linewidth}{!}{%
    \begin{tabular}{@{}lrccccc>{\columncolor{methodhighlight}}c@{}}
        \toprule
        \textbf{Marker program} & \textbf{Share} & \textbf{scVI} & \textbf{Geneformer} & \textbf{scGPT} & \textbf{CellPLM} & \textbf{scPRINT} & \textbf{\proposed{}} \\
        \midrule
        CD4 T cell & 19.3 & 0.8832 & \emph{0.8983} & 0.8869 & 0.8820 & 0.8575 & \textbf{0.9070} \\
        CD8/cytotoxic T & 15.0 & 0.8411 & \textbf{0.8581} & 0.8256 & 0.8032 & 0.7853 & \emph{0.8468} \\
        B cell & 13.5 & \emph{0.5883} & 0.5850 & 0.5730 & 0.5752 & 0.5511 & \textbf{0.5885} \\
        NK cell & 11.7 & 0.7811 & \textbf{0.7819} & 0.7713 & 0.7637 & 0.7686 & \emph{0.7811} \\
        Classical monocyte & 5.8 & 0.6446 & 0.6518 & 0.6554 & 0.6457 & \textbf{0.6679} & \emph{0.6656} \\
        Non-classical mono. & 3.8 & 0.7471 & \textbf{0.7620} & 0.7548 & 0.7422 & 0.7271 & \emph{0.7605} \\
        Dendritic cell & 7.8 & 0.7087 & \emph{0.7130} & 0.7076 & 0.7003 & 0.6978 & \textbf{0.7250} \\
        Platelet & 3.2 & \textbf{0.5989} & 0.5510 & 0.5186 & 0.4510 & 0.5370 & \emph{0.5778} \\
        Plasma cell & 1.6 & 0.4923 & \textbf{0.5039} & 0.4993 & 0.4846 & 0.4974 & \emph{0.5002} \\
        \midrule
        \textbf{Macro mean} & -- & 0.6984 & \emph{0.7006} & 0.6880 & 0.6720 & 0.6766 & \textbf{0.7058} \\
        \bottomrule
    \end{tabular}
    }
\end{table*}

\paragraph{Cross-donor disease-state retrieval.}
We next test whether disease-associated state remains locally recoverable when assay, anatomical tissue, and cell type are held fixed, across four held-out non-blood cohorts.

Table~\ref{tab:rq2-disease-cohorts} reports both cohort-level results and equal-cohort summaries.
\proposed{} achieves the highest equal-cohort mean for purity, accuracy, and balanced accuracy.
It leads in the colon-adenocarcinoma and interstitial-lung-disease cohorts.
It also gives the highest accuracy in bronchopulmonary dysplasia, whereas scPRINT performs best in ulcerative colitis.
These cohort-level results show that the aggregate advantage of \proposed{} reflects strong performance across multiple disease settings while retaining substantial cohort-specific heterogeneity.
Together, the two supplementary analyses reinforce the biological-identity and state demand by showing cross-donor lineage-program coherence and disease-state structure in analysis-specific cohorts.

\begin{table*}[htbp]
    \centering
    \caption{Cohort-level within-cell-type disease-state retrieval in the supplementary biological-state analysis. Each cohort contains 12,000 sampled cells, and same-donor neighbors are excluded. The final block reports equal-cohort means, with sample SD across cohorts shown in parentheses. ILD, BPD, and UC denote interstitial lung disease, bronchopulmonary dysplasia, and ulcerative colitis. Bold and italics mark the highest and second-highest values within each cohort and metric, respectively.}
    \label{tab:rq2-disease-cohorts}
    \small
    \newcommand{\tabmeanstdapp}[2]{\begin{tabular}[c]{@{}c@{}}$#1$\\[-2pt]{\scriptsize $(#2)$}\end{tabular}}
    \begin{tabular}{@{}llccccc>{\columncolor{methodhighlight}}c@{}}
        \toprule
        \textbf{Cohort} & \textbf{Metric} & \textbf{scVI} & \textbf{Geneformer} & \textbf{scGPT} & \textbf{CellPLM} & \textbf{scPRINT} & \textbf{\proposed{}} \\
        \midrule
        Colon & Purity & \emph{0.6432} & 0.6371 & 0.6312 & 0.6310 & 0.6249 & \textbf{0.6583} \\
                             & Accuracy & 0.7276 & 0.7267 & 0.7274 & 0.7093 & \emph{0.7299} & \textbf{0.7507} \\
              & Balanced acc. & 0.6499 & \emph{0.6511} & 0.6496 & 0.6220 & 0.6482 & \textbf{0.6733} \\ [4pt]
        ILD & Purity & \emph{0.6083} & 0.6082 & 0.5994 & 0.5952 & 0.5962 & \textbf{0.6385} \\
            & Accuracy & 0.6920 & \emph{0.6941} & 0.6868 & 0.6831 & 0.6830 & \textbf{0.7359} \\
            & Balanced acc. & 0.5802 & \emph{0.5828} & 0.5750 & 0.5659 & 0.5667 & \textbf{0.6294} \\ [4pt]
        BPD & Purity & 0.5088 & \textbf{0.5395} & 0.5177 & 0.5149 & 0.4972 & \emph{0.5363} \\
            & Accuracy & 0.5228 & \emph{0.5728} & 0.5431 & 0.5411 & 0.5129 & \textbf{0.5772} \\
            & Balanced acc. & 0.4469 & \textbf{0.5230} & 0.4869 & 0.4698 & 0.4294 & \emph{0.5192} \\ [4pt]
        UC & Purity & 0.4929 & \emph{0.5583} & 0.5379 & 0.5070 & \textbf{0.6050} & 0.5439 \\
           & Accuracy & 0.4736 & \emph{0.5784} & 0.5500 & 0.5129 & \textbf{0.6450} & 0.5589 \\
           & Balanced acc. & 0.3967 & \emph{0.5044} & 0.4770 & 0.4440 & \textbf{0.5758} & 0.4862 \\
        \midrule
        \textbf{Mean} & Purity & \tabmeanstdapp{0.5633}{0.0738} & \tabmeanstdapp{\mathit{0.5858}}{0.0449} & \tabmeanstdapp{0.5716}{0.0528} & \tabmeanstdapp{0.5620}{0.0609} & \tabmeanstdapp{0.5808}{0.0570} & \tabmeanstdapp{\mathbf{0.5943}}{0.0631} \\
                          & Accuracy & \tabmeanstdapp{0.6040}{0.1246} & \tabmeanstdapp{\mathit{0.6430}}{0.0790} & \tabmeanstdapp{0.6268}{0.0942} & \tabmeanstdapp{0.6116}{0.0989} & \tabmeanstdapp{0.6427}{0.0932} & \tabmeanstdapp{\mathbf{0.6557}}{0.1016} \\
                          & Balanced acc. & \tabmeanstdapp{0.5184}{0.1169} & \tabmeanstdapp{\mathit{0.5653}}{0.0663} & \tabmeanstdapp{0.5471}{0.0813} & \tabmeanstdapp{0.5254}{0.0830} & \tabmeanstdapp{0.5550}{0.0913} & \tabmeanstdapp{\mathbf{0.5770}}{0.0887} \\
        \bottomrule
    \end{tabular}
\end{table*}

\subsection{Supplementary Context-Invariance Results (RQ3)}
\label{app:supp-context-results}

This analysis complements the donor-conditioned common-cohort context-invariance results in Table~\ref{tab:joint-trilemma} by isolating acquisition modality as an orthogonal technical context.
Donor identity, the within-dataset context label used in RQ1, mixes technical effects with genuine biological differences between individuals, so it does not isolate nuisance variation by itself.
We therefore compare scRNA-seq and snRNA-seq with the study collection, anatomical source, donor, and annotated cell type held fixed, which leaves measurement modality as the dominant remaining source of technical variation.
The cohort-selection and matched-population procedures are described in Appendix~\ref{app:supp-context-setup}, and the metric definitions are provided in Appendix~\ref{app:supp-context-metrics}.
Because the two modalities are matched as populations rather than as individual cells, the comparison is between cells and nuclei drawn from the same donor and cell type, not between paired measurements of the same cell.

Table~\ref{tab:rq3-modality-invariance} reports the equal-weight average across the matched strata, and is the primary result of this analysis.
\proposed{} achieves the lowest modality $R^2$ and the highest modality ASW.
These two metrics summarize modality separation at different levels---centroid displacement and cell-level distance geometry---so their agreement indicates that measurement modality plays a consistently smaller role in structuring the evaluated embedding, rather than an advantage that holds only under a single geometric summary.
Conditional iLISI probes a fixed local neighborhood scale and yields a different ordering, which we read through the caveat in Appendix~\ref{app:supp-context-metrics}.

\begin{table*}[htbp]
    \vspace{-1ex}
    \centering
    \caption{Matched scRNA-seq/snRNA-seq context invariance. Values are means across 20 balanced cell resamples. Arrows indicate the favorable direction. Bold and italics mark the best and second-best values, respectively.}
    \label{tab:rq3-modality-invariance}
    \small
    \begin{tabular}{lccccc>{\columncolor{methodhighlight}}c}
        \toprule
        \textbf{Metric} &
        \textbf{scVI} & 
        \textbf{Geneformer} & 
        \textbf{scGPT} & 
        \textbf{CellPLM} & 
        \textbf{scPRINT} &
        \textbf{\proposed{}} \\ 
        \midrule
        Modality $R^2$ $\downarrow$ & $0.4192$ & $\mathit{0.3645}$ & $0.5357$ & $0.7190$ & $0.3913$ & $\mathbf{0.2677}$ \\
        Modality ASW $\uparrow$ & $0.4265$ & $\mathit{0.4765}$ & $0.3169$ & $0.1696$ & $0.4523$ & $\mathbf{0.5932}$ \\
        Conditional iLISI $\uparrow$ & $\mathbf{0.0755}$ & $0.0324$ & $0.0259$ & $\mathit{0.0356}$ & $0.0270$ & $0.0329$ \\
        \bottomrule
    \end{tabular}
\end{table*}

\begin{figure}[htbp]
    \centering
    \includegraphics[width=0.99\linewidth]{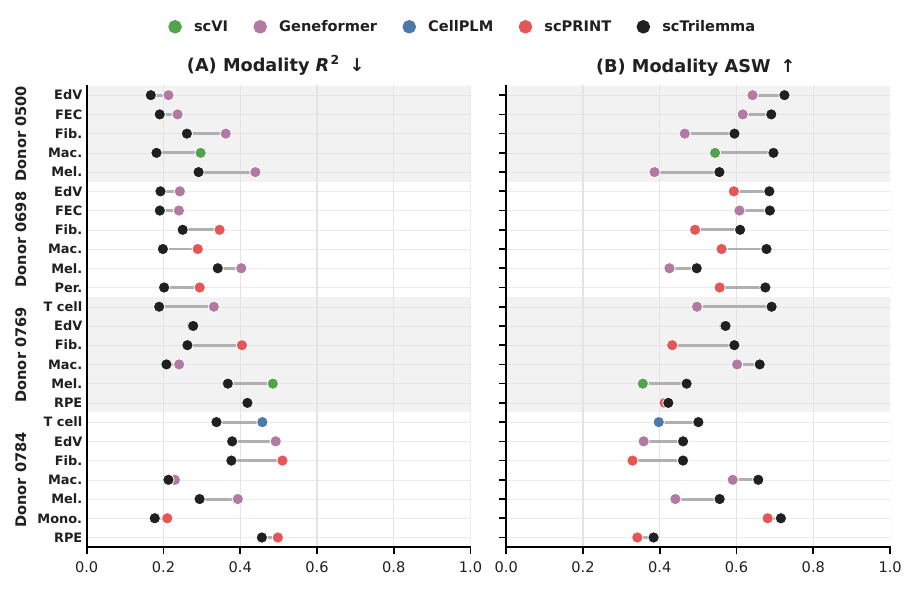}
    \caption{Stratum-level matched-modality results in RQ3. Values are means over 20 balanced cell resamples. (A) Modality $R^2$ and (B) modality ASW for each donor--cell-type stratum; donor labels omit the common \texttt{BCM\_22\_} prefix. Black points are \proposed{}, and colored points are the strongest external result per stratum among scVI, Geneformer, scGPT, CellPLM, and scPRINT, connected to \proposed{} by a line. EdV, endothelial cell of venule; FEC, fenestrated endothelial cell; Fib., fibroblast; Mac., macrophage; Mel., melanocyte; Per., pericyte; Mono., monocyte; RPE, retinal pigment epithelial cell.}
    \label{fig:rq3-stratum-context}
\end{figure}

Because an equal-weight average can be carried by a few strata, Figure~\ref{fig:rq3-stratum-context} resolves the two aggregate separation measures into their 24 constituent donor--cell-type strata.
\proposed{} gives the lowest modality $R^2$ in 23 of them and the highest modality ASW in all of them, across strata spanning vascular, stromal, myeloid, lymphoid, and pigmented populations.
In the single $R^2$ exception, an endothelial-cell-of-venule stratum from donor \texttt{BCM\_22\_0769}, \proposed{} and scVI are effectively tied.
The aggregate advantage in Table~\ref{tab:rq3-modality-invariance} therefore reflects a consistent stratum-level pattern rather than a particular donor or biological population.

All metrics are computed from the same cell embedding used in RQ1 and RQ2, so the observed modality robustness is a property of the evaluated representation rather than of a separately batch-corrected auxiliary space.
Overall, \proposed{} most strongly limits the extent to which measurement modality structures the matched embedding geometry, with its advantage clearest at the within-stratum scale rather than in every local neighborhood.

\subsection{Supplementary Expression-Fidelity Results (RQ4)}
\label{app:supp-expression-results}

This analysis complements the common-cohort expression-fidelity results in Table~\ref{tab:joint-trilemma} by evaluating biologically interpretable gene-level variation in analysis-specific disease cohorts and, as a secondary diagnostic, direct agreement with observed expression across the full held-out benchmark.
Expression fidelity does not mean literal reproduction of observed counts; it denotes the preservation of gene-level biological variation required for expression-level analysis.
The cohort selection, contrast construction, and reconstruction procedures are described in Appendix~\ref{app:supp-expression-setup}, and the metric definitions are provided in Appendix~\ref{app:supp-expression-metrics}.

\paragraph{Differential-expression and pathway concordance.}
We first test whether reconstructed profiles retain disease-associated differential-expression contrasts and their pathway-level organization across seven release-held-out disease cohorts.

\begin{table*}[htbp]
    \centering
    \caption{Cohort-level and equal-cohort differential-expression and pathway concordance in the supplementary expression-fidelity analysis. Each cohort value averages its within-cell-type disease-versus-normal contrasts; the final block reports equal-cohort means with sample SD across cohorts in parentheses. AD, Alzheimer disease; CRC, colorectal cancer; IFALD, intestinal failure-associated liver disease; mCRC, metastatic colorectal carcinoma; OAG, open-angle glaucoma; ON, obstructive nephropathy; T2D, type 2 diabetes mellitus. Arrows indicate the favorable direction, and bold and italics mark the highest and second-highest values within each block and metric.}
    \label{tab:rq4-deg-pathway-concordance}
    \small
    \newcommand{\tabmeanstdapp}[2]{\begin{tabular}[c]{@{}c@{}}$#1$\\[-2pt]{\scriptsize $(#2)$}\end{tabular}}
    \resizebox{0.99\linewidth}{!}{%
    \begin{tabular}{@{}llccccc@{}}
        \toprule
        \textbf{Cohort}
        & \textbf{Method}
        & \shortstack{\textbf{logFC}\\\textbf{Spearman}}$\uparrow$
        & \shortstack{\textbf{top-100 DEG}\\\textbf{Jaccard}}$\uparrow$
        & \shortstack{\textbf{top-100 DEG}\\\textbf{Sign}}$\uparrow$
        & \shortstack{\textbf{Pathway}\\\textbf{Jaccard}}$\uparrow$
        & \shortstack{\textbf{Pathway score}\\\textbf{Spearman}}$\uparrow$ \\
        \midrule
        OAG
        & scVI         & 0.1573 & 0.1379 & 0.6367 & 0.2406 & -0.1662 \\
        & scPRINT      & \textbf{0.5559} & \emph{0.2163} & \emph{0.7550} & 0.2535 & \emph{-0.1124} \\
        & CellPLM      & 0.2194 & 0.1655 & 0.7000 & \emph{0.2594} & -0.1456 \\
        & \cellcolor{methodhighlight}\proposed{} & \cellcolor{methodhighlight}\emph{0.4302} & \cellcolor{methodhighlight}\textbf{0.2498} & \cellcolor{methodhighlight}\textbf{0.8683} & \cellcolor{methodhighlight}\textbf{0.3261} & \cellcolor{methodhighlight}\textbf{0.1589} \\
        \addlinespace
        IFALD
        & scVI         & 0.1932 & 0.1401 & 0.7233 & 0.2247 & -0.1954 \\
        & scPRINT      & \textbf{0.4062} & \emph{0.2528} & \emph{0.7600} & \emph{0.3823} & \emph{0.2449} \\
        & CellPLM      & 0.2383 & 0.2180 & 0.7117 & 0.3207 & 0.1989 \\
        & \cellcolor{methodhighlight}\proposed{} & \cellcolor{methodhighlight}\emph{0.3861} & \cellcolor{methodhighlight}\textbf{0.4171} & \cellcolor{methodhighlight}\textbf{0.8783} & \cellcolor{methodhighlight}\textbf{0.4788} & \cellcolor{methodhighlight}\textbf{0.4989} \\
        \addlinespace
        ON
        & scVI         & 0.1634 & 0.0893 & 0.6933 & 0.1563 & -0.2840 \\
        & scPRINT      & \textbf{0.4947} & \emph{0.2715} & \textbf{0.8233} & 0.1649 & -0.1958 \\
        & CellPLM      & 0.2607 & 0.1902 & 0.7250 & \textbf{0.2193} & \textbf{0.0580} \\
        & \cellcolor{methodhighlight}\proposed{} & \cellcolor{methodhighlight}\emph{0.3338} & \cellcolor{methodhighlight}\textbf{0.2949} & \cellcolor{methodhighlight}\emph{0.8150} & \cellcolor{methodhighlight}\emph{0.1964} & \cellcolor{methodhighlight}\emph{-0.0128} \\
        \addlinespace
        mCRC
        & scVI         & 0.2605 & 0.1770 & 0.7933 & 0.2576 & -0.0732 \\
        & scPRINT      & \textbf{0.5233} & \emph{0.2951} & \emph{0.8167} & 0.2832 & 0.0033 \\
        & CellPLM      & 0.2819 & 0.2463 & 0.8100 & \emph{0.3027} & \emph{0.0737} \\
        & \cellcolor{methodhighlight}\proposed{} & \cellcolor{methodhighlight}\emph{0.3919} & \cellcolor{methodhighlight}\textbf{0.3899} & \cellcolor{methodhighlight}\textbf{0.9150} & \cellcolor{methodhighlight}\textbf{0.3905} & \cellcolor{methodhighlight}\textbf{0.3156} \\
        \addlinespace
        AD
        & scVI         & 0.1538 & 0.0818 & \textbf{0.7050} & 0.1453 & -0.2944 \\
        & scPRINT      & \textbf{0.3558} & \emph{0.1722} & \emph{0.6900} & 0.1620 & \emph{-0.2248} \\
        & CellPLM      & 0.1450 & 0.1256 & 0.6450 & \emph{0.1766} & -0.2312 \\
        & \cellcolor{methodhighlight}\proposed{} & \cellcolor{methodhighlight}\emph{0.2107} & \cellcolor{methodhighlight}\textbf{0.2056} & \cellcolor{methodhighlight}0.6533 & \cellcolor{methodhighlight}\textbf{0.2047} & \cellcolor{methodhighlight}\textbf{-0.1227} \\
        \addlinespace
        CRC
        & scVI         & 0.3313 & 0.1518 & 0.8150 & 0.1965 & -0.1436 \\
        & scPRINT      & 0.2664 & \emph{0.2576} & 0.7433 & 0.2657 & -0.1169 \\
        & CellPLM      & \emph{0.4183} & 0.2482 & \emph{0.8350} & \emph{0.2833} & \emph{0.0361} \\
        & \cellcolor{methodhighlight}\proposed{} & \cellcolor{methodhighlight}\textbf{0.4487} & \cellcolor{methodhighlight}\textbf{0.4095} & \cellcolor{methodhighlight}\textbf{0.9167} & \cellcolor{methodhighlight}\textbf{0.4526} & \cellcolor{methodhighlight}\textbf{0.4783} \\
        \addlinespace
        T2D
        & scVI         & 0.1320 & 0.0886 & 0.6180 & 0.1849 & -0.2782 \\
        & scPRINT      & \textbf{0.3772} & \emph{0.2370} & \emph{0.7720} & 0.2139 & \emph{-0.1804} \\
        & CellPLM      & 0.1492 & 0.1500 & 0.6440 & \emph{0.2195} & -0.2323 \\
        & \cellcolor{methodhighlight}\proposed{} & \cellcolor{methodhighlight}\emph{0.2900} & \cellcolor{methodhighlight}\textbf{0.3181} & \cellcolor{methodhighlight}\textbf{0.7900} & \cellcolor{methodhighlight}\textbf{0.3230} & \cellcolor{methodhighlight}\textbf{0.1977} \\
        \midrule
        \textbf{Mean}
        & scVI         & \tabmeanstdapp{0.1988}{0.0717} & \tabmeanstdapp{0.1238}{0.0371} & \tabmeanstdapp{0.7121}{0.0733} & \tabmeanstdapp{0.2008}{0.0422} & \tabmeanstdapp{-0.2050}{0.0840} \\
        & scPRINT      & \tabmeanstdapp{\mathbf{0.4257}}{0.1035} & \tabmeanstdapp{\mathit{0.2432}}{0.0400} & \tabmeanstdapp{\mathit{0.7658}}{0.0453} & \tabmeanstdapp{0.2465}{0.0764} & \tabmeanstdapp{-0.0831}{0.1630} \\
        & CellPLM      & \tabmeanstdapp{0.2447}{0.0927} & \tabmeanstdapp{0.1920}{0.0477} & \tabmeanstdapp{0.7244}{0.0743} & \tabmeanstdapp{\mathit{0.2545}}{0.0518} & \tabmeanstdapp{\mathit{-0.0346}}{0.1683} \\
        & \cellcolor{methodhighlight}\proposed{} & \cellcolor{methodhighlight}\tabmeanstdapp{\mathit{0.3559}}{0.0839} & \cellcolor{methodhighlight}\tabmeanstdapp{\mathbf{0.3264}}{0.0823} & \cellcolor{methodhighlight}\tabmeanstdapp{\mathbf{0.8338}}{0.0926} & \cellcolor{methodhighlight}\tabmeanstdapp{\mathbf{0.3389}}{0.1110} & \cellcolor{methodhighlight}\tabmeanstdapp{\mathbf{0.2163}}{0.2343} \\
        \bottomrule
    \end{tabular}
    }
\end{table*}

Table~\ref{tab:rq4-deg-pathway-concordance} reports the equal-cohort means in its final block, and is the primary result of this analysis.
\proposed{} has the highest equal-cohort mean on DEG overlap, sign concordance, pathway overlap, and pathway-score correlation, and ranks second on logFC Spearman behind scPRINT.
LogFC Spearman scores the ordering of effect sizes across all shared genes, most of which change little between states, whereas the DEG and pathway metrics score the genes and processes that change most; \proposed{} therefore recovers the disease-associated changes themselves and their pathway organization more consistently, while scPRINT better preserves the ordering of effect sizes across the shared gene list.
The cohort blocks show the same division in nearly every cohort, so neither result reflects a single disease setting.

\paragraph{Direct reconstruction agreement.}
We next compare reconstructed profiles directly with observed expression across all release-held-out datasets, for the four models with reconstruction outputs.

In Table~\ref{tab:rq4-direct-reconstruction}, \proposed{} achieves the highest Pearson and Spearman correlations, indicating that its reconstructions most strongly preserve the relative organization and rank structure of expression within individual cells.
Restricting the comparison to genes detected in at least 20\% of cells widens the correlation gap in favor of \proposed{}, as the second block of Table~\ref{tab:rq4-direct-reconstruction} shows, so the advantage in relative expression structure is largest for the genes on which most analyses rely, while pointwise error remains lower for other models.

\begin{table}[htbp]
    \centering
    \caption{Direct reconstruction agreement across the release-held-out datasets. Values are equal-dataset means $\pm$ sample standard deviations. Arrows indicate the favorable direction, and bold and italics mark the highest and second-highest values within each block. The second block restricts each dataset to genes detected in at least 20\% of its cells.}
    \label{tab:rq4-direct-reconstruction}
    \small
    \begin{tabular}{lcccc}
        \toprule
        \textbf{Method} & \textbf{Pearson} $\uparrow$ & \textbf{Spearman} $\uparrow$ & \textbf{MAE} $\downarrow$ & \textbf{MSE} $\downarrow$ \\
        \midrule
        \multicolumn{5}{l}{\textbf{\textit{All shared genes}}} \\
        \hspace{1.5em} scVI
        & 0.3722 $\pm$ 0.0846
        & 0.3098 $\pm$ 0.0602
        & \emph{1.0041 $\pm$ 0.1157}
        & \emph{1.5788 $\pm$ 0.3013} \\
        \hspace{1.5em} scPRINT
        & 0.4751 $\pm$ 0.0577
        & 0.3962 $\pm$ 0.0617
        & 1.0475 $\pm$ 0.2220
        & 1.6569 $\pm$ 0.5433 \\
        \hspace{1.5em} CellPLM
        & \emph{0.5785 $\pm$ 0.1045}
        & \emph{0.4404 $\pm$ 0.0759}
        & \textbf{0.8135 $\pm$ 0.2247}
        & \textbf{1.3003 $\pm$ 0.5483} \\
        \rowcolor{methodhighlight}\hspace{1.5em} \proposed{}
        & \textbf{0.6380 $\pm$ 0.0863}
        & \textbf{0.4524 $\pm$ 0.0584}
        & 1.4162 $\pm$ 0.2310
        & 2.2828 $\pm$ 0.6831 \\
        \addlinespace[2pt]
        \multicolumn{5}{l}{\textbf{\textit{Genes detected in $\geq$20\% of cells}}} \\
        \hspace{1.5em} scVI
        & 0.3879 $\pm$ 0.1030
        & 0.3458 $\pm$ 0.0801
        & \textbf{1.4050 $\pm$ 0.2756}
        & \textbf{2.6807 $\pm$ 0.9607} \\
        \hspace{1.5em} scPRINT
        & 0.5054 $\pm$ 0.0665
        & 0.4721 $\pm$ 0.0704
        & 1.5870 $\pm$ 0.4450
        & 3.8269 $\pm$ 2.1478 \\
        \hspace{1.5em} CellPLM
        & \emph{0.5454 $\pm$ 0.1181}
        & \emph{0.4827 $\pm$ 0.0902}
        & \emph{1.4396 $\pm$ 0.4795}
        & \emph{3.2736 $\pm$ 2.2960} \\
        \rowcolor{methodhighlight}\hspace{1.5em} \proposed{}
        & \textbf{0.6865 $\pm$ 0.0794}
        & \textbf{0.5885 $\pm$ 0.0680}
        & 1.6907 $\pm$ 0.4413
        & 4.1604 $\pm$ 2.2677 \\
        \bottomrule
    \end{tabular}
\end{table}

\section{Ablation and Intervention Details}
\label{app:model-analysis-section}
\subsection{Demand-targeted interventions}
\label{app:demand-interventions}

Table~\ref{tab:demand-interventions} accompanies Figure~\ref{fig:demand-interventions} with every metric and the per-dataset consistency of each intervention.
The interventions use the stratified 2,500-cell sample of each dataset and the shared gene list of Appendix~\ref{app:supp-expression-setup}; centroid shrinkage uses as many k-means clusters as annotated cell types, so NMI is nearly fixed by construction and Label ASW is the pushed metric, donor centering subtracts the difference between the donor mean and the dataset mean, and latent refinement maximizes the correlation between the reconstruction and the observed profile over thirty optimization steps.
Strengths are limited to $\alpha\le0.75$ because full shrinkage collapses cells onto the centroids and donor centering beyond this point lowers BRAS itself, and a label-centroid push that raises NMI directly, reported in the second block of Table~\ref{tab:demand-interventions}, lowers fidelity in the same way while leaving cross-donor comparability nearly unchanged.
The remaining metrics move with the ones shown in the main text, and the fidelity push slightly raises cross-donor comparability while lowering identity.

\begin{table}[htbp]
    \centering
    \caption{Demand-targeted interventions. Each block pushes one demand at strength $\alpha$; the label-centroid block shrinks cells toward annotated cell-type centroids instead of k-means centroids, and each entry is the mean of the per-dataset change from the unmodified model, with the number of datasets in which the metric increases in parentheses.}
    \label{tab:demand-interventions}
    \resizebox{0.99\linewidth}{!}{%
    \begin{tabular}{@{}lcccccc@{}}
        \toprule
        & \textbf{NMI} $\uparrow$ & \textbf{Label ASW} $\uparrow$ & \textbf{BRAS} $\uparrow$ & \textbf{iLISI} $\uparrow$ & \textbf{Recon.\ Pearson} $\uparrow$ & \textbf{MSE} $\downarrow$ \\
        \midrule
        \multicolumn{7}{l}{\textit{Push identity (centroid shrinkage)}} \\
        \hspace{1em}$\alpha=0.25$ & $-0.002$ (40/89) & $+0.014$ (86/89) & $-0.024$ (0/75) & $+0.003$ (65/75) & $-0.007$ (0/89) & $+0.018$ (88/89) \\
        \hspace{1em}$\alpha=0.5$ & $-0.002$ (46/89) & $+0.031$ (85/89) & $-0.071$ (0/75) & $+0.009$ (69/75) & $-0.023$ (0/89) & $+0.048$ (89/89) \\
        \hspace{1em}$\alpha=0.75$ & $-0.001$ (46/89) & $+0.054$ (77/89) & $-0.175$ (0/75) & $+0.011$ (69/75) & $-0.053$ (0/89) & $+0.087$ (89/89) \\
        \addlinespace
        \multicolumn{7}{l}{\textit{Push identity (label centroids)}} \\
        \hspace{1em}$\alpha=0.25$ & $+0.105$ (89/89) & $+0.030$ (89/89) & $-0.003$ (1/75) & $+0.003$ (61/75) & $-0.004$ (7/89) & $+0.021$ (89/89) \\
        \hspace{1em}$\alpha=0.5$ & $+0.225$ (89/89) & $+0.087$ (89/89) & $-0.006$ (2/75) & $+0.009$ (66/75) & $-0.019$ (0/89) & $+0.053$ (89/89) \\
        \hspace{1em}$\alpha=0.75$ & $+0.306$ (89/89) & $+0.206$ (89/89) & $-0.009$ (4/75) & $+0.016$ (68/75) & $-0.047$ (0/89) & $+0.093$ (89/89) \\
        \addlinespace
        \multicolumn{7}{l}{\textit{Push context invariance (donor centering)}} \\
        \hspace{1em}$\alpha=0.25$ & $+0.006$ (50/75) & $-0.001$ (22/75) & $+0.021$ (74/75) & $+0.016$ (74/75) & $-0.000$ (33/75) & $+0.006$ (74/75) \\
        \hspace{1em}$\alpha=0.5$ & $+0.005$ (39/75) & $-0.002$ (12/75) & $+0.033$ (74/75) & $+0.024$ (72/75) & $-0.001$ (12/75) & $+0.013$ (74/75) \\
        \hspace{1em}$\alpha=0.75$ & $-0.002$ (33/75) & $-0.004$ (6/75) & $+0.036$ (70/75) & $+0.022$ (64/75) & $-0.005$ (1/75) & $+0.023$ (75/75) \\
        \addlinespace
        \multicolumn{7}{l}{\textit{Push expression fidelity (latent refinement)}} \\
        \hspace{1em}$\alpha=0.25$ & $-0.000$ (43/89) & $-0.000$ (39/89) & $+0.001$ (73/75) & $-0.000$ (29/75) & $+0.064$ (89/89) & $-0.049$ (0/89) \\
        \hspace{1em}$\alpha=0.5$ & $-0.002$ (41/89) & $-0.001$ (24/89) & $+0.004$ (75/75) & $-0.000$ (32/75) & $+0.116$ (89/89) & $-0.113$ (0/89) \\
        \hspace{1em}$\alpha=0.75$ & $-0.004$ (28/89) & $-0.002$ (14/89) & $+0.007$ (75/75) & $+0.000$ (41/75) & $+0.147$ (89/89) & $-0.187$ (0/89) \\
        \bottomrule
    \end{tabular}
    }
\end{table}

\subsection{PB-Cond controls}
\label{app:pbcond-controls}

Two controls ask whether the training-time effect of PB-Cond requires the routing architecture.
Table~\ref{tab:pbcond-onoff} switches PB-Cond on and off within three routing configurations, from the bare VAE backbone through E-Gate only to E-Gate with C-Route.
Switching it on changes gene-level reconstruction in every configuration, raising HVG Pearson on the bare backbone and lowering it together with MSE on the E-Gate-only model, whereas the NMI gain appears only in the fully routed model, so its effect depends on the surrounding architecture rather than being a fixed additive gain.
Table~\ref{tab:pbcond-scvi} instead gives the same pseudo-bulk code to a matched scVI as a decoder covariate or as a conditional prior; as a prior, the code trades biological conservation and cross-donor comparability for local donor mixing, with reconstruction Pearson slightly higher and MSE slightly worse.
Together, these controls place the effect of PB-Cond within the routing design rather than in the code itself.

\begin{table}[!htb]
    \centering
    \caption{PB-Cond switched off and on within three routing configurations (backbone, E-Gate only, E-Gate + C-Route). Values are dataset-level means of the on-minus-off change, with SEM in parentheses. Asterisks mark paired two-sided Wilcoxon tests after BH-FDR correction ($^{*}q<0.05$). MSE is the cell-wise mean squared error over the genes detected in at least 2\% of the sampled cells, before restriction to the four-model common list of Appendix~\ref{app:supp-expression-setup}, so its absolute value differs slightly from Table~\ref{tab:rq4-direct-reconstruction}.}
    \label{tab:pbcond-onoff}
    \resizebox{0.99\linewidth}{!}{%
    \begin{tabular}{@{}lcccccc@{}}
        \toprule
        & $\Delta$\textbf{NMI} $\uparrow$ & $\Delta$\textbf{ASW} $\uparrow$ & $\Delta$\textbf{BRAS} $\uparrow$ & $\Delta$\textbf{iLISI} $\uparrow$ & $\Delta$\textbf{HVG Pearson} $\uparrow$ & $\Delta$\textbf{MSE} $\downarrow$ \\
        \midrule
        Backbone & \begin{tabular}[c]{@{}c@{}}$+0.0008$\\[-2pt]{\scriptsize $(0.0015)$}\end{tabular} & \begin{tabular}[c]{@{}c@{}}$+0.0000$\\[-2pt]{\scriptsize $(0.0005)$}\end{tabular} & \begin{tabular}[c]{@{}c@{}}$\mathbf{-0.0035}^{*}$\\[-2pt]{\scriptsize $(0.0009)$}\end{tabular} & \begin{tabular}[c]{@{}c@{}}$-0.0005$\\[-2pt]{\scriptsize $(0.0004)$}\end{tabular} & \begin{tabular}[c]{@{}c@{}}$\mathbf{+0.0571}^{*}$\\[-2pt]{\scriptsize $(0.0061)$}\end{tabular} & \begin{tabular}[c]{@{}c@{}}$+0.0026$\\[-2pt]{\scriptsize $(0.0020)$}\end{tabular} \\
        E-Gate only & \begin{tabular}[c]{@{}c@{}}$-0.0014$\\[-2pt]{\scriptsize $(0.0014)$}\end{tabular} & \begin{tabular}[c]{@{}c@{}}$\mathbf{+0.0008}^{*}$\\[-2pt]{\scriptsize $(0.0002)$}\end{tabular} & \begin{tabular}[c]{@{}c@{}}$\mathbf{-0.0090}^{*}$\\[-2pt]{\scriptsize $(0.0005)$}\end{tabular} & \begin{tabular}[c]{@{}c@{}}$\mathbf{-0.0024}^{*}$\\[-2pt]{\scriptsize $(0.0005)$}\end{tabular} & \begin{tabular}[c]{@{}c@{}}$\mathbf{-0.0341}^{*}$\\[-2pt]{\scriptsize $(0.0037)$}\end{tabular} & \begin{tabular}[c]{@{}c@{}}$\mathbf{-0.0447}^{*}$\\[-2pt]{\scriptsize $(0.0044)$}\end{tabular} \\
        E-Gate + C-Route (full model) & \begin{tabular}[c]{@{}c@{}}$\mathbf{+0.0122}^{*}$\\[-2pt]{\scriptsize $(0.0012)$}\end{tabular} & \begin{tabular}[c]{@{}c@{}}$\mathbf{+0.0018}^{*}$\\[-2pt]{\scriptsize $(0.0002)$}\end{tabular} & \begin{tabular}[c]{@{}c@{}}$\mathbf{+0.0020}^{*}$\\[-2pt]{\scriptsize $(0.0005)$}\end{tabular} & \begin{tabular}[c]{@{}c@{}}$-0.0007$\\[-2pt]{\scriptsize $(0.0005)$}\end{tabular} & \begin{tabular}[c]{@{}c@{}}$\mathbf{+0.0046}^{*}$\\[-2pt]{\scriptsize $(0.0011)$}\end{tabular} & \begin{tabular}[c]{@{}c@{}}$\mathbf{-0.0234}^{*}$\\[-2pt]{\scriptsize $(0.0031)$}\end{tabular} \\
        \bottomrule
    \end{tabular}
    }
\end{table}

\vspace{-1ex}
\begin{table}[!htb]
    \centering
    \caption{Matched scVI configurations that receive the same dataset--donor pseudo-bulk code. The three configurations share the seed, training data, training budget, and VAE recipe and differ only in where group information enters the model. Values are means across the held-out datasets; reconstruction Pearson and MSE are the cell-wise Pearson and mean squared error over all genes on the 76 datasets with complete reconstructions. Asterisks mark paired two-sided Wilcoxon tests against the configuration without a batch covariate ($^{*}p<0.05$; $^{**}p<0.01$; $^{***}p<0.001$).}
    \label{tab:pbcond-scvi}
    \resizebox{0.99\linewidth}{!}{%
    \begin{tabular}{@{}lcccccc@{}}
        \toprule
        & \textbf{NMI} $\uparrow$ & \textbf{ASW} $\uparrow$ & \textbf{BRAS} $\uparrow$ & \textbf{iLISI} $\uparrow$ & \textbf{Recon.\ Pearson} $\uparrow$ & \textbf{MSE} $\downarrow$ \\
        \midrule
        scVI, no batch covariate & $0.6019$ & $0.5597$ & $0.7579$ & $0.1765$ & $0.5127$ & $0.3488$ \\
        + decoder batch covariate & $0.5973$ & $0.5634^{**}$ & $0.7452^{***}$ & $0.1775$ & $0.5243^{***}$ & $0.3497$ \\
        + PB-conditioned prior & $0.5765^{***}$ & $0.5509^{***}$ & $0.7425^{***}$ & $0.1832^{***}$ & $0.5185^{*}$ & $0.3524^{**}$ \\
        \bottomrule
    \end{tabular}
    }
\end{table}

\vspace{-1ex}

\subsection{PB-Cond target-side stress test}
\label{app:pbcond-stress}

Because the pseudo-bulk code enters only the KL prior, the target code cannot change the embedding or the reconstruction of a trained model.
Table~\ref{tab:pbcond-stress} verifies this on the frozen model by recomputing only the target-side code under donor subsampling to 50 cells, rare- and dominant-80\% compositions, a global permutation of the donor--code correspondence, and an all-zero code.
Cell representations, posterior means, and reconstructions are identical in every condition, and only the KL term responds, weakly to the realistic perturbations and more strongly to the shuffle and zero controls.
Table~\ref{tab:pbcond-shuffle} then retrains the model after globally shuffling the donor--code correspondence and scores it with the same protocol alongside the model without PB-Cond.
Shuffling the code leaves identity and cross-donor comparability at or slightly above the level of the full model, whereas removing PB-Cond lowers them, so the representation effect of PB-Cond arises from the learned group-conditional prior rather than from the donor-specific code.
Both reconstruction measures degrade whether the code is removed or shuffled.
Neither the target-side code nor the specific donor-to-code assignment therefore provides a shortcut; the remaining limitation of scope is discussed in Appendix~\ref{app:limitations-section}.

\begin{table*}[htbp]
    \centering
    \small
    \caption{Target-side code perturbations applied to the frozen \proposed{} model. Only the pseudo-bulk code is recomputed; the model, cells, and centroids are fixed. $\Delta$KL is the change in the KL term relative to the matched code, and the last row reports the maximum absolute change in the cell representation, posterior mean, and reconstruction (Max $|\Delta|$ outputs).}
    \label{tab:pbcond-stress}
    \resizebox{0.99\linewidth}{!}{%
    \begin{tabular}{@{}lcccccc@{}}
        \toprule
        & \textbf{Matched} & \textbf{N = 50} & \textbf{Rare-80\%} & \textbf{Dominant-80\%} & \textbf{Global shuffle} & \textbf{Zero code} \\
        \midrule
        $\Delta$KL (\%) & $0.000$ & $+0.027$ & $+0.436$ & $+0.086$ & $+3.931$ & $+3.821$ \\
        Mean $|\Delta \mathbf{c}_b|$ & $0.0000$ & $0.0012$ & $0.0070$ & $0.0024$ & $0.0254$ & $0.0313$ \\
        Max $|\Delta|$ outputs & -- & $0$ & $0$ & $0$ & $0$ & $0$ \\
        \bottomrule
    \end{tabular}}
\end{table*}

\begin{table}[htbp]
    \centering
    \small
    \caption{Retraining with a globally shuffled donor--code correspondence, scored with the protocol of Figure~\ref{fig:component-ablation}. $\Delta$ is the mean of the per-dataset difference from the full model for the model trained with shuffled codes and for the model without PB-Cond. Asterisks mark paired two-sided Wilcoxon tests after BH-FDR correction ($^{*}q<0.05$; $^{**}q<0.01$). MSE is the cell-wise mean squared error over the genes detected in at least 2\% of the sampled cells, before restriction to the four-model common list of Appendix~\ref{app:supp-expression-setup}, so its absolute value differs slightly from Table~\ref{tab:rq4-direct-reconstruction}.}
    \label{tab:pbcond-shuffle}
    \begin{tabular}{@{}lccc@{}}
        \toprule
        \textbf{Metric} & \textbf{Full} & $\Delta$ \textbf{Shuffled codes} & $\Delta$ \textbf{w/o PB-Cond} \\
        \midrule
        NMI $\uparrow$ & $0.6511$ & $+0.0021^{**}$ & $-0.0129^{**}$ \\
        Label ASW $\uparrow$ & $0.5404$ & $-0.0006^{**}$ & $-0.0018^{**}$ \\
        BRAS $\uparrow$ & $0.8396$ & $+0.0025^{**}$ & $-0.0020^{**}$ \\
        iLISI $\uparrow$ & $0.1814$ & $-0.0006^{*}$ & $+0.0007$ \\
        HVG Pearson $\uparrow$ & $0.8852$ & $-0.0043^{**}$ & $-0.0046^{**}$ \\
        MSE $\downarrow$ & $2.2754$ & $+0.0199^{**}$ & $+0.0234^{**}$ \\
        \bottomrule
    \end{tabular}
\end{table}

\subsection{Hyperparameter sensitivity}
\label{app:hparam-sensitivity}

\begin{figure}[htbp]
  \centering
  \includegraphics[width=0.90\linewidth]{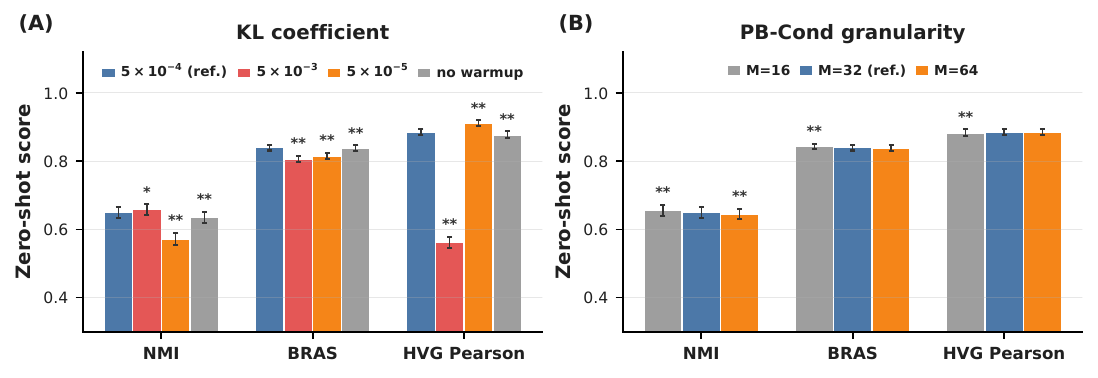}
  \caption{
  Hyperparameter sensitivity.
  (A) KL coefficient: zero-shot scores (mean $\pm$ SEM across datasets) for the reference $\lambda_{\mathrm{KL}}=5\times10^{-4}$ with a 2,500-step warmup, a ten-fold larger and smaller coefficient, and no warmup, with the same seed, batch size, and 25,000 training steps; asterisks mark paired two-sided Wilcoxon tests against the reference after BH-FDR correction ($^{*}q<0.05$; $^{**}q<0.01$).
  (B) PB-Cond codebook size: zero-shot scores for $M=16$, $M=32$, and $M=64$ centroids under the same protocol as (A), with the same tests against $M=32$.
  }
  \label{fig:hparam-sensitivity}
\end{figure}

The objective contains a single balancing coefficient, $\lambda_{\mathrm{KL}}$, and PB-Cond changes the KL reference distribution but not this role.
Figure~\ref{fig:hparam-sensitivity}(A) varies the coefficient around the reference.
A ten-fold larger coefficient collapses expression fidelity and lowers comparability while leaving identity intact, and a ten-fold smaller coefficient trades identity and comparability for expression fidelity.
Removing the warmup lowers all three metrics slightly.
The reference setting therefore sits at the balance point of a reconstruction--representation trade-off governed by the magnitude of $\lambda_{\mathrm{KL}}$, and the warmup adds a small but consistent gain on top of it.
Figure~\ref{fig:hparam-sensitivity}(B) varies the number of pseudo-bulk centroids $M$ used by PB-Cond; across $M\in\{16,32,64\}$ the zero-shot scores remain close on NMI, BRAS, and HVG Pearson, with a smaller codebook tilting slightly toward identity and comparability and the reference toward fidelity, so PB-Cond is not highly sensitive to this granularity in the tested range and we retain $M=32$ as the default.

\section{Mechanistic Analyses}
\label{app:mechanistic-section}
\label{app:mechanistic-analysis}

This appendix asks how each route produces the contribution that the ablation of Section~\ref{sec:ablation} measures.
Every analysis starts from the trained full model and changes one route at a time, either by retraining the model with that route replaced under the same protocol or by intervening on the trained checkpoint at evaluation time, so that the differences isolate the mechanism of that route rather than a design history.
The four parts follow the order of the routes: what E-Gate changes before compression, which demands the pooling and decoder-query routes of C-Route govern, how latent queries route gene-token information through co-expression modules, and how PB-Cond acts through the prior.


\subsection{Expression-gated gene encoding (E-Gate)}
\label{app:egate-analysis}

\begin{figure}[htbp]
  \centering
  \includegraphics[width=\linewidth]{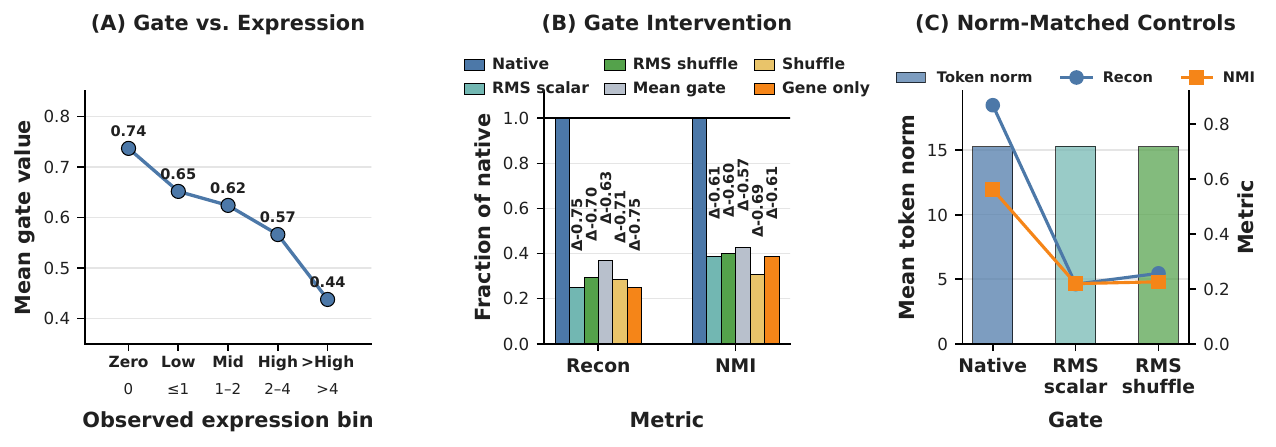}
  \vspace{-5ex}
  \caption{
  Gate profile and same-checkpoint counterfactuals of E-Gate in the full model on the held-out blood/protocol subset.
  (A) Mean gate value per bin of observed expression: zero counts, and input values of at most 1, 1 to 2, 2 to 4, and above 4.
  (B) Reconstruction and NMI retained relative to the native gate after replacing the gate of the trained full model with a norm-matched scalar (RMS scalar), a norm-matched shuffle (RMS shuffle), its feature-wise mean (Mean gate), a shuffle across cell--gene positions (Shuffle), or one (Gene only).
  (C) Mean token norm (bars) with reconstruction and NMI (lines) for the native gate and the two norm-matched controls.
  }
  \label{fig:gene-gate-counterfactual}
\end{figure}

E-Gate is examined on a held-out blood/protocol subset: ten 2025-11-08 release datasets from blood that share major immune cell types but differ in assay, so that dataset identity is not reducible to tissue.
Figure~\ref{fig:gene-gate-counterfactual}(A) reads the gate of Equation~\ref{eq:method-egate} off the trained full model and averages it within bins of observed expression.
Panels (B) and (C) intervene on the same checkpoint, replacing the native gate by controls that remove its feature-wise modulation, its cell--gene assignment, or the gate altogether, two of which are matched to the native gate magnitude on the per-position root-mean-square value.

The gate attenuates tokens more strongly as observed expression rises, so E-Gate acts as expression-dependent gain control before compression, and the same-checkpoint interventions show that the contribution of E-Gate depends on this expression-specific modulation: replacing the native gate with an averaged or shuffled gate sharply reduces reconstruction and NMI, so the gain is not explained by the mere presence of a multiplicative parameterization.
The norm-matched controls keep token norms close to the native gate yet still degrade both metrics, showing that token-magnitude compression alone does not account for the E-Gate gain.

\begin{figure}[htbp]
  \centering
  \includegraphics[width=0.5\linewidth]{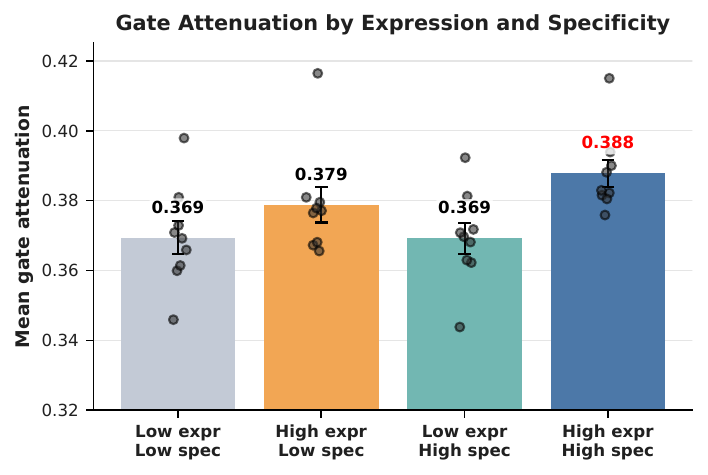}
  \caption{
  Expression-specificity quadrant analysis for E-Gate in the full model on the held-out blood/protocol subset.
  Genes are stratified within each cell type by observed expression magnitude and one-vs-rest cell-type specificity.
  Bars show the mean gate attenuation across analyzed cell types, and points show per-cell-type values.
  }
  \label{fig:gene-gate-quadrants}
\end{figure}

To examine whether the learned gate tracks only expression magnitude or also cell-type-specific biology, we stratify genes within each cell type by observed mean expression and one-vs-rest specificity (Figure~\ref{fig:gene-gate-quadrants}).
Mean gate attenuation is highest for genes that are both highly expressed and cell-type-specific, consistent with marker-aligned gain control: E-Gate limits high-abundance expression from dominating gene-token construction while retaining cell-type-specific salience.

\subsection{Cell-representation routing (C-Route)}
\label{app:croute-analysis}

\begin{figure}[htbp]
  \centering
  \includegraphics[width=\linewidth]{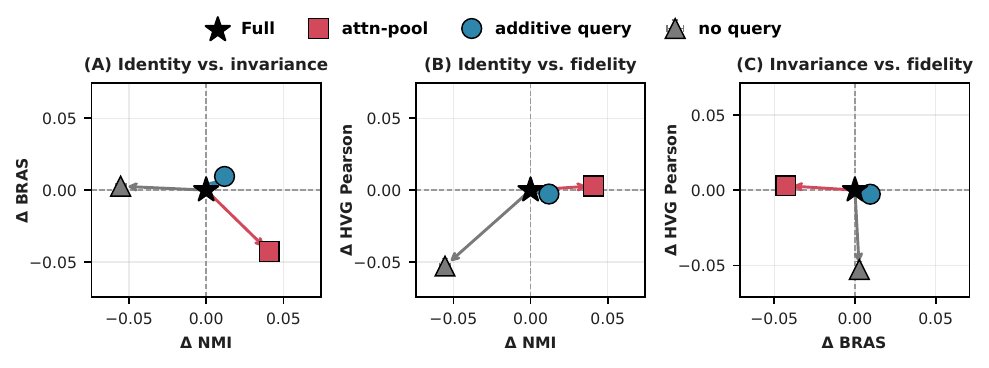}
  \caption{
  C-Route leave-one-out comparison.
  Each point is one retrained model in which a single C-Route component of the full model (Full) is replaced: attention pooling (attn-pool), an additive decoder query (additive query), or no pooled contribution to the decoder queries (no query).
  Axes give the change from Full in dataset means, with Full at the origin, for each pairwise projection of the three demands: (A) $\Delta$NMI against $\Delta$BRAS, (B) $\Delta$NMI against $\Delta$HVG Pearson, and (C) $\Delta$BRAS against $\Delta$HVG Pearson.
  Error bars are standard errors of the paired per-dataset differences.
  }
  \label{fig:croute-pareto}
\end{figure}

\begin{figure}[htbp]
  \centering
  \includegraphics[width=\linewidth]{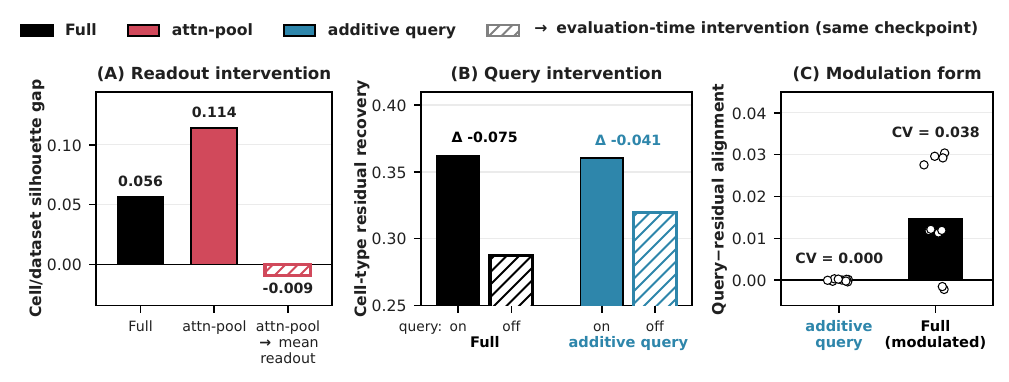}
  \caption{
  Evaluation-time counterfactuals of latent-token readout and decoder-query routing on the held-out blood/protocol subset.
  Colors denote the trained model (Full, attn-pool, additive query) and hatched bars denote an intervention applied to the same checkpoint at evaluation time.
  (A) Cell/dataset silhouette gap of Full, of attn-pool under its native attention readout, and of attn-pool with a forced mean readout of the same latent tokens.
  (B) Cell-type residual recovery of Full and additive query with the pooled representation kept in the decoder queries (on) or removed (off); $\Delta$ gives off minus on.
  (C) Correlation between query-modulation magnitude and the dataset residual for additive query and Full; points are per-dataset values, and CV is the across-gene coefficient of variation of the modulation magnitude.
  }
  \label{fig:cell-routing-counterfactual}
\end{figure}

C-Route is examined with two complementary experiments.
Figure~\ref{fig:croute-pareto} replaces one C-Route component of the full model at a time under the same training protocol, switching the pooling readout to attention (attn-pool), the multiplicative modulation of Equation~\ref{eq:method-croute-query} to an additive query (additive query), or removing the pooled representation from the decoder queries (no query), and scores each retrained model on the zero-shot benchmark.
Figure~\ref{fig:cell-routing-counterfactual} intervenes on these trained checkpoints at evaluation time on the blood/protocol subset introduced above.

The leave-one-out comparison separates the demands each route governs.
Replacing the pooling readout moves comparability the most: attn-pool raises NMI but lowers BRAS, so attention pooling gains cell-state resolution by admitting dataset-sensitive variation into the evaluated embedding, whereas mean pooling keeps the representation comparable across datasets at a small cost in resolution.
Removing the pooled representation from the decoder queries moves identity and fidelity together, lowering both NMI and HVG Pearson with only a small gain in comparability, so the contribution of C-Route to expression fidelity comes from routing the pooled representation into the decoder queries rather than from the form of that routing: the additive query stays close to Full, slightly higher in NMI and BRAS and slightly lower in HVG Pearson.
The two routes therefore trade against different demands rather than competing for the same one.
Mean pooling is retained because it keeps comparability, and the multiplicative form is retained because it gates each gene query separately, which gives the decoder a gene-specific pathway from the pooled representation that the additive form, which adds the same vector to every gene, cannot provide, as the counterfactuals below show.

The counterfactuals confirm that these effects arise from the intended routes.
Forcing a mean readout on attn-pool collapses its cell/dataset silhouette gap below that of Full, with cell-type separation falling as dataset separation rises, so the resolution gained by attention pooling resides in the readout itself rather than in the latent tokens it summarizes.
Removing the cell-representation contribution from the decoder queries reduces the recovery of cell-type residual structure, most clearly under multiplicative modulation, showing that the pooled representation contributes to the native reconstruction path while full latent-token memory retains a second route for expression detail.
Consistently, only multiplicative modulation yields nonzero alignment between query modulation and the dataset residual, indicating a structured decoder-query pathway rather than a contribution routed only through latent memory.


\subsection{Latent-query routing through co-expression modules}
\label{app:latent-query-analysis}

\begin{figure}[htbp]
  \centering
  \includegraphics[width=0.95\linewidth]{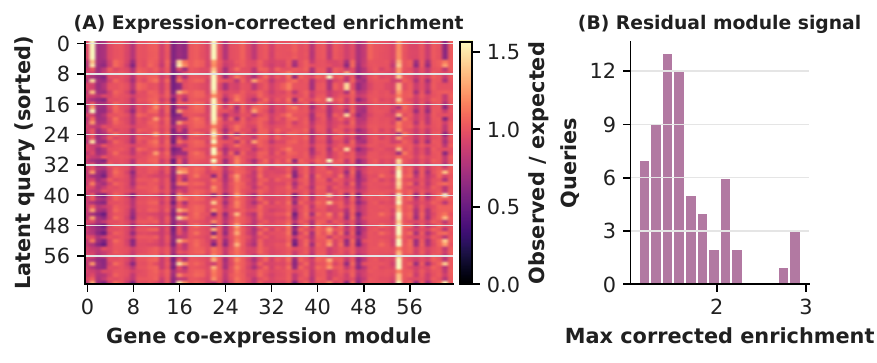}
  \caption{
  Encoder cross-attention enrichment over gene co-expression modules after controlling for mean expression.
  (A) Observed-over-expected attention mass for every latent query (rows) and co-expression module (columns) in the last encoder layer; queries are sorted by their most enriched module.
  (B) Distribution across latent queries of the maximum corrected enrichment attained by each query.
  }
  \label{fig:encoder-crossattn-wgcna}
\end{figure}

Because the encoder compresses gene tokens through cross-attention from learned latent queries, we ask which gene groups the bottleneck reads during cell-level forward passes.
We capture the encoder cross-attention weights of the full model on all held-out datasets and aggregate attention mass over pseudo-bulk-derived gene co-expression modules.
Raw attention is strongly biased toward broadly expressed genes: across last-layer queries, gene attention mass correlates with mean log-expression at a Spearman correlation of $0.656$, and the top-200 attended genes lie at the $98.4$th mean-expression percentile on average.
Figure~\ref{fig:encoder-crossattn-wgcna} therefore reports an expression-corrected enrichment that preserves each query's attention mass across mean-expression deciles and compares the observed module mass with the expected mass within those deciles.

After correction, every latent query still favors a preferred module above its expression-matched expectation, although the residual enrichment is modest in magnitude.
The preferred modules recur across queries, and Table~\ref{tab:latent-module-enrichment} annotates the recurrent ones by post-hoc over-representation analysis of their top attended protein-coding genes, using Enrichr with GO, Reactome, MSigDB Hallmark, CellMarker, and PanglaoDB libraries~\citep{kuleshov2016enrichr,gene2015gene,gillespie2022reactome,liberzon2015molecular,zhang2019cellmarker,franzen2019panglaodb}.
The module preferred by most queries carries a neuronal projection and cell-adhesion signature (module 54), the module with the largest residual enrichment a translation and mitochondrial RNA-processing signature (module 22), and a third an oligodendrocyte/glial signature (module 16), with an additional immune module (module 1).
These results indicate that the latent bottleneck does not attend to genes uniformly; gene-token information is instead routed through partially structured co-expression axes, which we use as interpretability evidence for the bottleneck route rather than as a causal account.

\begin{table}[htbp]
  \centering
  \small
  \setlength{\tabcolsep}{4pt}
  \caption{
  Post-hoc enrichment-backed annotation of recurrent latent attention modules.
  Gene sets are constructed from the top protein-coding genes receiving encoder cross-attention in each module.
  Enriched terms are used only to describe the modules qualitatively.
  }
  \resizebox{0.99\linewidth}{!}{%
  \begin{tabular}{
    c
    L{0.28\linewidth}
    L{0.35\linewidth}
    L{0.22\linewidth}
  }
  \toprule
  \textbf{Module} & \textbf{Top genes} & \textbf{Source-backed enriched terms} & \textbf{Descriptive label} \\
  \midrule
  54 &
  \textit{PTPRD}, \textit{NPAS3}, \textit{NRXN1}, \textit{NLGN1}, \textit{PCDH9}, \textit{LSAMP} &
  Brain and oligodendrocyte-precursor cell-marker terms; immature-neuron terms; neuron projection and adhesion genes &
  Neuronal projection / adhesion \\
  22 &
  \textit{MT-CO1}, \textit{MT-CO2}, \textit{MT-ATP6}, \textit{MT-ND2}, \textit{CALM2}, \textit{B2M} &
  rRNA processing; metabolism of RNA; mitochondrial RNA processing and degradation &
  Translation / mitochondrial RNA axis \\
  16 &
  \textit{EDIL3}, \textit{NCAM2}, \textit{SLC1A2}, \textit{PTN}, \textit{PLP1}, \textit{PTPRZ1} &
  Oligodendrocyte and oligodendrocyte-progenitor marker terms; astrocyte and radial-glia terms &
  Oligodendrocyte / glial \\
  1 &
  \textit{HLA-B}, \textit{PTPRC}, \textit{CD83}, \textit{CCL4}, \textit{MT-ATP8}, \textit{MT-ND4L} &
  Dendritic-cell, NK-cell, and T-cell marker terms; IL-2/STAT5 and TNF-alpha signaling &
  Immune / inflammatory \\
  \bottomrule
  \end{tabular}
  }
  \label{tab:latent-module-enrichment}
\end{table}


\subsection{Pseudo-bulk-derived conditioning (PB-Cond)}
\label{app:pbcond-analysis}

\begin{figure}[htbp]
  \centering
  \includegraphics[width=\linewidth]{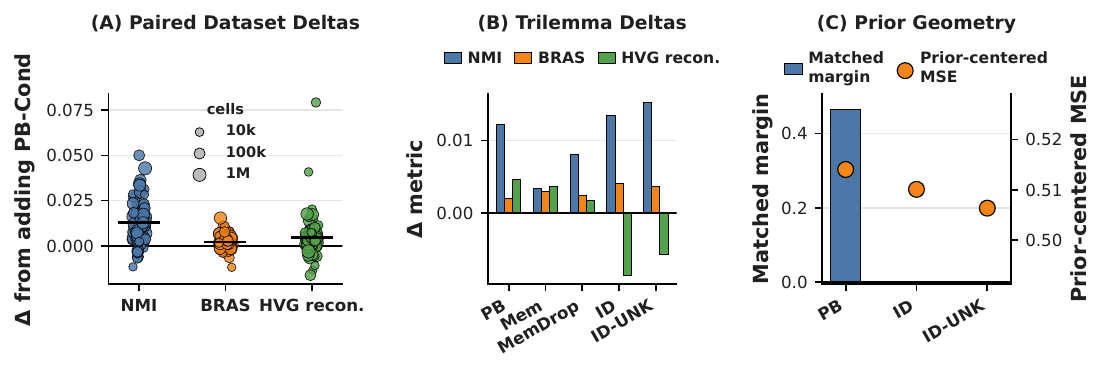}
  \vspace{-5ex}
  \caption{
  PB-Cond ablation and prior-geometry diagnostics.
  (A) Paired per-dataset NMI, BRAS, and HVG Pearson deltas for PB-Cond relative to the E-Gate+C-Route configuration, with dot size proportional to cell count.
  (B) NMI, BRAS, and HVG-reconstruction deltas relative to the same configuration for PB-Cond and for memory-based and learned-identifier context routes.
  (C) Posterior--prior geometry of PB-Cond and learned-identifier prior controls: bars give the matched-random prior margin and points the prior-centered posterior MSE.
  }
  \label{fig:pbcond-prior-ablation}
\end{figure}

Unlike the E-Gate and C-Route counterfactuals above, PB-Cond acts through the KL prior rather than through the encoder or decoder path.
We therefore examine it through three lenses: per-dataset representation deltas, representation and fidelity changes against alternative context routes, and posterior--prior geometry under the conditional prior of Equation~\ref{eq:method-pbcond}, which Figure~\ref{fig:pbcond-prior-ablation} summarizes from a trilemma perspective.

\begin{figure}[htbp]
  \centering
  \includegraphics[width=0.90\linewidth]{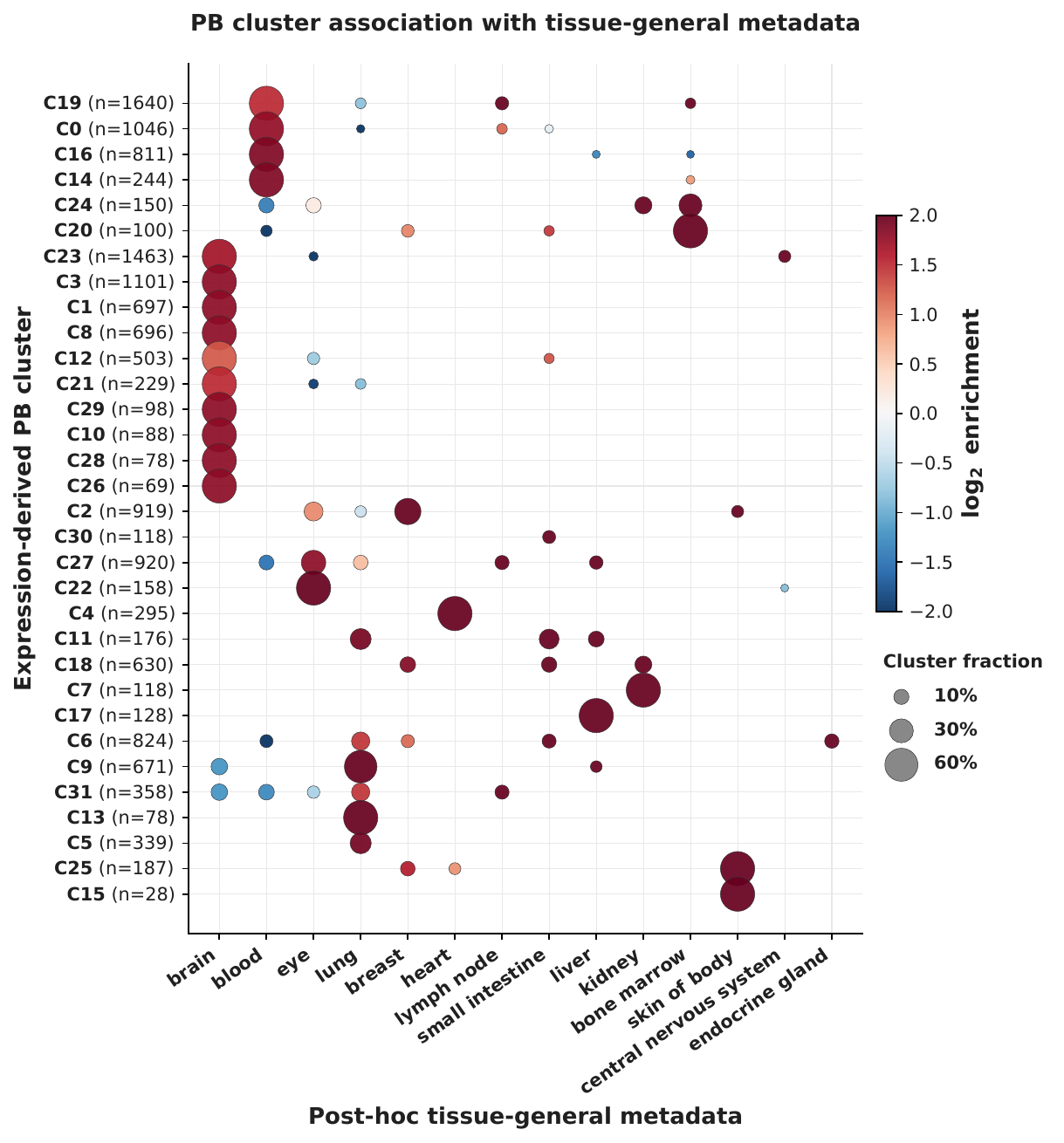}
  \caption{
  Post-hoc metadata annotation of expression-derived PB-Cond clusters.
  Each row is a pseudo-bulk codebook cluster, and each column is a top tissue-general metadata term.
  Dot size denotes the fraction of cells in the cluster assigned to that tissue term, while color denotes enrichment relative to the global background.
  Metadata are used only for annotation after clustering.
  }
  \label{fig:pbcond-cluster-annotation}
\end{figure}

PB-Cond raises per-dataset identity and HVG Pearson relative to the E-Gate+C-Route configuration in most datasets while leaving cross-donor comparability nearly unchanged.
The memory-based routes give the same pseudo-bulk code to the decoder as an additional memory token instead of the prior, with and without token dropout, and the learned-identifier priors replace the code by a one-hot dataset identifier, either covering every dataset or mapping held-out datasets to a shared unknown identifier.
The alternative context routes do not reproduce this profile: the memory-based routes recover only part of the identity gain, and the learned-identifier priors match or exceed it but lower HVG Pearson, so PB-Cond is the only route that improves identity and fidelity together.
The prior-geometry diagnostics show that the matched pseudo-bulk prior lies much closer to the posterior than a random context prior, giving a large matched-random margin, whereas the learned-identifier priors give no margin at all, so their identity gain is not accompanied by dataset-specific alignment between posterior and prior; the prior-centered posterior MSE is similar across the three.
PB-Cond therefore provides a zero-shot-compatible prior-side route that raises per-dataset identity and expression fidelity together without replacing the evaluated cell embedding, and the matched-random margin shows that this gain comes through a prior that is aligned with the posterior, whereas identifier priors reshape the posterior without such alignment and at a cost in fidelity.

We also annotate the PB-Cond codebook after training to ask whether its expression-derived clusters correspond to recognizable tissue-level biology.
Each dataset--donor pseudo-bulk profile is assigned to the cluster with the largest soft {k-means} weight, and Census metadata are then summarized within each cluster; the metadata are used only for this post-hoc annotation and never to construct the codebook.
Figure~\ref{fig:pbcond-cluster-annotation} shows that the clusters recover interpretable tissue-level structure, including blood-, brain-, eye-, breast-, lung-, and heart-enriched groups, supporting PB-Cond as an expression-derived route for tissue-associated context rather than a learned identifier lookup.


\section{Limitations}
\label{app:limitations-section}

Our evidence is limited to expression-derived routing within a reconstruction-trained latent-bottleneck VAE; whether denoising or masked-prediction objectives, or other backbones, route information in the same way is untested.
The gains are clearest at the global scale, in clustering agreement and cross-context comparability, whereas fixed-neighborhood metrics such as isolated-label silhouette and local mixing are less consistent; since these metrics are also sensitive to annotation noise, label imbalance, and neighborhood scale, how much of this difference reflects the representation rather than the metrics remains open.
Expression fidelity is posed as a demand on any expression-level output, but we could verify it only through the reconstruction of a VAE, scoring differential-expression and pathway structure in reconstructed profiles rather than pointwise agreement; whether denoised, imputed, or masked-prediction outputs preserve the same structure, and whether reconstructions can serve as calibrated estimates of absolute expression, remain untested.
PB-Cond relies on pseudo-bulk summaries of dataset--donor groups, which carry both patient-level disease state and study-specific effects~\citep{liu2026learning}; this is why they enter only the prior, and Appendix~\ref{app:pbcond-stress} shows that the trained model does not depend on the target-side summary and that its representation contribution does not require matched codes, so only its reconstruction contribution can be attributed to the code content.
Its remaining limitation is one of scope, since PB-Cond needs a grouping from which pseudo-bulk can be formed and we have not tested groupings beyond dataset--donor.
These boundaries mark where the routing principle should be stress-tested next: perturbation data, disease-enriched cohorts, cross-species transfer, and lower-resource collections.


\end{document}